%% file: main.tex
\documentclass[10pt,twocolumn,letterpaper]{article}

\usepackage[pagenumbers]{wacv} % To force page numbers, e.g. for an arXiv version

\usepackage{graphicx}
\usepackage{amsmath}

\usepackage{booktabs}
\usepackage{rotating}
\usepackage{mathtools}
\usepackage{CJKutf8}
\newtheorem{definition}{Definition}
\usepackage{multirow}
\usepackage{tikz}
\usepackage{amssymb}
\usepackage{mathrsfs}
 
\usepackage{algpseudocode}
\usepackage{algorithm}
\usepackage[accsupp]{axessibility} 
\DeclarePairedDelimiter\abs{\lvert}{\rvert}%

\newcommand{\appendixhead}
    {\centering\textbf{\Large I Spy With My Little Eye\\A Minimum Cost Multicut Investigation of Dataset Frames}\vspace{0.25in}\par}

\usepackage[pagebackref,breaklinks,colorlinks]{hyperref}

\usepackage[capitalize]{cleveref}
\crefname{section}{Sec.}{Secs.}
\Crefname{section}{Section}{Sections}
\Crefname{table}{Table}{Tables}
\crefname{table}{Tab.}{Tabs.}

\def\wacvPaperID{26} % *** Enter the WACV Paper ID here
\def\confName{WACV}
\def\confYear{2025}

\begin{document}

\title{I Spy With My Little Eye\\A Minimum Cost Multicut Investigation of Dataset Frames}

\author{
Katharina Prasse$^{1}$, 
Isaac Bravo$^{2}$, 
Stefanie Walter$^{2}$, 
Margret Keuper$^{1,3}$\\
$^{1}$University of Mannheim, Mannheim, Germany, \texttt{\small katharina.prasse@uni-mannheim.de}\\
$^{2}$Technical University Munich, Munich, Germany, 
\texttt{\small \{isaac.bravo, stefanie.walter\}@tum.de}\\
$^{3}$Max-Planck-Institute for Informatics, Saarland Informatics Campus, \texttt{\small keuper@uni-mannheim.de}
}
\maketitle

%%%%%%%%% ABSTRACT
\begin{abstract}
      Visual framing analysis is a key method in social sciences for determining common themes and concepts in a given discourse.
   To reduce manual effort, image clustering can significantly speed up the annotation process.
   In this work, we phrase the clustering task as a Minimum Cost Multicut Problem [MP]. 
   Solutions to the MP have been shown to provide clusterings that maximize the posterior probability, solely from provided local, pairwise probabilities of two images belonging to the same cluster.
   We discuss the efficacy of numerous embedding spaces to detect visual frames and show its superiority over other clustering methods.
   To this end, we employ the climate change dataset \textit{ClimateTV} which contains images commonly used for visual frame analysis.
   For broad visual frames, DINOv2 is a suitable embedding space, while ConvNeXt V2 returns a larger number of clusters which contain fine-grain differences, \ie speech and protest.
   Our insights into embedding space differences in combination with the optimal clustering - by definition - advances automated visual frame detection. Our code can be found at \url{https://github.com/KathPra/MP4VisualFrameDetection}.
\end{abstract}

%%%%%%%%% BODY TEXT
\section{Introduction}
\label{sec:intro}
% 8 pages
Frame analysis \cite{goffman1974frame} plays a key role in social science research.
This method extracts the main concepts from a dataset, which is generally collected for a single analysis and not shared with the community.
While frame analysis was originally text-centric, visual frame analysis has gained traction in the field, as images have become ever more present in communication.
Visual frames can be formal/stylistic or content-oriented \cite{schaefer2017frame}. 
The detection of formal/stylistic frames can be easily automated with the help of style detection algorithms.
Content-oriented frame detection is a much harder task to automate.
Authors can either use zero-shot classification by selecting suitable frames from previous works, rely on clustering approaches, or a combination thereof.
Automated frame detection is not well-researched and datasets are often manually annotated.
The detection of abstract and diverse concepts remains a challenging task \cite{zhang2022omnibenchmark} and an active field of research.

We argue that clustering is the most suitable approach as it does not rely on pre-defined frames and thus imposes less bias on the outcome of the investigation.
The emergence of novel frames, as investigated among others by O'Neill \cite{o2020more}, can only be detected through this approach - or manually.
Mooseder \etal~employ clustering in their work to reduce the number of manual annotations.
%This branch focuses on images
%This not so novel phrase exemplifies the important nature of data for many applications.
%Anomaly detection, visual communication research, wild life monitoring, social media research all require efficient recognition of patterns in the data.
%Several recent works recognized the potential of clustering embedded images or image patches in order to better understand previously un-lablled data or data with noisy labels.
Similarly, Oquab \etal~\cite{oquab2023dinov2} use clustering to extend their dataset, while Zhou \etal~employ clustering to evaluate their online tokenizer.
Zhang \cite{zhang2022omnibenchmark} employs clustering to clean their dataset, by removing outliers.
%We argue that clustering in its purest form can be leveraged in order to recognize patterns in datasets.
%While previous works have demonstrated the helpfulness of clustering, we argue that use of a probabilistic framework is more beneficial than \textit{k-means}.
To provide clusters within a probabilistically meaningful framework, we propose to phrase the clustering problem as a Minimum Cost Multicut Problem [MP] \cite{chopra1993partition}, in which the images are the nodes of a graph with weighted edges indicating the images' probability of being in the same cluster.
Without any hyperparameters, one can obtain a clustering with the maximum posterior probability. %generating the optimal clustering.

The MP takes image similarities as an input which we generate using several vision or vision and language models. %, \ie~ResNet-50, VGG19-BN, Vision Transformer, ConvNeXt V2, DINOv2, and the CLIP ViT image encoder \cite{radford2021learning, oquab2023dinov2, he2015deepresiduallearningimage, dosovitskiy2020image, woo2023convnext, vgg}. 
%While the distribution of image similarities plays a key role, inter-model differences also have to be accounted for when calculating the edge weights.
The annotated datasets ImageNette \cite{Howard_Imagenette_2019} and ImageWoof \cite{Howard_Imagewoof_2019} are our independent validation  and test set for finding the optimal calibration.
On the dataset ClimateTV \cite{prasse2023towards}, we show and discuss the effectiveness of our approach for social science research.
%With this approach, we can \eg~to detect which animal species have passed a specific wild life watch point or how many frames have been utilized in a topic's communication.
%We show the effectiveness of MP as a clustering objective on several datasets, both meticulously curated and web-scraped.
Our work further investigates embedding space differences both in the embedding and the resulting clusterings.
%Image clustering detects pattern useful for many real life applications, while it can also help to better understand large model's embedding spaces.
%Recently Liang \etal~have reported the cone effect in multi-modal models, as they observed all images' cosine similarities result in a narrow cone, and thus the image embeddings to only occupy a small part of the embedding space \cite{liang2022mind}.
%We further investigate how \textit{narrow cone effect} \cite{ModalityGap} influences the expressiveness of the embeddings and discuss which embedding models are most suitable for which use-case.

Our contributions are: 
(1) We advance automated visual frame detection by extensive clustering analysis.
(2) To this aim, we formulate semantic embedding-based clustering as a Minimum Cost Multicut Problem that maximizes the posterior probability of the clustering.
(3) We analyse the efficacy of powerful vision foundation models for this novel application and provide concrete recommendations on which embedding spaces are most suitable for this task.

\section{Related Work}

Visual frame analysis investigates frames in communication.
Alone in the context of climate change, numerous works have used this method \cite{born2019bearing, schaefer2017frame, o2020more, rebich2015image, mooseder2023social, weaver2022sponsored, hayes2021greta, wang2018public, Casas2019, o2023visual, mcgarry2024fire}.
Advances in automating the annotation process have recently started.
Mooseder \etal~\cite{mooseder2023social} employ the \textit{k-means} algorithm to cluster their images' VGG16 features and then manually annotate 100 random images per cluster.
%The \textit{k-means} algorithm requires the a priori definition of the number of clusters.
They set k= 5,000, which results in the manual annotation of 50,000 images. 
In total, they found 21 distinct frames with 8\% of images excluded based on the clustering results.
Given an optimal clustering for their data, their manual workload would have been drastically lowered, with a lower bound of 2,100 images to manually annotate.
%On top of that, they also have the frame \textit{miscelaneous} to which they assign clusters of images that they are not interested in.

Phrasing a problem as an MP has many applications in computer vision.
The most prominent use case is multi-person tracking \cite{tang2016multi, tang2017multiple, ho2020unsupervised, ho2020two, nguyen2022lmgp}, where it is used to link and cluster person hypotheses over time.
Furthermore, Andres \etal~employ the MP to generate a probabilistic image segmentation \cite{andres2011probabilistic}, followed by \cite{jung2022optimizing, Galasso_2014_CVPR,jung2022learning, kardoost2019solving}.
Keuper \etal~build upon this work and use the MP for efficient image and mesh graph decomposition \cite{keuper2015efficient} and motion segmentation~\cite{KB15b} with several follow-up works~\cite{K17,KB19,kardoost21,kardoost22}. 
Ho \etal~have employed the MP for image clustering \cite{ho2020learning,ho2021msm}.
In contrast to their work, we do not train a deep neural network when computing the inputs for the MP.
While MP clustering has been done in the past \cite{ho2021estimating}, we are the first to use MP clustering in conjunction with foundation model embedding spaces.
Given the recent advances in computer vision, we employ models with highly expressive embedding spaces to create image features and use their pair-wise cosine similarities as inputs.
This approach to obtaining edge costs has been proven successful by Swoboda \etal \cite{abbas2023clusterfug}, who use ResNet-50 features \cite{he2015deepresiduallearningimage}.
The strength of cosine similarities has been ever present and its efficacy for foundation models was highlighted by Radford \etal \cite{radford2021learning}. 

In this work, we leverage vision models' and VLMs' embedding spaces for feature generation.
While clustering using traditional vision models is aptly, foundation models clustering is just gaining traction within the community \cite{yuan2024foundationmodelmakesclustering, baker2024usingmultimodalfoundationmodels, wagner2024just}.
We want to highlight the concurrent work of Wagner \etal who show the efficacy of DINO features for data exploration \cite{wagner2024just}.
In our work, we compare the expressiveness of performant vision model's embedding spaces w.r.t. their capabilities to represent abstract visual concepts.
Visual frame analysis is precisely interested in such concepts, as its goal is to understand the common themes in a given dataset. 
The discussion of embedding space differences both adds to our understanding and advances their applicability in data exploration.
We further strengthen our point by evaluating on OmniBenchmark \cite{zhang2022omnibenchmark}, a computer vision dataset which is designed to test how universal vision features are.

We compare CLIP's general-purpose features which have been trained on a web-scale image-text dataset and have a high zero-shot classification accuracy \cite{radford2021learning} to DINOv2 features, which are trained using optimized training data collection with the goal of increasing features' robustness \cite{oquab2023dinov2}.
With ConvNeXt V2 we include another model which achieves a high classification accuracy \cite{woo2023convnext}.
This architecture contains fully convolutional masked autoencoders and a global response normalization.
%In our investigation of embedding spaces, we analyse the models' embeddings w.r.t.~the cone effect in multi-modal models.
Moreover, we employ ResNet-50, VGG19-BN, and ViT features to assess the expressiveness differences between smaller and larger models.
Liang \etal~observe the CLIP images' cosine similarities resulting in a narrow cone, and conclude that the image embeddings occupy a small part of the embedding space \cite{ModalityGap}.
Based on data distribution comparisons and image cluster analysis, we provide a guideline for the choice of embedding model. 
%, we visualize the image embeddings using UMAP \cite{2018arXivUMAP} in line with \cite{ModalityGap}.

\section{Methods}
The Minimum Cost Multicut problem is a graph problem which involves finding the cutting of the graph into distinct clusters such that the cost, the sum of the cut edges' weights, is minimal.
As we phrase the clustering problem as a Minimum Cost Multicut Problem [MP]\cite{chopra1993partition}, we map the images to a graph structure.
To this end, we embed all images and construct a fully connected graph which edge weights are the images' cosine similarities.
We investigate embedding differences with respect to their expressiveness and effectiveness of finding visual frames.
%To this end, we analyse the clustering of images across embedding models and settings.
We use six models, differing in terms of parameters and classification performance.
ResNet-50 \cite{he2015deepresiduallearningimage}, VGG19-BN \cite{vgg}, Vision Transformer B/32 \cite{vit}, ConvNeXt V2 \cite{woo2023convnext}, and DINOv2 \cite{oquab2023dinov2} are pure vision models, while CLIP ViT-B/32 \cite{radford2021learning} is trained in a multi-modal setting using image-text pairs.
An overview of the employed models and their characteristics can be found in \cref{app:models}.

\subsection{Image Clustering}
We phrase the image clustering task as a MP, also referred to as weighted Correlation Clustering.

\begin{definition}
A finite, undirected graph G = (V,E) with cost $w: E  \rightarrow \mathbb{R} $ associated with the edges is separated into detached components such that the cost is minimal
\begin{equation}
    \min_{\textbf{y} \in \{0,1\}^{\abs{E}}} \hspace{1em} c(\textbf{y}) = \textbf{y}^T\textbf{w} = \sum_{e \in E} w_e y_e ,
\end{equation}
where $y$ is the binary edge label indicating whether the edge should be cut. This is subject to the linear constraint
\begin{equation}
    \forall C \in \text{cycles}(G), \forall e \in C: (1-y_e) \leq \sum_{e' \in C \setminus \{e\}} (1-y_{e'}) .
\end{equation}
\end{definition}
In line with previous work \cite{andres2011probabilistic, keuper2015efficient}, the MP can be understood as a Bayesian Network, where the optimal partitioning $\mathscr{Y}$ depends on the individual edge decisions $y_e \in \{0,1\}$.
They are dependent on the image-pair features $x_e \in \mathbb{R}^n$ for all $e \in E$ of the graph G, as shown in \cref{fig:MP_bayesiannetwork}.

Given appropriately set edge costs 
$w_e=
\mathrm{log}
\left(
\frac{1-p(y_e\mid x_e)}{p(y_e\mid x_e)}
\right)$, 
solving the MP is equivalent to maximizing the posterior probability $p_{y\mid x,\mathscr{Y}}$ with
\begin{equation}
    p(y\mid x,\mathscr{Y})  \propto p(\mathscr{Y}\mid y) \cdot p(x, y) 
\end{equation}
which can be rewritten as
\begin{equation}
    %p_(y\mid x,\mathscr{Y})  \propto p(\mathscr{Y}\mid y) \cdot \prod_{e \in E} p(y_e\mid x_e) .
    p_(y\mid x,\mathscr{Y})  \propto p(\mathscr{Y}\mid y) \cdot p(x \mid y) \cdot p(y) .    
\label{eq:probMP}
\end{equation}
This proportionality holds under the assumption that $x$ and $\mathscr{Y}$ are conditionally independent. %, $x \independent \mathscr{Y}$.
The right-hand side of \cref{eq:probMP} contains three parts, the likelihood of a clustering $p(\mathscr{Y}\mid y)$ which is set to zero, if $y$ differs from the optimal clustering, and to a constant otherwise, as we do not have any prior knowledge about the clustering. 
The second part is the likelihood of the image similarity feature $p(x\mid y)$ and the third part is the bias term $p(y)$.
By choosing strong embedding models, we assume all $p(x\mid y)$ are high for their respective $\mathscr{Y}$ and that different embedding spaces allow to detect different frames.
We compare the visual frames detected using MP clustering to centroid-based k-means, density-based DBSCAN, and hierarchical agglomerative clustering using WARD linking. 
\input{Fig/bayesian_network}
\subsection{Image-graph mapping}
\label{sec:imagegraphmapping}
To formulate our clustering task as a MP, we map the images to nodes in a fully connected graph as Ho \etal suggests \cite{ho2021msm}.
In the graph, the edge weights represent the cosine similarity between image embeddings.
While the cosine similarity is defined for the range $-1 \leq c_s \leq 1$, our analysis in \cref{subsec:emb} shows that its distribution differs greatly between embedding models.
%In order to increase comparability between models, we define a normalization which we apply to all weights.
First, min-max scaling is used to confine the weights to the range $[0,1]$.
Then, the weights are transformed such that the decision boundary for cutting an edge is at zero, with positive weights corresponding to the likelihood of images belonging to the same cluster and negative weights to different ones.
Na\"ively, the decision boundary is at the transformed, normalized cosine similarity of 0.5, using 
\begin{equation}
   w_{ab} = \log \frac{1-p(y_{ab}\mid x_{ab})}{p(y_{ab}\mid x_{ab})} \propto \log \frac{s_c(a,b)}{1 - s_c(a,b)}
\end{equation}
where $a$,$b$ are nodes in the graph, corresponding to images, and $s_c(\cdot,\cdot)$ is their cosine similarity.
Depending on the embedding space, the inherent decision boundary can be located at different positions. 
To assign appropriate pseudo-probabilities and account for different calibration of the cosine similarity w.r.t. probabilities, we ablate a calibration term $\mathit{cal}$ for each embedding space and set  
\begin{equation}
   w_{ab} =  \log \frac{s_c(a,b)}{1 - s_c(a,b)} +  \log\frac{1- \mathit{cal}}{\mathit{cal}}.
\end{equation}
%We aim to determine $\mathit{cal}$ independent of any edge features corresponding to unbiased pseudo-probabilities, i.e. $p(y)=0.5$. 
We ablate for $0.1 \leq \mathit{cal} \leq 0.9$ in \cref{subsec:graph_abl} on independent validation and test sets.
To this end, we use two annotated datasets and compare the clusterings to the dataset's classes.
%Our design of using the cosine similarities of high quality embeddings ensures a high quality likelihood approximation of image features. 
%We will further discuss this in \cref{subsec:perf}.

\subsection{Solvers}
Efficient heuristics \cite{keuper2015efficient} are used to solve the MP \ie~finding image clusters.
We employ the Greedy Additive Edge Contraction [GAEC] and the Kerninghan-Lin [KL] algorithm \cite{kernighan1970efficient} to efficiently cut the graph into distinct components in the implementation of Keuper \etal \cite{keuper2015efficient} in \cite{graph_mcmc}. 
More algorithm details can be found in the documentation.
\begin{algorithm}
  \caption{Greedy Additive Edge Contraction}
  \label{alg:GAEC}
\begin{algorithmic}
\Require $G=(V,E)$, Edge Weights $w_e \ \forall e \in E$
\Ensure Final set of clusters $C$ and total cost
\State Initialize $C$ with each vertex in its own cluster
\State Initialize total cost = $0$
\State Create a priority queue $Q$ to store edges $(u, v)$ sorted by their weights in descending order
\While{$Q$ is not empty \textbf{and} highest edge weight $\geq 0$}
    \State Extract edge $(u, v)$ with the highest weight from $Q$
    \State Merge the clusters containing $u$ and $v$
    \State Update the cluster set $C$ accordingly
    \For{each edge $(x, y)$ adjacent to $u$ or $v$}
        \State Update the weight of the edge $(x, y)$ if needed
        \State Reinsert updated $(x, y)$ into $Q$
    \EndFor
    \State Update total cost += weight of merged edge $(u, v)$
\EndWhile
\State \Return the final set of clusters $C$ and total cost
\end{algorithmic}  
\end{algorithm}

First, we use GAEC, described in \cref{alg:GAEC}, to compute a preliminary clustering.
This algorithm starts with each image in a separate cluster and iteratively merges the two clusters connected by the largest, positive edge weight. 
The graph size is continuously reduced during the execution and the algorithm stops when merging any two additional clusters would have a negative weight.
The resulting, preliminary clustering is then optimized using the KL algorithm described in \cref{alg:KLj}, where three types of changes are possible, (1) exchange nodes of two neighbouring clusters, (2) move nodes to an new cluster, and (3) join two neighbouring clusters. 
The clustering refinement using the KL algorithm improves the clustering results, as shown in \ref{app:solver}.
Conceptually, the KL algorithm tries to improve an initial clustering by iteratively making small adjustments to the clustering and tracking their effect in terms of overall clustering cost.
The cost of a clustering consists of the the sum of all cut edges' weights and the algorithm aims to minimize it.

\begin{algorithm}
  \caption{Kernighan-Lin Algorithm with Joins (KLj)}
  \label{alg:KLj}
\begin{algorithmic}
\Require Graph $G = (V, E)$, Edge Weights $w_e \forall e \in E$
\Ensure Partitioning $P$ with $|P| > 1$
\Procedure{External Cost}{$a, P$}
    \State \Return $\sum_{v \in P \setminus \{P_a\}} w(a,v)$
\EndProcedure
\Procedure{Internal Cost}{$a, P$}
    \State \Return $\sum_{v \in P_a} w(a,v)$
\EndProcedure
\State Compute initial D-values for all $v \in V$ where $D(v) = \textsc{ExternalCost}(v, P) - \textsc{InternalCost}(v, P)$
\Repeat
    \For {each edge $e = (a, b) \in E$} 
       \If {node\_changed($a$) \textbf{or} node\_changed($b$)}  
            \State Update $D(a)$ and $D(b)$
            \State Update partition $P'$
        \EndIf
    \EndFor
    \For {each node $a \in V$} 
        \If {node\_changed($a$)} 
            \State Update $D(a)$
            \State Update partition $P'$
        \EndIf
    \EndFor
\Until{no further changes in $P'$}
\State \Return the optimized partitioning $P'$
\end{algorithmic}  
\end{algorithm}

\subsection{Metrics and Evaluation}
To measure the differences between clusterings, we use the variation of information [VI] proposed by Meil\u{a} \cite{meilua2003comparing}
\begin{equation}
    VI(C, C')=H(C\mid C')+H(C'\mid C),
\label{eq:vi}
\end{equation}
where $VI$ is the sum of the two conditional entropies of the two clusterings.
Each clustering result may consist of 1 to x clusters.
The $VI = 0 \iff C = C'$ has the upper bounded $VI(C, C') \leq \log n$ where n is the number of nodes in the graph \ie the number of images.
It is a true metric and the triangle inequality holds.
We ablate the calibration term by investigating the two conditional entropies, comp. \cref{eq:condh} on an independent validation and test set.
\begin{equation}
\label{eq:condh}
    H(Y\mid X) = -\sum_{x\in X} \sum_{y\in Y}
p(x, y)\log p(y\mid x)
\end{equation}
% The comparability between clusterings is ensured by mapping the cluster labels to the class label that they are most similar.
% This mapping assigns the class label to the largest cluster containing its images. 
% All clusters beyond the number of classes are assigned a new number.
% When comparing clusterings of two different embedding models, we employ the Hungarian matching algorithm  \cite{kuhn1955hungarian} with a cost functions containing the top 3 most similar clusters in descending order.
% Given that the cluster overlaps with less than three clusters of the other model, all missing costs are set to zero.
We use the implementation of \cite{vi_git}, which iterates over all combinations of clusters in the two clusterings.

Additionally, we use standard cluster statistics to compare the clusterings in terms of cluster sizes and diversity.
%During We compare the clustering to the dataset's class labels, if available, and to each other using \cref{eq:vi}.
For ClimateTV, we manually investigate the largest clusters by randomly selecting 10 images per cluster.
The differences between embedding models are further assessed in terms for common clusters and clustering differences. 
We extend this by investigating which clusters are entirely contained within another dataset's larger cluster.

\subsection{Datasets}
\begin{CJK*}{UTF8}{gbsn}
%\Image网
We employ two curated datasets, ImageNette \cite{Howard_Imagenette_2019} and ImageWoof \cite{Howard_Imagewoof_2019}, for determining the optimal $cal$ term for each embedding space.
ImageNette is selected to assess the efficacy for the distinction between broad concepts and ImageWoof for fine-grained concepts.
$Cal$ is selected based on the training set clustering and its effectiveness is validated using the validation set.
%are the curated image datasets, whereof ImageNette contains 10 classes which are deemed easy to classify, as they contain distinct classes, \ie tench, English springer, cassette player, chain saw, church, French horn, garbage truck, gas pump, golf ball, parachute.
%ImageWoof contains 10 dog classes which are harder to classify as they are more similar to each other. %, and Image网 is their combination.
In both cases we compare the MP clusterings to the dataset's classes.
%Given that we do unsupervised clustering, we combine the images from the train, validation, and test set.
%We evaluate the clusters in terms of their overlap to the dataset's classes.
Moreover, we use OmniBenchmark to measure how many realms and concepts the assessed embedding spaces can distinguish between.
It contains 21 realms which each consist of several concepts, in total 7,372 non-overlapping concepts are included and classification top-1 accuracy is currently below 50\%.
To this end, we randomly select 10k images from the train set and compare the clusters to the original labels, both on realm level and concept level.
Finally, we employ the ClimateTV dataset \cite{prasse2023towards} to exemplify the efficacy of our method in detecting visual frames.
This classification dataset contains animal classes, social media visuals of political protest, conferences, climate change solutions such as wind energy, and several climate change consequences, \eg floods, droughts, economic instability, and human rights infringements.
The authors have collected all images that were shared on X (formerly Twitter) in the year 2019 in the context of climate change. 
We focus our evaluation on the images tweeted in January 2019.
More details can be found in their work.
\end{CJK*}
\subsection{Experimental Setup}
We create image embeddings using ResNet50, VGG19-BN, ViT, ConvNeXt V2, CLIP ViT, and DINOv2 (details in \cref{app:models}), and build complete image graphs with weighted edges based on the image embedding's cosine similarity.
This graph is then divided into clusters using heuristic solvers for the MP.
The experiments were run on Intel Xeon CPU E5 and the image embeddings are created using NVIDIA GeForce RTX 4090.
The clusterings are compared with respect to their VI.
Additionally, we report further cluster statistics, \eg~cluster sizes, distribution, and cleanliness.
The comparison of quantitative and qualitative results concludes our investigation.

\label{subsec:graph_abl}
\begin{figure*}[ht]
    \centering
    \includegraphics[width=0.18\linewidth, trim={4.5cm 0 4.5cm 0},clip]{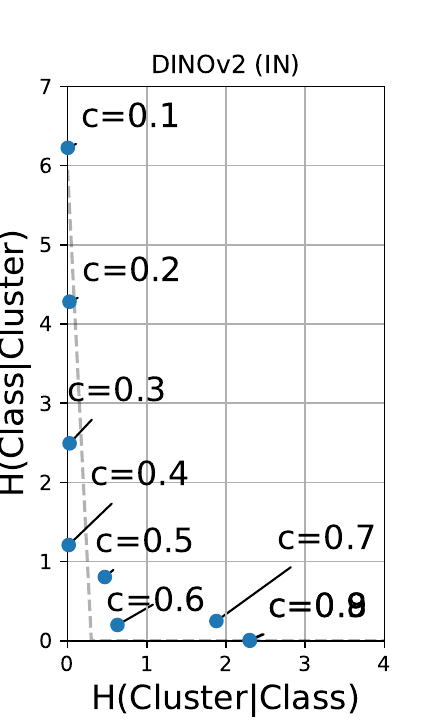}
    \includegraphics[width=0.18\linewidth, trim={4.5cm 0 4.5cm 0},clip]{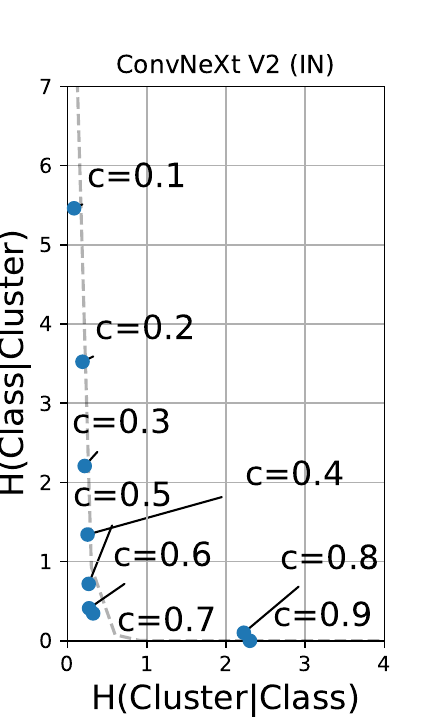}
    \includegraphics[width=0.18\linewidth, trim={4.5cm 0 4.5cm 0},clip]{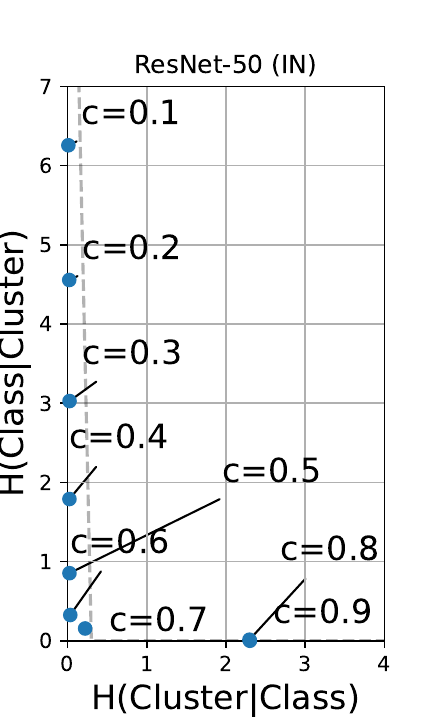}
    \includegraphics[width=0.18\linewidth, trim={4.5cm 0 4.5cm 0},clip]{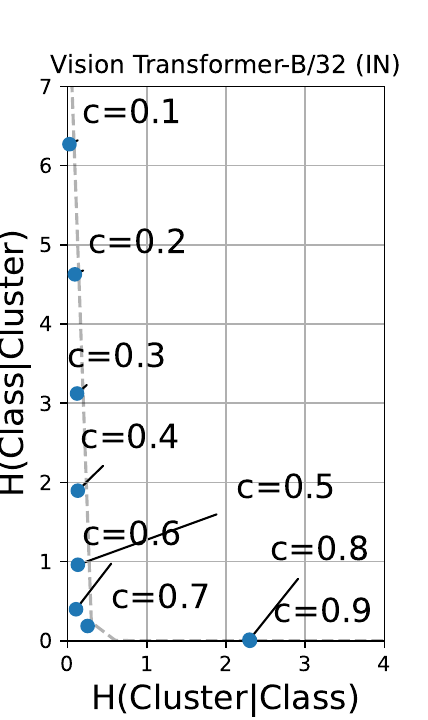}
    \includegraphics[width=0.18\linewidth, trim={4.5cm 0 4.5cm 0},clip]{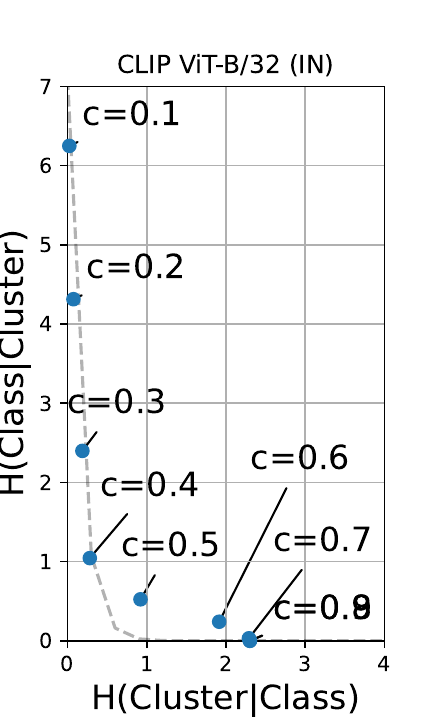}
    \caption{Calibration term [c] ablation across embedding spaces on ImageNette's train set shows that embedding spaces where the distance between different data points is increased during training require a smaller $cal$ compared to traditionally trained embedding spaces.}
    \label{fig:bias_abl}
\end{figure*}

\section{Results}
\label{sec:res}
We determine the calibration term $\mathit{cal}$ (defined in~\cref{sec:imagegraphmapping}) for each embedding space using the annotated ImageNette \cite{Howard_Imagenette_2019} and ImageWoof datasets \cite{Howard_Imagewoof_2019}.
Moreover, we report on the characteristics of the embedding spaces such as their data distributions and the overlap between embedding spaces using UMAP \cite{2018arXivUMAP} visualizations using the official implementation \cite{mcinnes2018umap-software}.
%An ablation of the most efficient graph construction in terms of edges can be found in the supplemental material \cref{app:edge_abl}.
Finally, we discuss clusters for the ClimateTV \cite{prasse2023towards} dataset to show how our method can support social science research.
We compare the cluster statistics and results to other clustering approaches.
This includes an excursion into multi-modality where we combine image and text input for the CLIP model.

\subsection{Calibaration Term Validation}
\label{ssec:cal}
We determine $\mathit{cal}$ by assessing the similarity of image classes and clusters.
The train set is used for experiments using  $0.1 \leq \mathit{cal} \leq 0.9$ with a step size of 0.1, whereof the best performing $cal$ is independently validated.
This is done for ImageNette, which has highly diverse classes, and for ImageWoof, which has fine-grained differences between classes.
%Our ablation assess the embeddings spaces efficacy in distinguishing between large- and small-grained differences.
%We ablate the graph quality in terms of finding the optimal clustering by selecting the adequate $\mathit{cal}$ term. % and graph efficiency in terms of run-time by removing unneeded edges before the MP is solved.
%The calibration term determines the spread of the data. 
%$cal=0.5$ equals the un-calibrated cosine similarities, where $w > 0$ are positive edge weights and $w < 0$ are negative ones.
Conceptually, the larger $\mathit{cal}$ is chosen, the more edge weights are larger than zero; reducing $\mathit{cal}$ has the opposite effect.
%This influences partitioning due to its effect on $p(y\mid x)$.
\cref{fig:bias_abl} shows the most optimal $\mathit{cal}$ term in terms of two conditional entropies, $H(Class \mid Cluster)$ and $H(Cluster\mid Class)$.
Given that the framework is probabilistic, we expect minor differences between runs. 
We select $cal$ such that $H(Class \mid Cluster)$ and $H(Cluster \mid Class)$ are balanced, as we aim for a high overlap between the clusters and the classes, \ie a low VI.
\begin{table}[ht]
    \centering
    \begin{tabular}{lcccc}
    \toprule
    Emb. model & $cal$ & $\Delta_{tr} H_1,H_2$  & $VI_{train}$ & $VI_{val}$ \\
    \midrule
    CLIP ViT-B-32 & $0.5$ &$0.40$& $1.55$ & $1.34$  \\
    DINOv2 & $0.6$ &$0.43$& $0.89$& $1.19$\\
    ConvNeXt V2 & $0.7$ &$0.02$ &$0.28$ & $0.42$ \\
    ViT-B-32 & $0.7$ &$0.07$& $0.46$ & $0.26$ \\
    ResNet-50 & $0.7$ &$0.07$& $0.44$ & $0.54$ \\
    Inc.-ResNetv2 & $0.5$ & $0.17$ & $0.49$  & $0.40$\\
    VGG19-BN & $0.7$ & $0.19$&$0.73$ & $0.94$ \\
    \bottomrule
    \end{tabular}
    \caption{Clustering ImageNette using ConvNeXt V2 closely fits the training set's classes. We use $\Delta H_1, H_2 = H(Class \mid Cluster) - H (Cluster \mid Class)$ as an indicator of clustering performance. ResNet-50 has the best validation VI.}
    \label{tab:in_clust}
    \vspace{-1em}
\end{table}
VI's differ slightly between the train and the validation set. 
Overall, CLIP ViT-B/32, DINOv2, and VGG19-BN have the worst clustering performance in terms of overlap with the original classes.
All other models perform well, with ResNet-50 achieving the highest VI overall, as \cref{tab:in_clust} shows.
The same trends can be observed for fine-grained differences, as the experiments on ImageWoof show (\cref{tab:iw_clust}). 
Again, CLIP ViT-B/32 achieves the worst VI and ResNet-50 the best.
However, in this setting, all embedding models' performances, except CLIP's, are more alike.
For the majority of embedding spaces, reducing $cal$ by a factor
0.1 improves the clustering similarity to the original classes.
We suggest selecting $cal$ based on our ablation with the application in mind.
If the data contains small differences, $cal$ can be reduced by 0.1.

We hypothesize that the slightly larger $cal$ aids the clustering of broad concepts, as it allows more images to be clustered together by GAEC. 
This algorithm does not join any negative edge, thus any two image representations connected by it, cannot initially be in the same cluster. 
The KL algorithm may alter the initial cluster assignment, however, the larger the negative cost, the less likely is this scenario.
Likewise, in the setting with fine-grained differences between classes, it appears optimal to have a lower number of initially positive edge weights to avoid clustering different classes together.
We expected that CLIP and DINOv2 would require lower $cal$ terms, as their training includes contrastive loss and the KoLeo regularizer respectively.

\begin{figure*}[ht!]
    \centering
    \begin{subfigure}[t]{0.3\linewidth}
        \includegraphics[width=\linewidth, trim={0 0 0 0.77cm},clip]{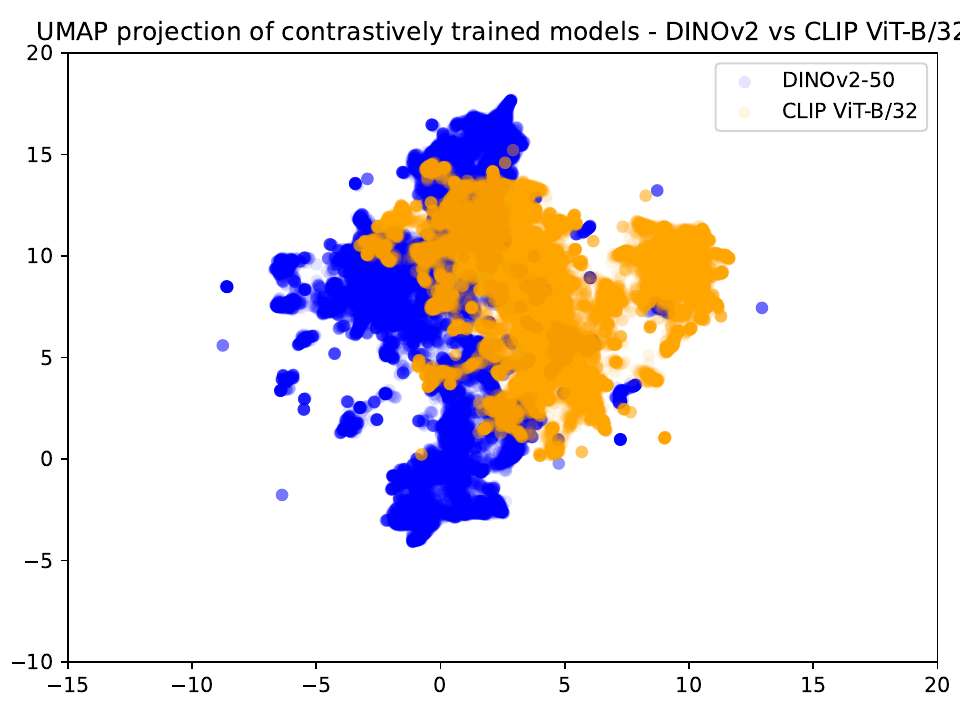}
        \caption{Contrastively trained spaces}
    \end{subfigure}
    \hfill
    \begin{subfigure}[t]{0.3\linewidth}
        \includegraphics[width=\linewidth, trim={0 0 0 0.77cm},clip]{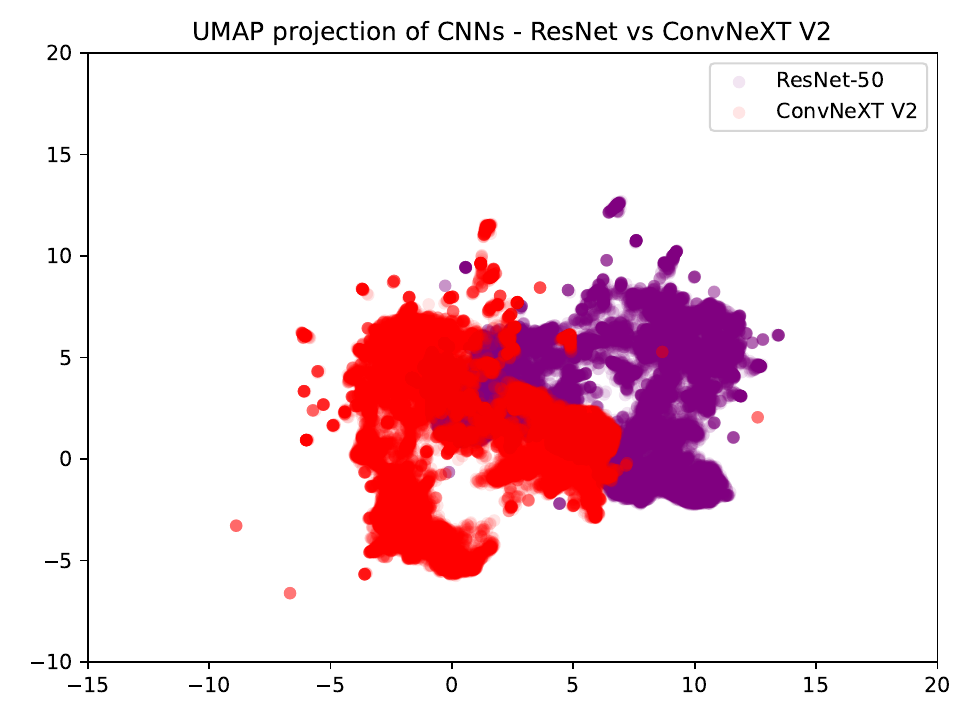}
        \caption{CNN spaces: ResNet vs. ConvNeXt V2}
    \end{subfigure}
    \hfill
    \begin{subfigure}[t]{0.3\linewidth}
        \includegraphics[width=\linewidth, trim={0 0 0 0.77cm},clip]{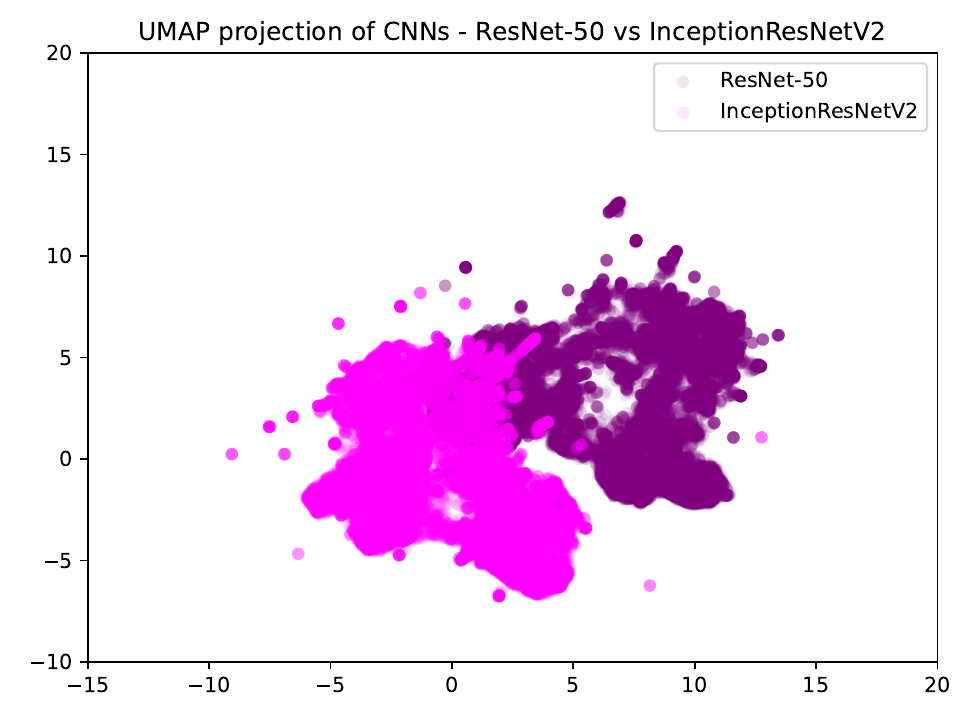}
        \caption{ResNet variants}
    \end{subfigure}
    \caption{UMAP visualization for image embeddings on ClimateTV reveals the differences in embedding space occupancy between different embedding models. DINOv2 and CLIP ViT-B/32 have a similar $s_c$ distribution but only a small overlap. The embedding space comparison of CNNs pre-trained on ImageNet1k shows almost no overlap.}
    \label{fig:umap}
\end{figure*}

\subsection{Embedding Space Analysis}
\label{subsec:emb}

\begin{table}[ht]
    \centering
    \begin{tabular}{lcccc}
    \toprule
    Emb. model & $cal$ & $\Delta_{tr} H_1,H_2$ & $VI_{train}$ & $VI_{val}$ \\
    \midrule
    CLIP ViT-B-32 & $0.4$ &$1.05$& $2.31$ &$2.26$ \\
    DINOv2 & $0.5$ &$0.40$& $1.45$ &$1.39$ \\
    ConvNeXt V2 & $0.6$ &\textbf{$0.01$} &\textbf{$0.61$} & $1.03$ \\
    ViT-B-32 & $0.6$ &$0.04$& $1.04$ &$1.10$ \\
    ResNet-50 & $0.7$ &$0.03$& $0.86$ & $0.73$ \\
    Inc.-ResNetv2 & $0.5$ & $0.04$ & $0.72$ & $0.97$ \\
    VGG19-BN & $0.6$ & $0.41$& $1.25$ & $1.13$ \\
    \bottomrule
    \end{tabular}
    \caption{Clustering ImageWoof using ResNet-50 closely fits the dataset's classes. $\Delta H_1, H_2 = H(Class \mid Cluster) - H (Cluster \mid Class)$ is an indicator of clustering performance.}
    \label{tab:iw_clust}
\end{table}
We analyse all image embeddings' un-normalized cosine similarities to asses the dataset's native distribution, as visualized in \cref{fig:cosinesim_IN}.
All cosine similarity distributions have a long-tail towards $c_s = 1$.
This is anticipated, as each image is only similar to the other $10\%$ of images which have the same class and are dissimilar to the remaining $90\%$ of the images.
The tail of the ViT and the ConvNeXt V2 model appear almost disjoint from the remaining distribution.
While some cosine similarity distributions appear more bell-shaped (DINOv2, CLIP RN50, CLIP ViT, Inception-ResNet-v2), others appear more skewed (ConvNeXt V2, RN50, ViT, VGG19). 
It appears that the more PDF resembles the normal distribution, the lower $cal$ term is optimal.
Models, which have less similar embeddings require a higher $cal$ to have a large enough number of positive edge weights.
Our findings also show \textit{narrow cone effect}~\cite{ModalityGap}, \ie embeddings having an above average cosine similarity and thus only cover a narrow cone in the embedding space hypersphere. 
%\textcolor{red}{First define and add reference for the narrow cone effect here!}
The narrow cone effect \cite{ModalityGap} can be observed in the CLIP (ViT \& RN50) embedding space and in the DINOv2 embeddings space, as their $\mu$ values are far beyond zero.
When the same architecture (ViT \& RN50) is pre-trained without contrastive loss, the narrow cone is not formed.
We have included the ResNet-50 CLIP encoder here, to show another instance of the narrow cone forming through multi-modal training. 
The narrow cone is apparent for some of the self-supervised models, but not for ConvNeXt V2.
Further causal investigations are beyond the scope of this work.

We find that the choice of embedding model has a large effect on the clusterings' VI, as no clear relation between $\mu$ and VI can be observed.
\cref{fig:umap} shows how little different embedding spaces overlap.
We can observe, that DINOv2 embeddings have the largest spread, while CNN embedding spaces appear more dense.

The models with a more Gaussian distribution are naturally less affected by an alteration of $\mathit{cal}$.
More details on the $s_c$ distributions can be found in \cref{apps:distri}.

Based on this ablation, we advocate for setting $\mathit{cal}$ according to the $s_c$ distribution of the embedding model.
When the inherent differences between embeddings are large, $\mathit{cal} = 0.7$ is required to create an impactful clustering.
Models whose embeddings are approximately normally distributed require no calibration, \ie $\mathit{cal} = 0.5$.
We find that the shape of the distribution is more influential than its mean value.
The granularity of the embedding space has a large effect on the final clustering.

\subsection{Clustering using MP}
\label{subsec:perf}
We compare the clusterings' statistics to better understand model differences to further investigate on image level.
For the annotated dataset, we compare the clustering to the classes.
Here, the number of clusters was larger than the number of classes (10).% in the original datasets.
This can be helpful in detecting outliers, \ie unlikely representations of the class, as for all classes clusters of size one exist.
Overall, the clusterings depend on the employed embedding model, as both statistics and image-level investigations show.
The clusterings are also subject to the diversity of the dataset, as VIs are generally larger for ImageWoof than for ImageNette.

This section highlights selected findings, additional results can be found in \cref{app:res}.

\begin{figure}[ht]
    \centering
    \includegraphics[width=0.3\linewidth, trim={0 4cm 0 0},clip]{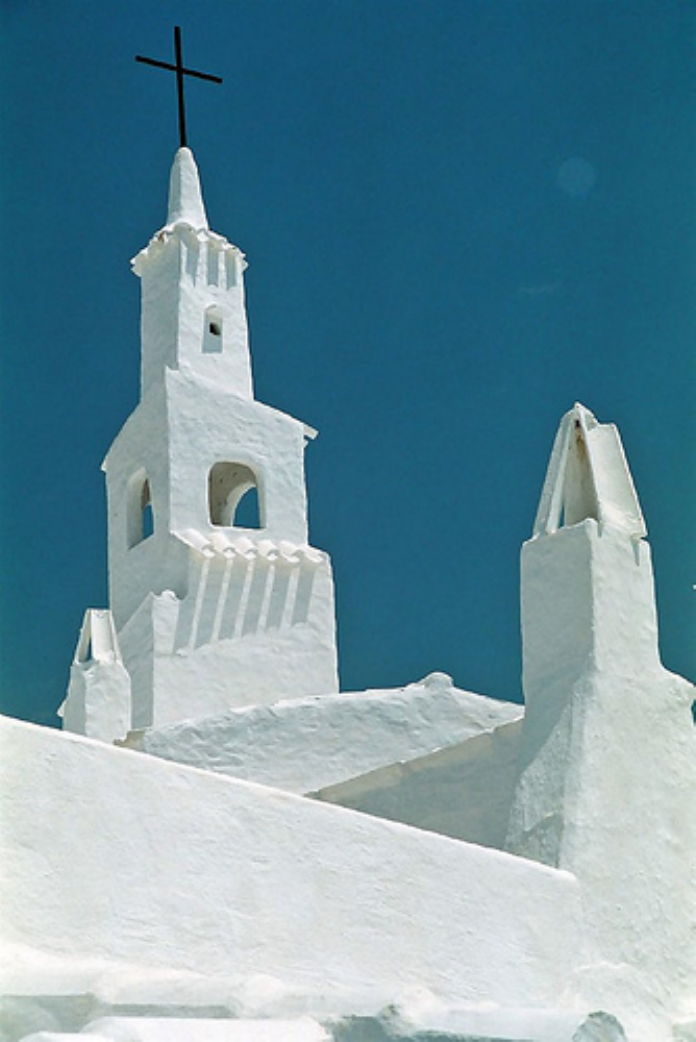}
    \hspace{0.1em}
    \includegraphics[width=0.51\linewidth]{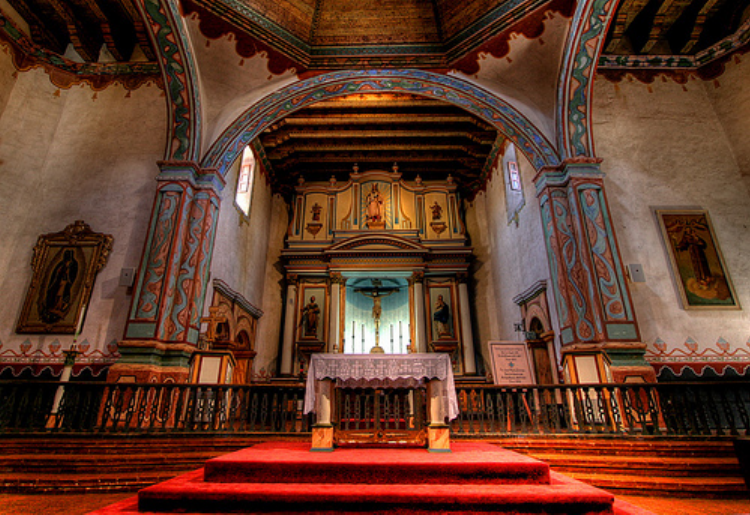} \\
    \vspace{0.1em}
    \includegraphics[width=0.51\linewidth]{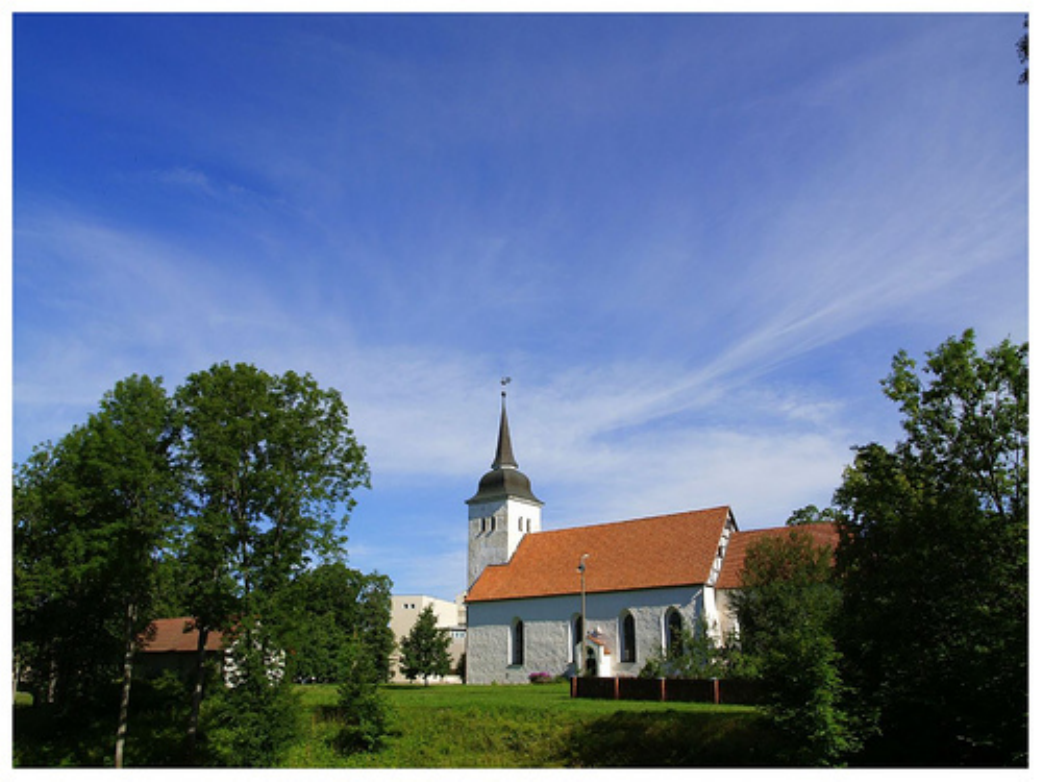}
    \hspace{0.1em}
    \includegraphics[width=0.3\linewidth, trim={0 0.6cm 0 0},clip]{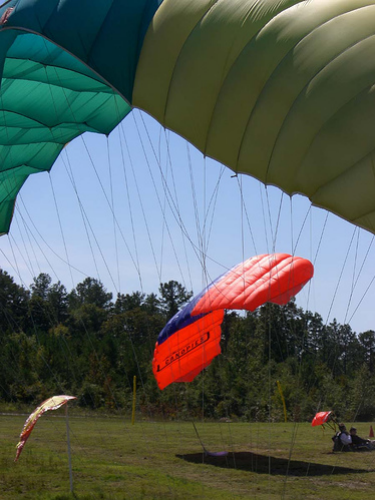}
    \\
    \label{fig:in_ex}
    \caption{The ImageNette church class is grouped into two single image clusters (r1), one fn as parachute and one fp parachute (r2).}
\end{figure}

\textbf{ImageNette} clusters confirm our expectations that unusual representations are separated into single image clusters (first row), while images that contain visual elements common for other classes might be mis-clustered (\cref{fig:in_ex}).
The image of the plain church, without any cross, which contains a beautiful sky is mixed with most of the parachute images, while the image with a parachute resembling a roof like structure is added into the large church cluster.
This stands in contrast to the clustering obtained using the CLIP $s_c$, where 41\% of clusters are mixed.
They contain up to 9 classes with at times equal contributions.

\begin{figure}[ht]
    \centering
    \includegraphics[width=0.42\linewidth, trim={0 2cm 0 0},clip]{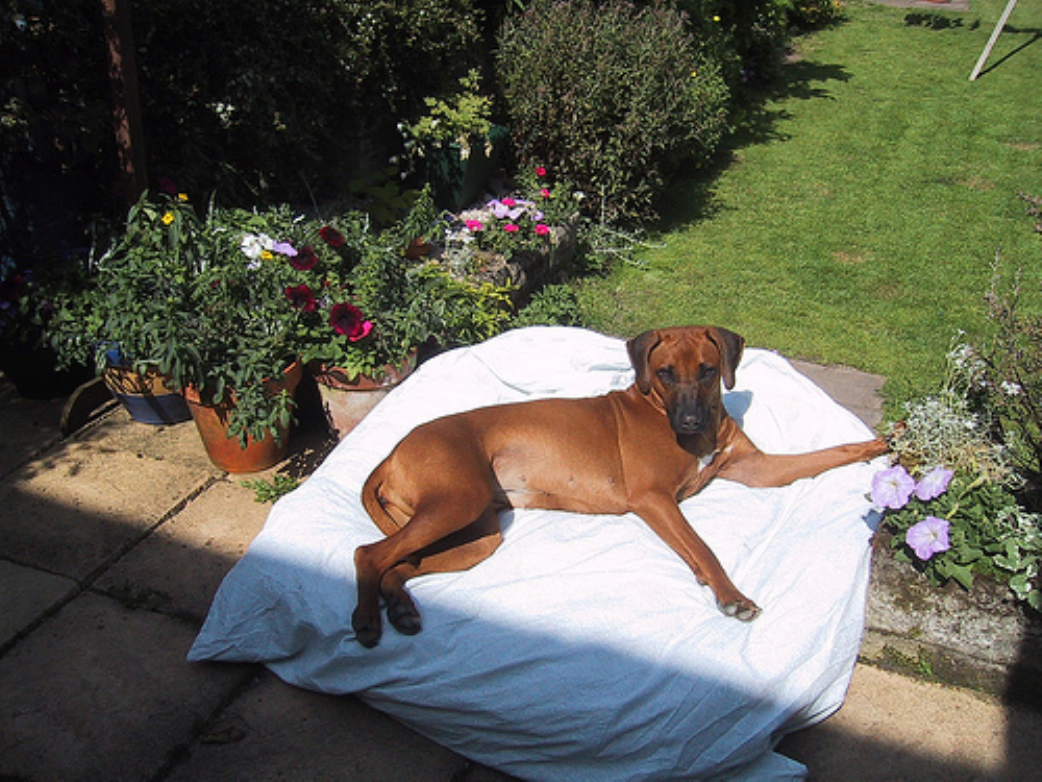}
    \hspace{0.1em}
    \includegraphics[width=0.42\linewidth, trim={0 3cm 0 0},clip]{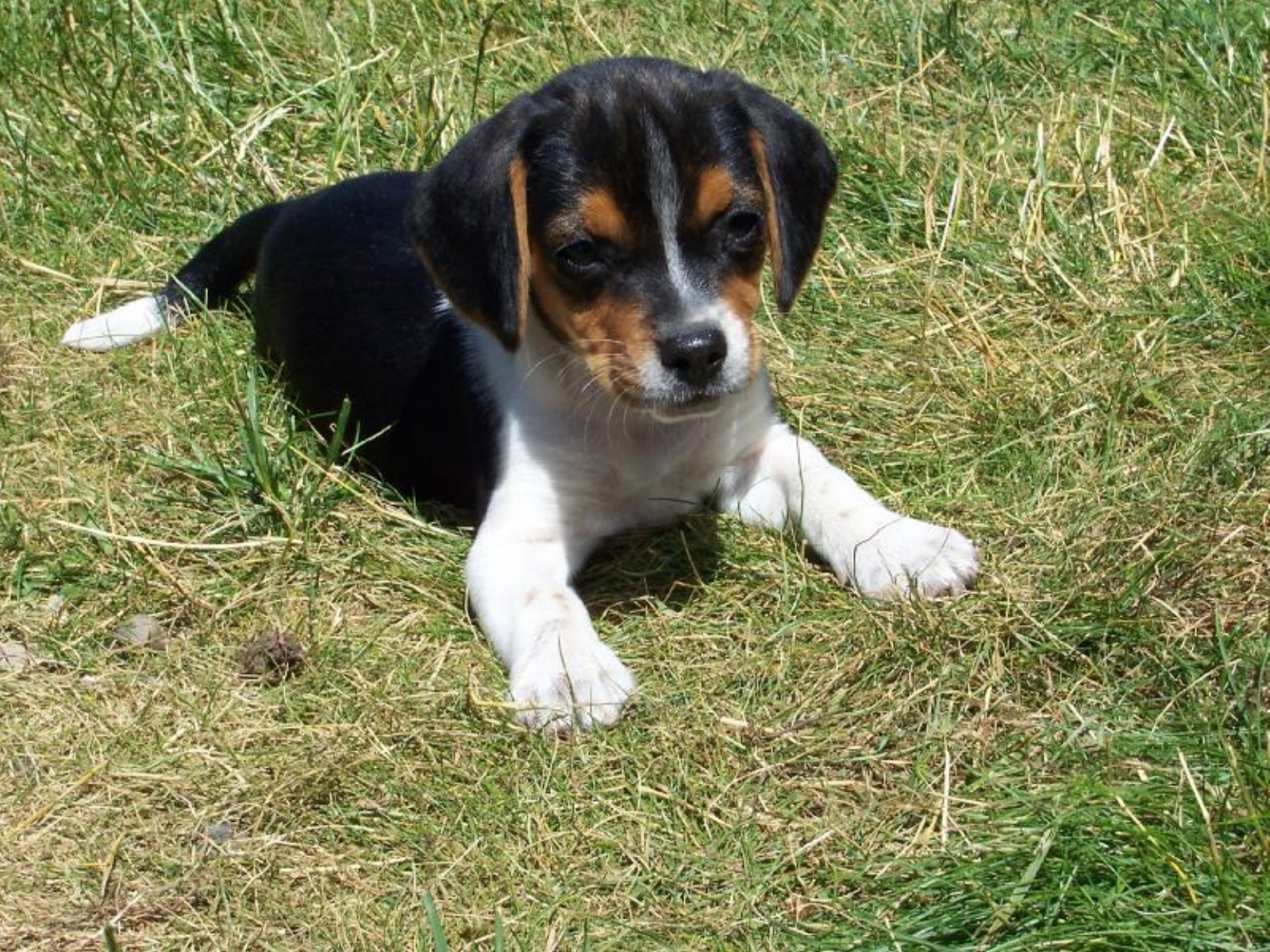} \\
    \vspace{0.3em}
    \includegraphics[width=0.42\linewidth]{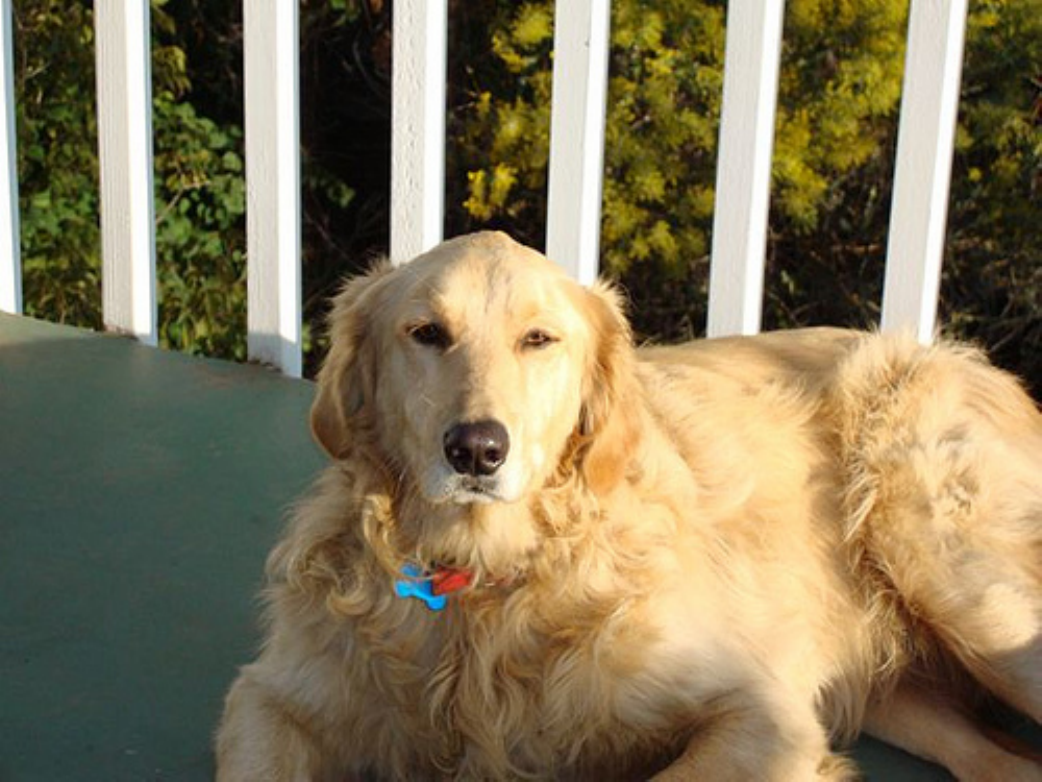}
    \hspace{0.1em}
    \includegraphics[width=0.42\linewidth, trim={0 4.5cm 0 0 },clip]{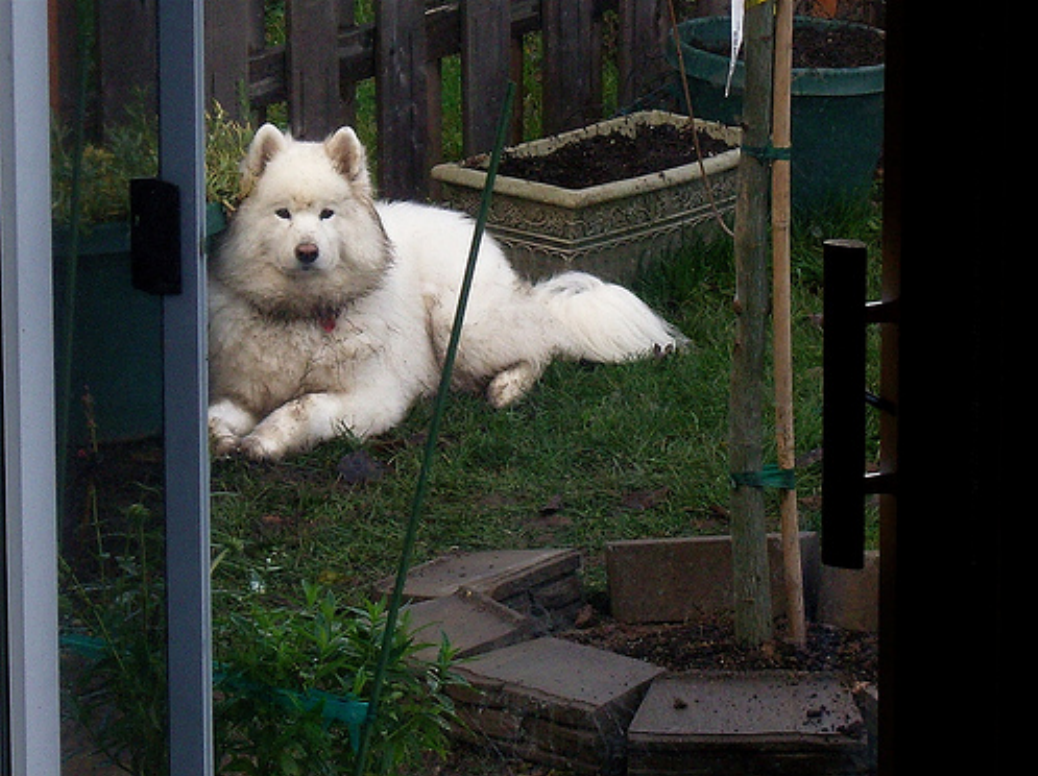}
    \caption{DINOv2 features appear to encode dog position and background, as r1 and r2 each show one cluster for ImageWoof.}
    \label{fig:iw_ex}
\end{figure}

\textbf{ImageWoof} clusterings contain fewer clusters for most models. 
Exclusively for DINOv2, more clusters are generated for ImageWoof as ImageNette.
These clusters contain many images of a single class in combination with a few outliers. %, or a small number of outliers.
While this results in a high VI compared to the class labels, the clusters appear meaningful, as they are based on common features shown in \cref{fig:iw_ex}

\textbf{ClimateTV image} clusterings are highly diverse, with $VI=3.99$ between ConvNeXt V2 and DINOv2 clusterings.
The number of clusters returned is more 2x higher for ConvNeXt V2 than for DINOv2.
While the largest cluster's size is comparable between models (approx. 11,5k), the cluster size distributions differ.
DINOv2 contains more large clusters in comparison, but also has a lower median value, due to more image clusters of size 1.
The largest cluster's contents differ greatly, as the DINOv2 cluster contains \textit{persons}, while the ConvNeXt V2 cluster contains \textit{computer generated content} \eg posters, visualizations, text.
The next larger ConvNeXt V2 clusters contains speakers (5k), outdoor photographs (5k), protest (1k), portraits (427), satellite images/earth visualizations. 
The largest agreement ($81\%$) between the two model's clusters is observed for frogs.
Both clusters contain few additional images without frogs.
Several ConvNeXt V2 clusters are contained in DINOv2 clusters, \ie frequently in \textit{persons}, but also in \textit{animals}.

\begin{figure}[ht]
    \centering
    \includegraphics[width=0.47\linewidth]{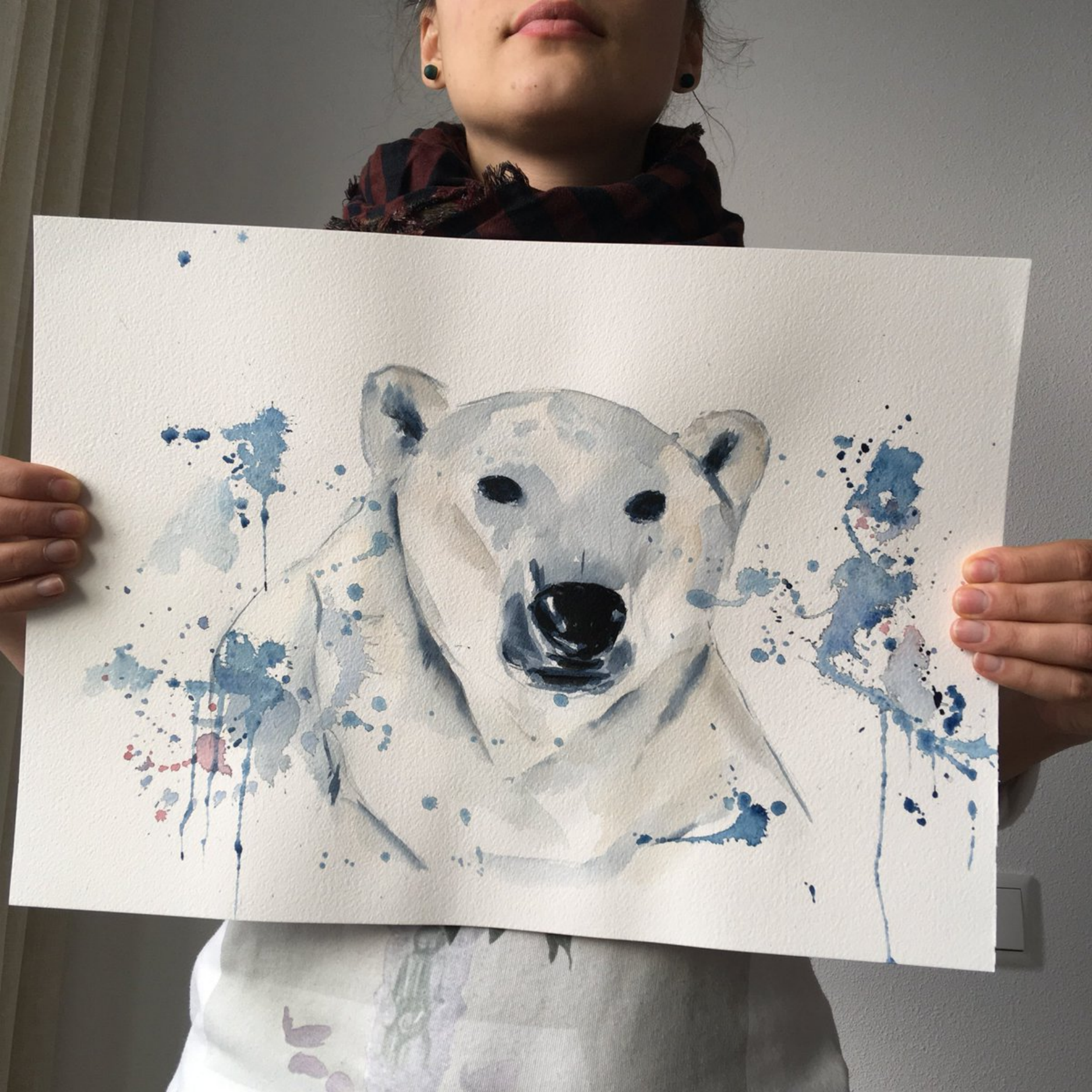}
    \hspace{0.1em}
     \includegraphics[width=0.47\linewidth]{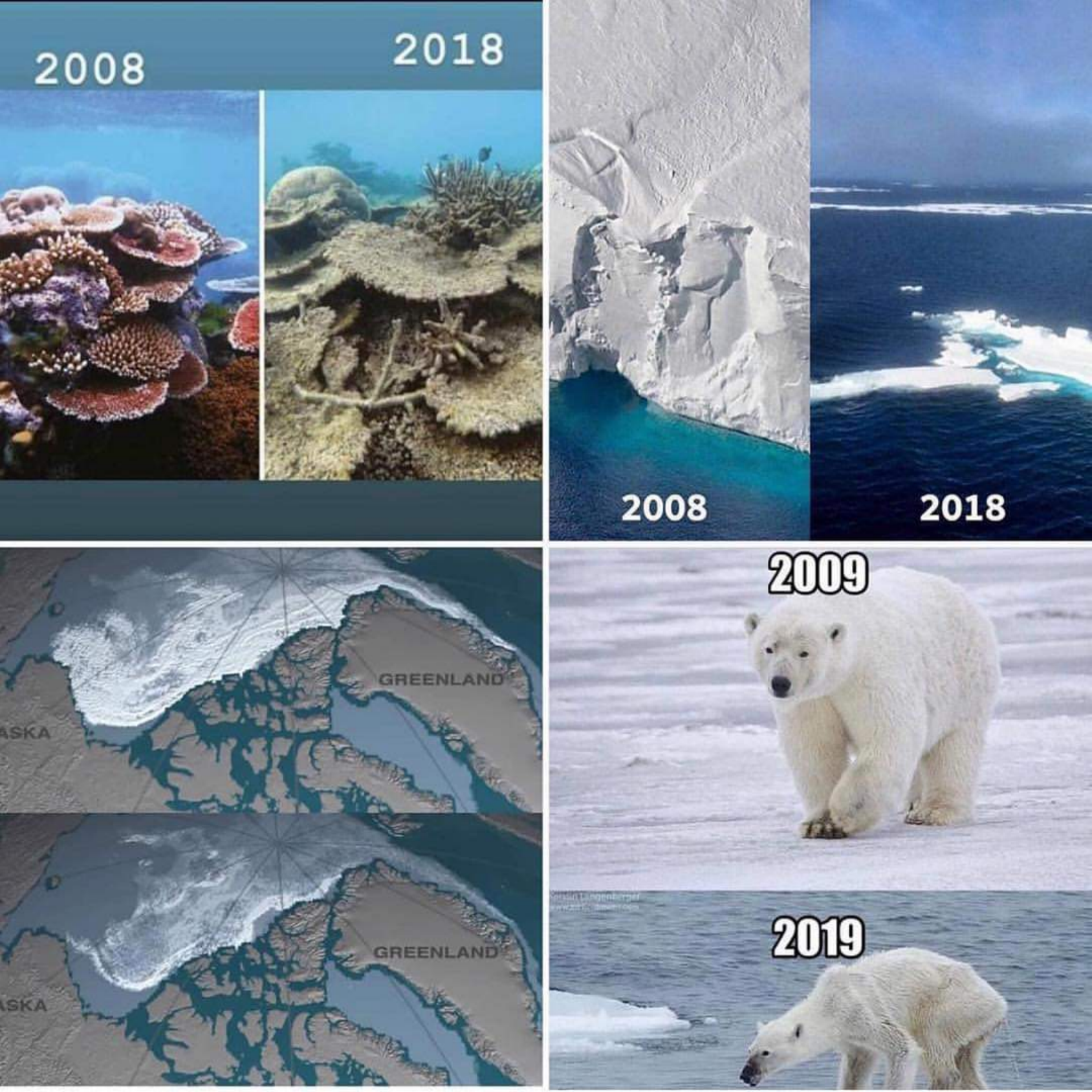} \\
     \vspace{0.1em}
    \includegraphics[width=0.47\linewidth]{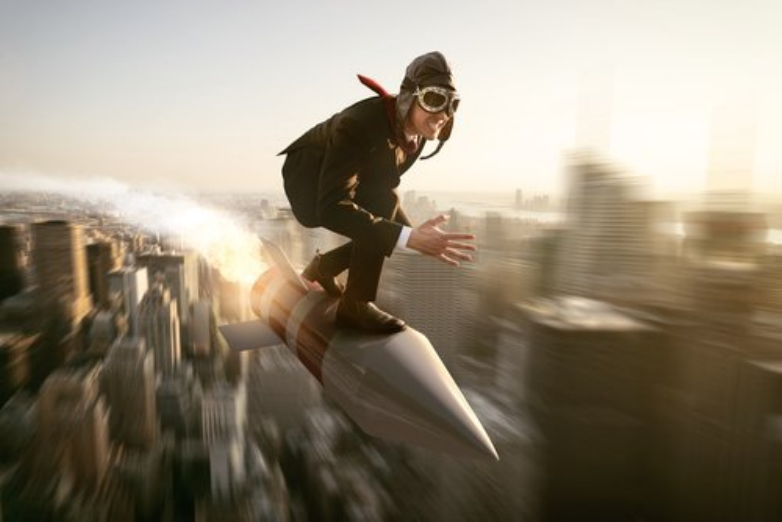}
    \hspace{0.1em}
    \includegraphics[width=0.47\linewidth]{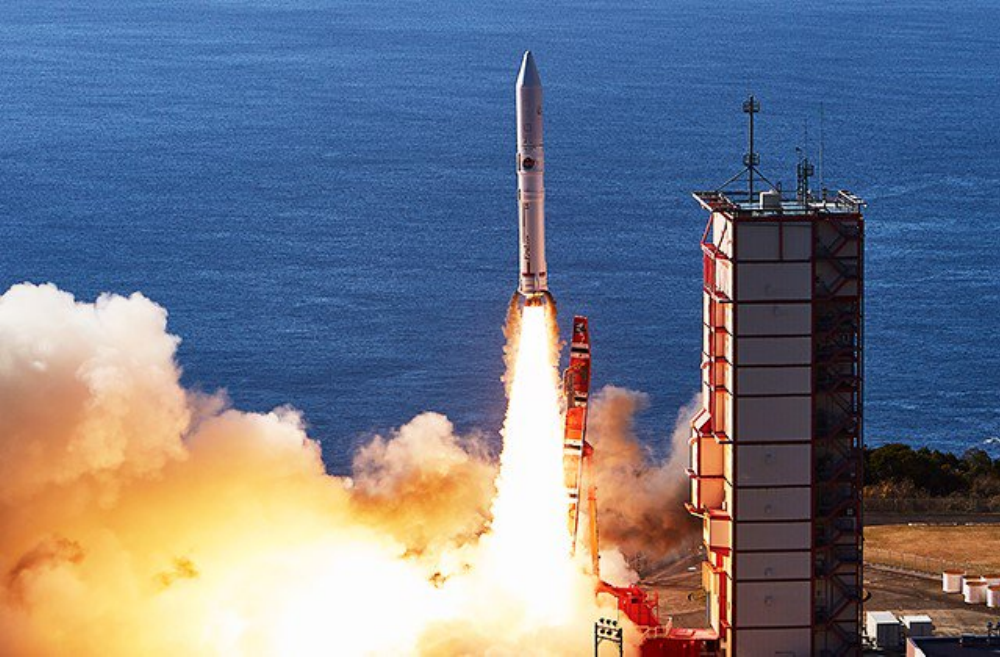} \\
    \vspace{0.1em}
    \includegraphics[width=0.47\linewidth]{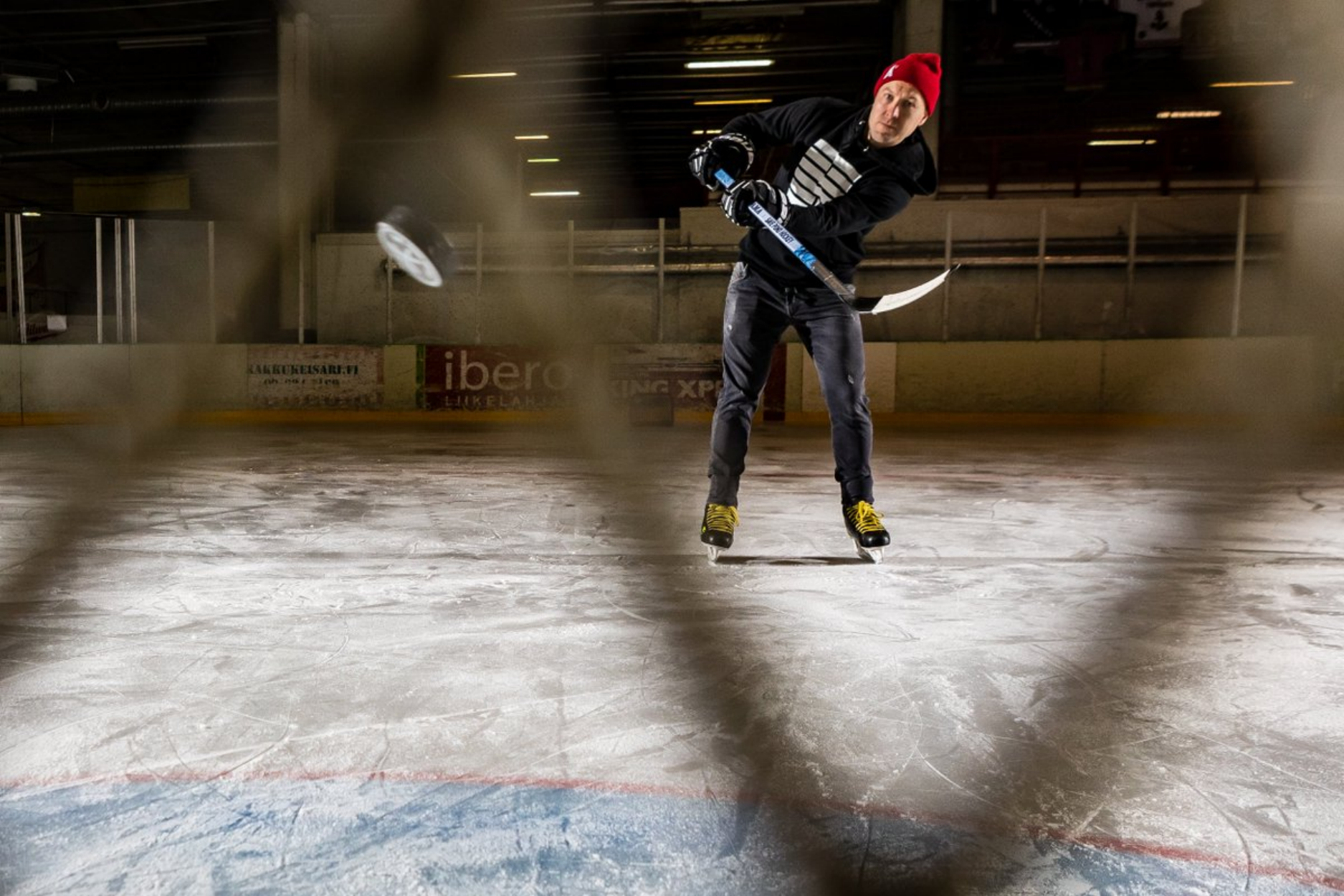}
    \hspace{0.1em}
    \includegraphics[width=0.47\linewidth]{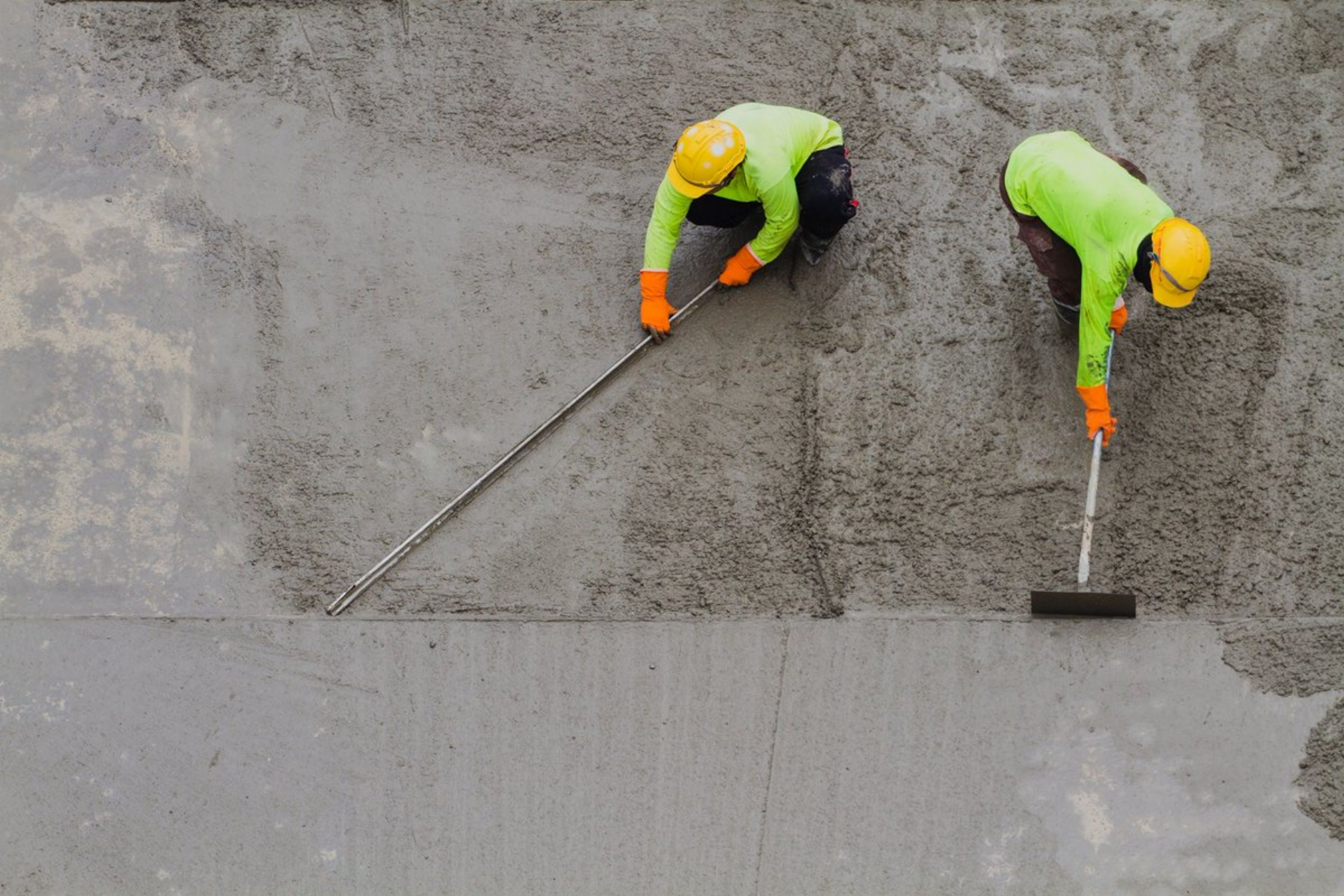} \\
    \caption{ConvNeXt V2 image features are clustered into content based topics, \eg polar bears (r1) and rockets (r2). Image background and common object can lead to mis-clustering (r3) where a hockey player and road construction workers form a cluster.}
    \label{fig:tv_ex}
\end{figure}

In line with previous experiments, the ConvNeXt V2 features result in meaningful clusters, examples shown in \cref{fig:tv_ex}, even for smaller cluster sizes.
When comparing the polar bear cluster between models, its images appear in 12 DINOv2 clusters.
All images containing persons (\cref{fig:tv_ex}, r1, left) are in \textit{persons}.
This includes both a person wearing a polar bear costume and a theater performance with stuffed polar bears, which we find remarkable.
Polar bear photographs in nature are in both cases clustered together.
However, for DINOv2, this cluster also contains images of other animals.
The combination of the two clusterings further allows us to identify noise.

\begin{figure}[ht]
    \centering
    \begin{minipage}[b]{0.47\linewidth}
        \includegraphics[width=\linewidth]{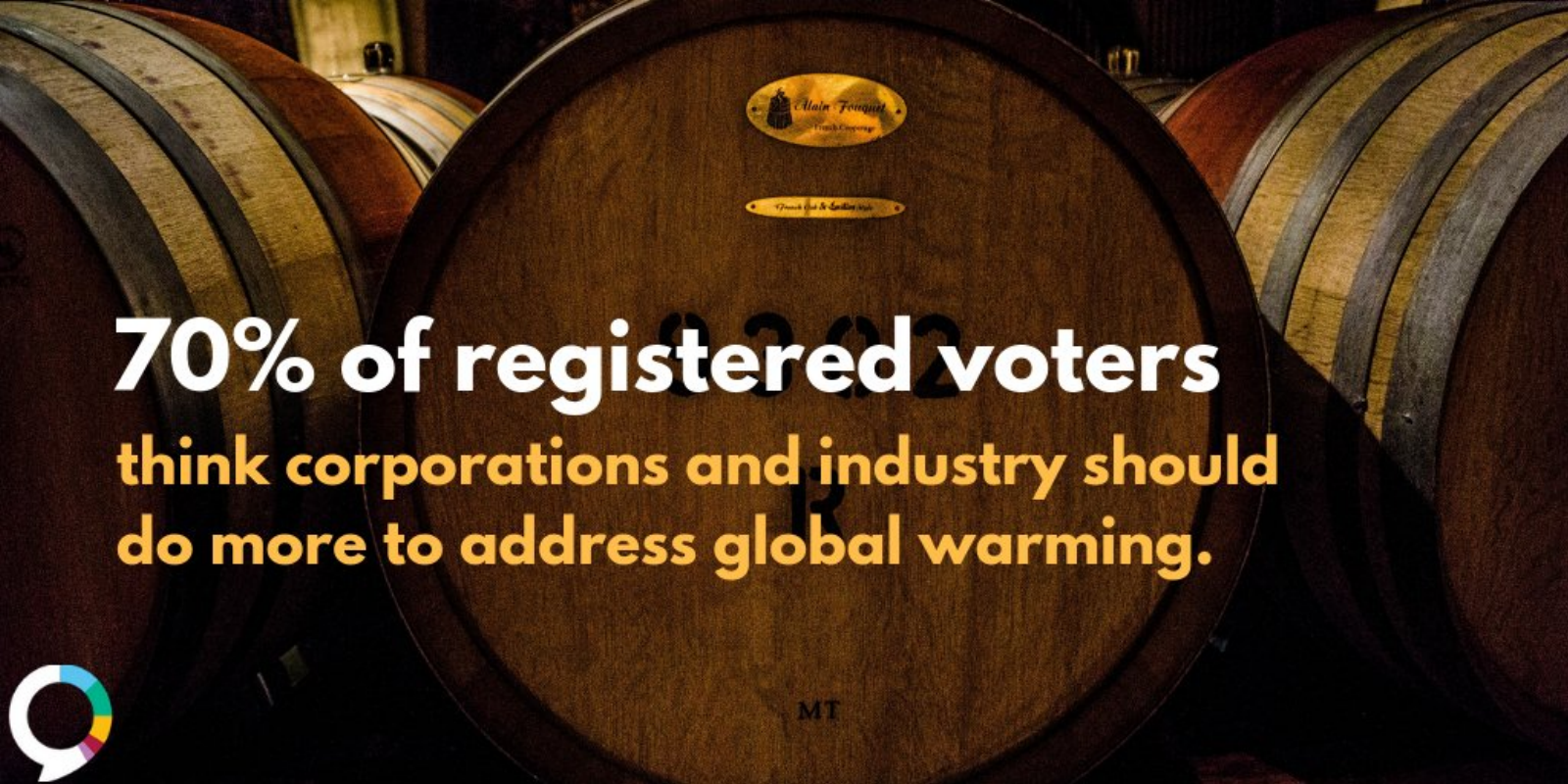}
        \vspace{0.5em}
        \includegraphics[width=\linewidth]{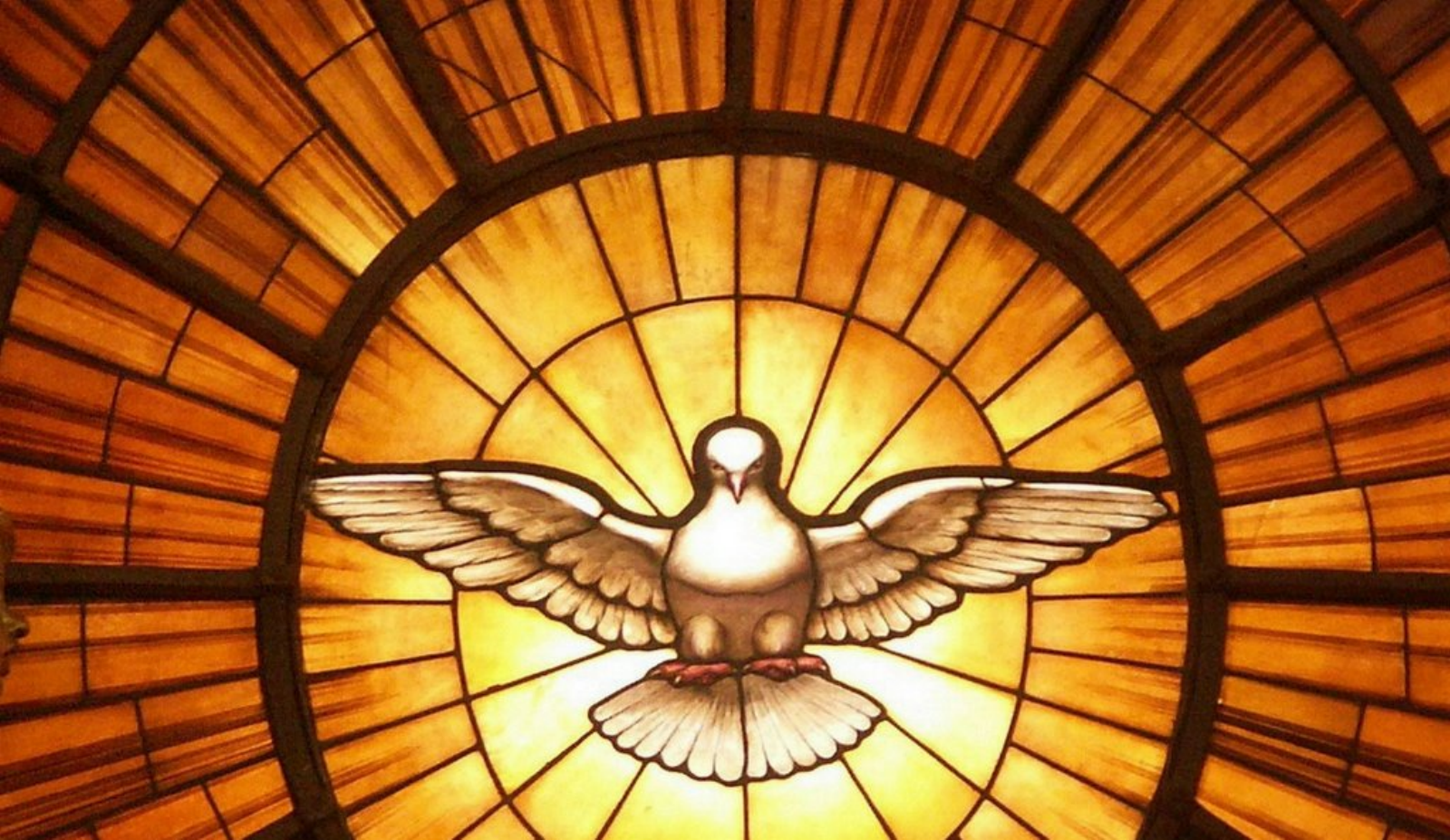}
    \end{minipage}
    \hspace{0.1em}
    \begin{minipage}[b]{0.47\linewidth}
        \includegraphics[width=\linewidth]{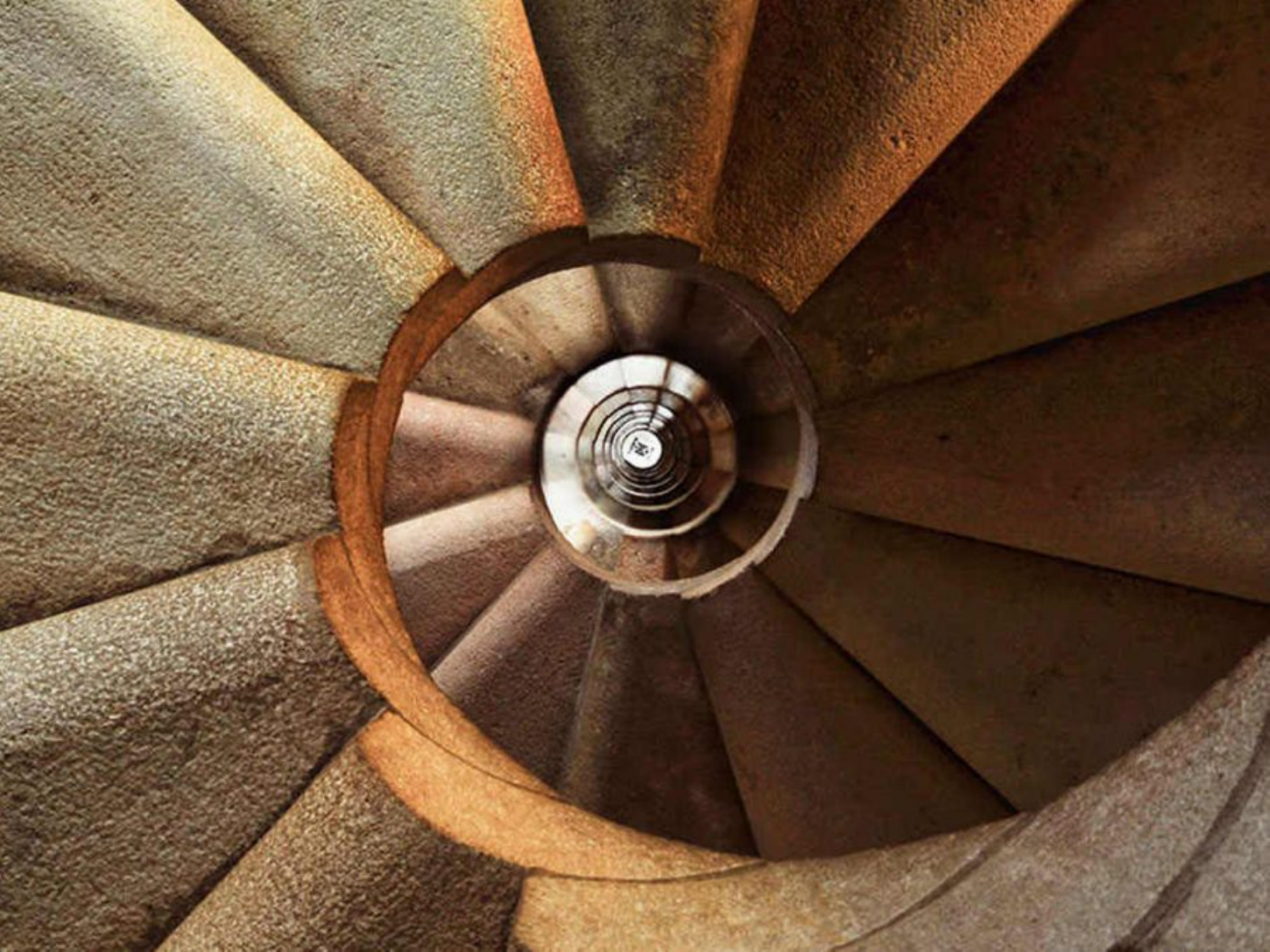}
        \vspace{0.5em}
        \includegraphics[width=\linewidth]{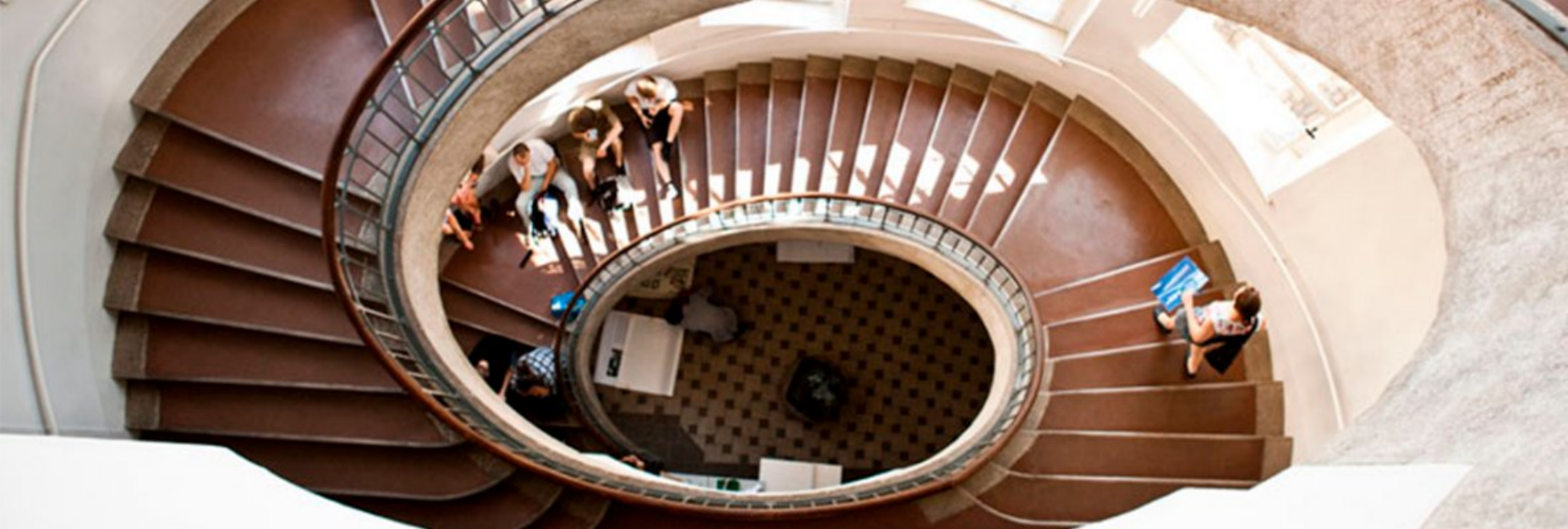}
    \end{minipage}
    \caption{While both models have a certain shape bias, DINOv2 strong tendency to clusters persons together offsets this.
    All images without humans form one DINOv2 cluster. The staircase images (r2, right \& r1, right) by ConvNeXt V2.}
    \label{fig:fail_ex}
\end{figure}

The small ConvNeXt V2 clusters of mixed images (\cref{fig:tv_ex}, r3) can again be found in the same DINOv2 cluster, however, here in combination with almost 12k other images containing humans.
Other large DINOv2 clusters contain event information or natural images, which indicates that DINOv2 clusters represent common patterns in the data.
\cref{fig:fail_ex} shows that both embeddings have a shape bias which results in small clusters of uncommon objects with similar shapes.
The DINOv2 cluster (\cref{fig:fail_ex}, r1 \& r2 left) is one example thereof, while ConvNeXt V2 clusters contain more variations of the curly shape (\cref{fig:fail_ex}, c1).

\textbf{Clustering Comparison}
When comparing our proposed MP clustering to KMEANS, DBSCAN, and agglomeration clustering, we can observe the same trends.
The CLIP embeddings' clusterings are the least similar to the other clusterings, with DINOv2's embedding space being to closest to it.
While the clusterings of KMEANS and agglomerative clustering are highly diverse, the results of DBSCAN all have inter-model VI's of less than 0.4.
The main advantage of MP clustering over KMEANS is that it can detect clusters of varying length, \eg~ DINOvs's people class would not be possible in this way.
Additionally, KMEANS clustering is less suitable for outlier detection as it - by design - only has few small clusters.

\textbf{Multimodal clustering}
We obtained the text corresponding to the ClimateTV images and used both as input to the MP.
The CLIP ViT-B/32 embedding space is already aligned and we use $cal=0.5$ as for the uni-modal model.
Due to the increased computational cost of using twice the nodes in the graph, we selected a random subset of 1,000 images to investigate the performance of the multi-modal MP.
We confirm prior findings of the modality gap \cite{ModalityGap} by having only image or text clusters.

\section{Discussion}
\label{sec:diss}
Image clusterings strongly depend on the image embedding model.
We observe a \textit{narrow cone} for both CLIP models, but also for the uni-modal DINOv2 and Inception-ResNet-v2 model.
Given that the clusterings using CLIP embeddings were most distinct when compared to the class labels, our results of it reducing embedding expressiveness are in line with previous works \cite{ModalityGap}.
The analysis of the ClimateTV dataset shows that DINOv2 features allow for a high-level understanding of the dataset by producing clusters such as \textit{person}, \textit{outdoor scene}, and \textit{nature}.
In spite of ResNet-50's strong performance on ImageNette and ImageWoof, its clustering on the abstract ClimateTV dataset contained a strong shape bias
The investigation of ImageNette clustering appeared to still contain too specific classes, so this effect was only identified when manually analysing the clusterings.
ConvNeXt V2 clusters however depend on fine-grained differences by \eg differentiating between \textit{speech} and \textit{protest}, which contain shared visual elements. 
The analysis of ImageWoof indicates that DINOv2 embeds the background and position of elements in the image, which can be helpful both in its own clustering or the validation of clusterings returned by another model.
The \textit{polar bear} cluster was divided and cleaned using the DINOv2 clustering.
In the context of climate change, we argue that the high-level DINOv2 and the fine-grained ConvNeXt V2 clusterings combined are a good starting point for frame analysis.

\section{Conclusion}
\label{sec:conc}
In this work, we propose a new method for social scientists to automate visual frame detection.
Our probabilistic clustering is phrased as a MP and uses image similarities of strong vision (and language) foundation models as a proxy for clustering probabilities.
In our experiments, we show the efficacy of MP clustering for detecting visual frames. 
We find that especially abstract frames such as \textit{speech} can only be detected by foundation models.
The intersection of clusterings can be used in order to reduce the number of noise in the clusters, given that the two clusterings are sufficiently distinct.
Our analyses have shown the potential usefulness of inter and intra embedding model multi-stage clustering, which we plan to investigate in future work. 
\section*{Acknowledgements}
This work is supported by the BMBF project 16DKWN027b Climate Visions. All experiments were run on University of Mannheim's server.

%%%%%%%%% REFERENCES
{\small
\bibliographystyle{ieee_fullname}
\bibliography{main}
}

\onecolumn
\appendix
\appendixhead
{\Large{--- Supplemental Material ---}}

\raggedright

\section{Embedding Models}
\label{app:models}

For all models, we use the recommended pre-training procedure without any modifications. 
The CLIP model is pre-trained on a web-scale dataset consisting of scraped image-text pairs. 
DINOv2 is pretrained on a combination of curated datasets.
All details can be found in \cref{tab:models}.
If part of the architecture, the classification head is replaced by torch.nn.Identity().

\begin{table}[ht]
    \centering
    \begin{tabular}{llll}
    \toprule
    Model & Architecture & Pre-training & Source \\
    \midrule
    CLIP RN-50 & ConvNet & openai & \small{\url{https://github.com/mlfoundations/open_clip}} \\
    \vspace{0.2em}
    CLIP ViT-B/32 & ViT & openai & \small{\url{https://github.com/openai/CLIP}} \\
    \vspace{0.2em}
    ConvNeXt V2 & ConvNet & ImageNet-1K & transformers.ConvNextV2ForImageClassification \\
    DINOv2 &ViT&custom&\small{\url{https://github.com/facebookresearch/dinov2}}\\
    ResNet-50 &ConvNet&IMAGENET1K\_V2& torchvision.models.resnet50 \\
    ViT-B/32 & ConvNet&IMAGENET1K\_V1&  torchvision.models.vit\_b\_32 \\
    VGG19-BN & ConvNet & IMAGENET1K\_V1 & torchvision.models.vgg19\_bn \\
    Inception-ResNet-V2 & ConvNet & IMAGENET1K\_V1 & timm \\
    \bottomrule
    \end{tabular}
    \caption{Details and sources for the employed embedding models.}
    \label{tab:models}
\end{table}
\newpage

\subsection{Distributions}
\label{apps:distri}
\begin{figure}[ht]
    \centering
    \includegraphics[width=0.3\linewidth]{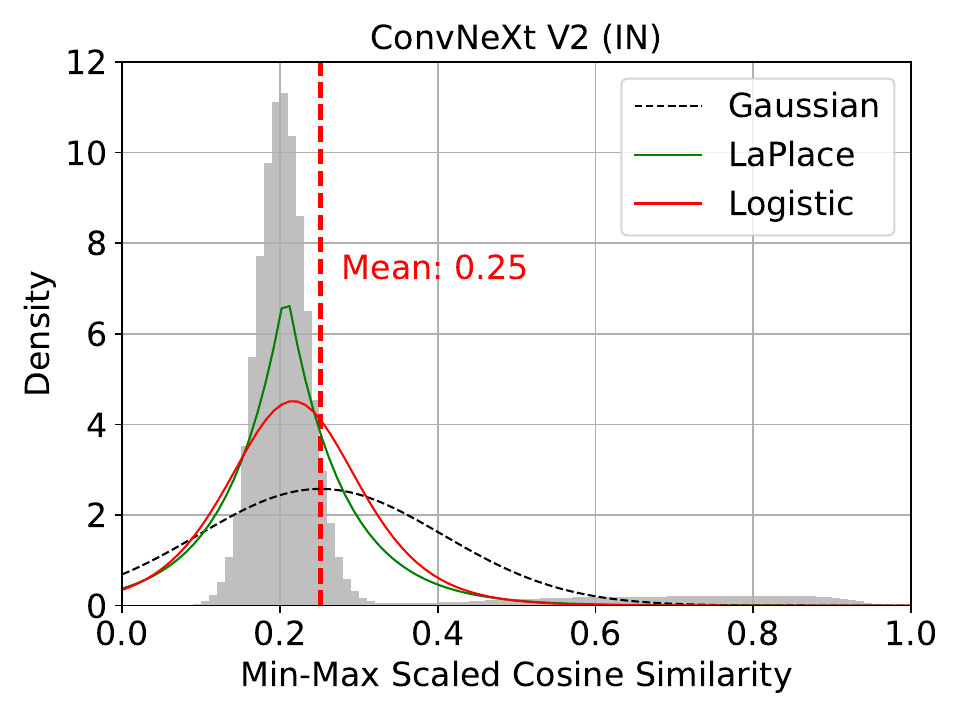}
    \includegraphics[width=0.3\linewidth]{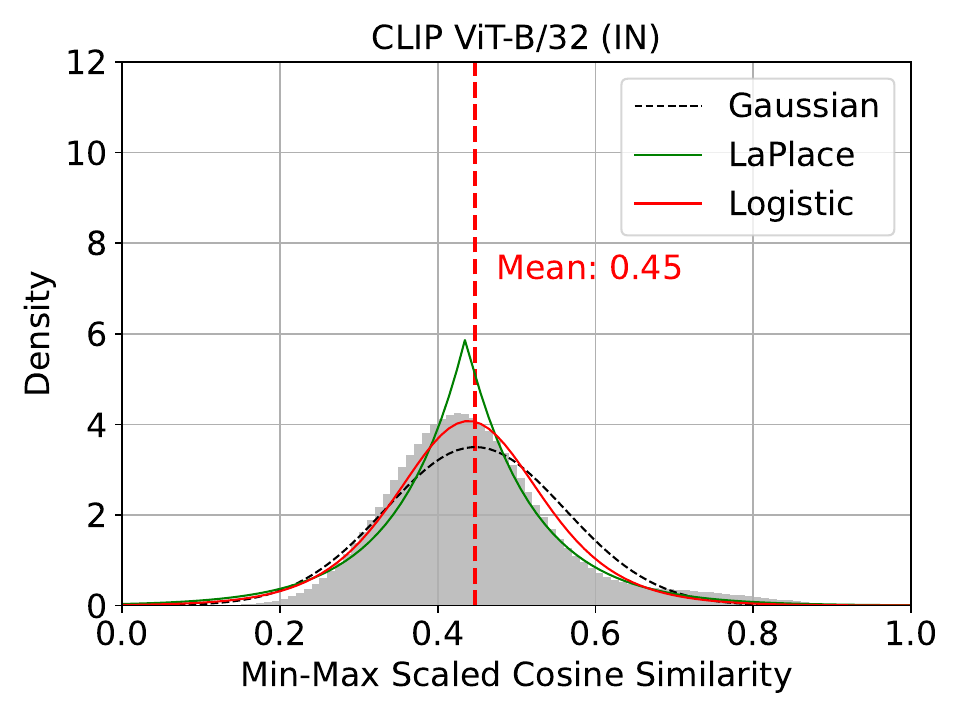}
    \includegraphics[width=0.3\linewidth]{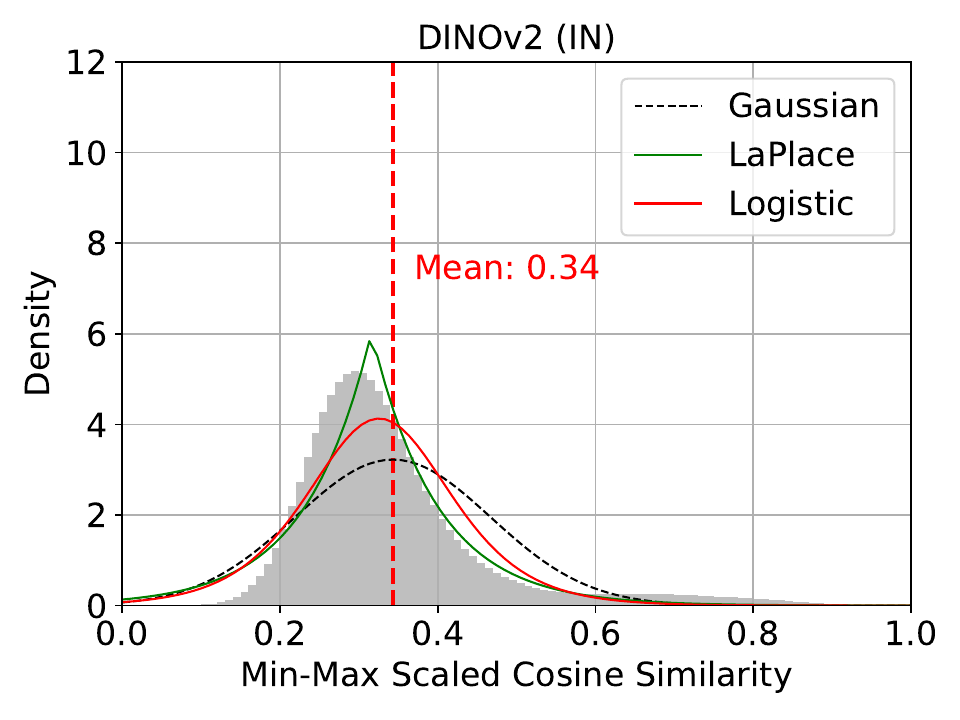}\\
    \includegraphics[width=0.3\linewidth]{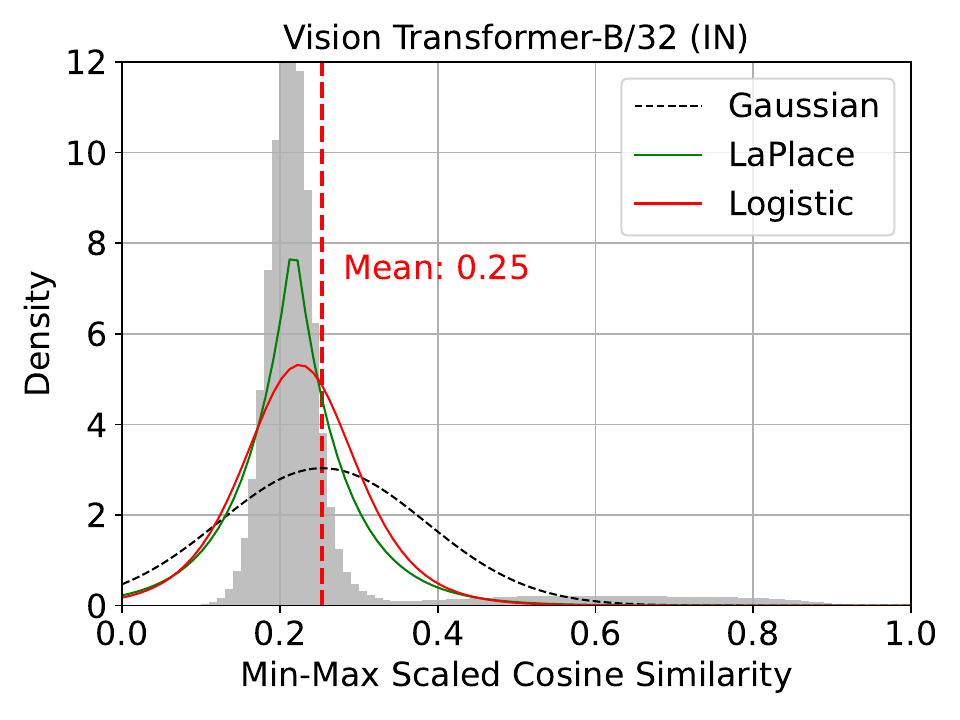}
    \includegraphics[width=0.3\linewidth]{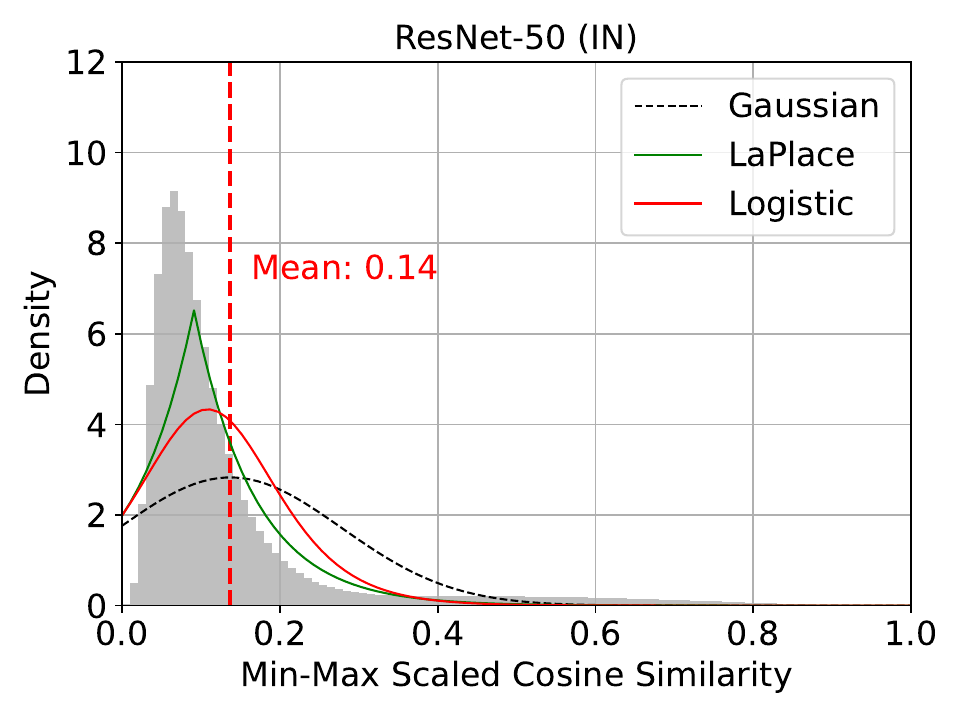}
    \caption{Distributions and mean for normalized
 ImageNette}
    \label{fig:distr_in}
\end{figure}

\begin{figure}[ht]
    \centering
    \includegraphics[width=0.3\linewidth]{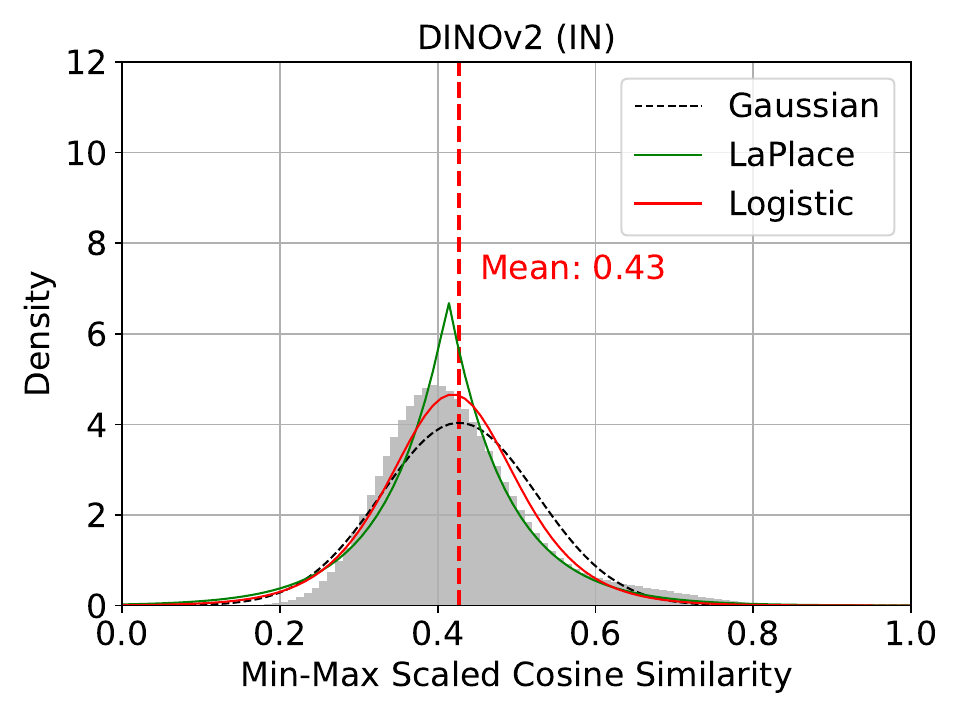}
    \includegraphics[width=0.3\linewidth]{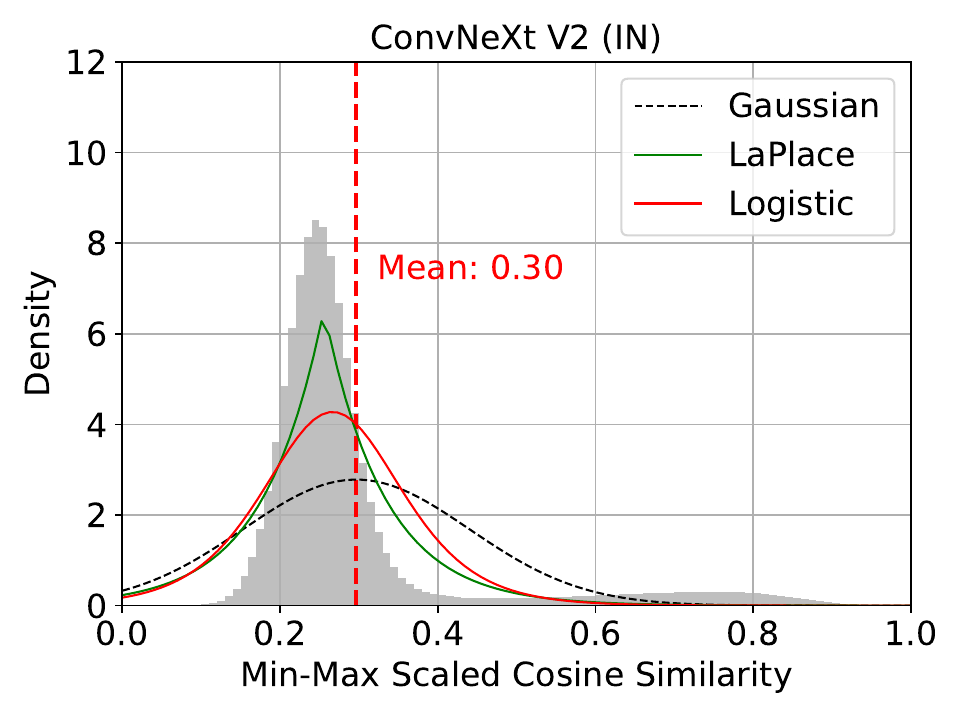}
    \includegraphics[width=0.3\linewidth]{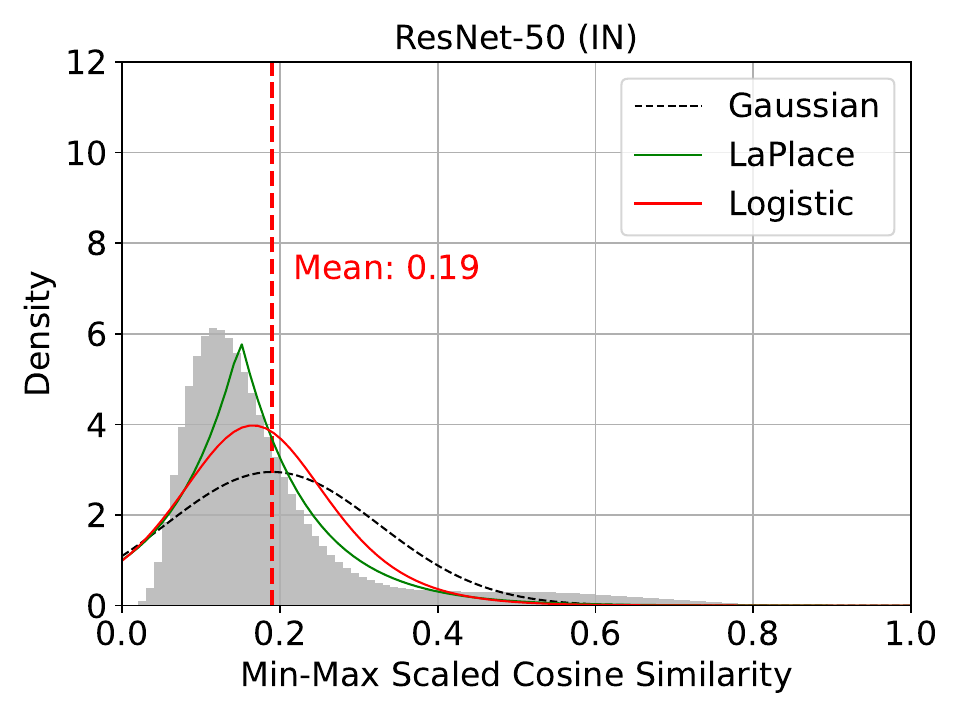}\\
    \includegraphics[width=0.3\linewidth]{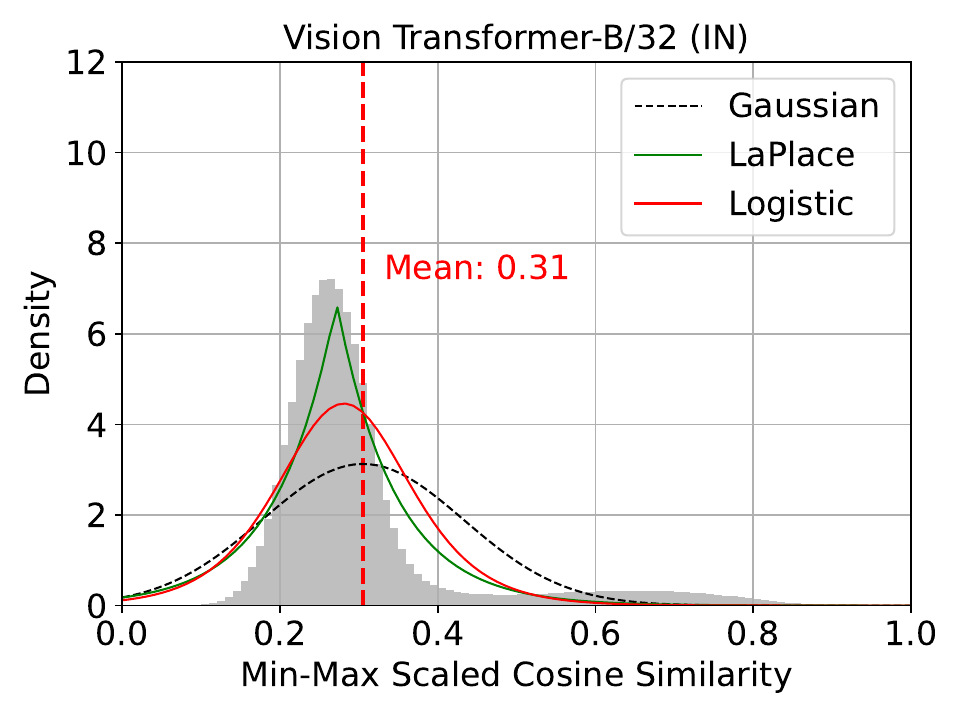}
    \includegraphics[width=0.3\linewidth]{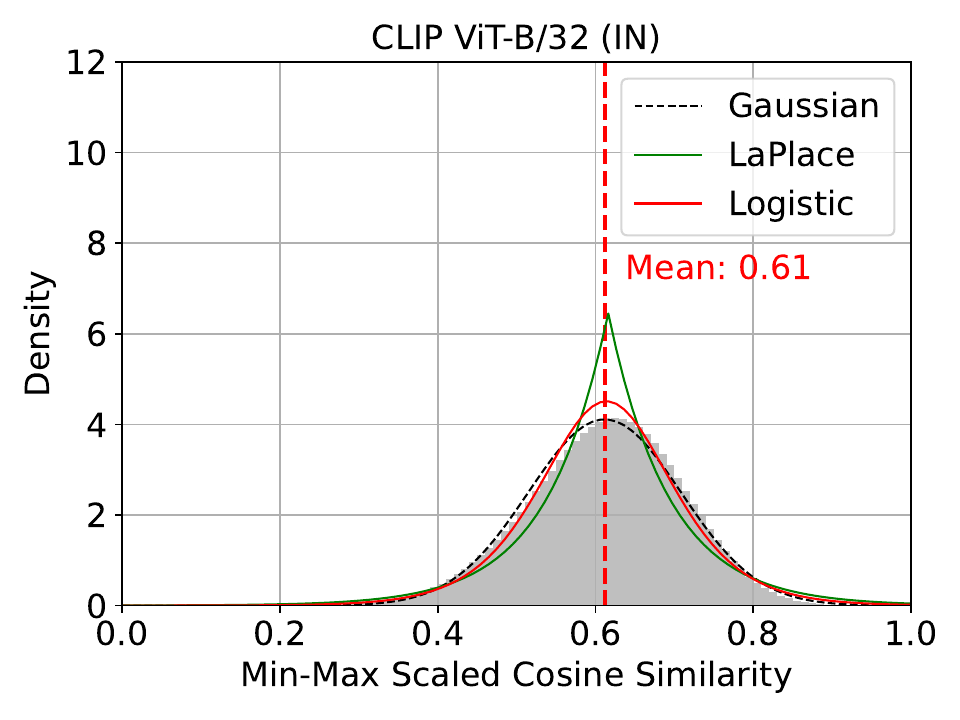}
    \caption{Distributions and mean for normalized ImageWoof}
    \label{fig:distr_iw}
\end{figure}

We ablate whether the mean of the normalized $s_c$ distribution is decisive for selecting the optimal $cal$.
During the normalization, the data mean may shift, depending on the distribution.
\cref{fig:distr_in} shows all normalized probability density plots and compares it to several data distributions.
We compare the clusterings for the same $\mathit{cal}$ as for ImageNette and $\mathit{cal} =0.94 - 0.98 * \mu$.
The normalized $\mu$ are visualized in \cref{fig:distr_iw}.

\newpage
\section{Embedding Space Analysis}
\begin{figure*}[ht]
    \centering
    \includegraphics[width=0.3\linewidth]{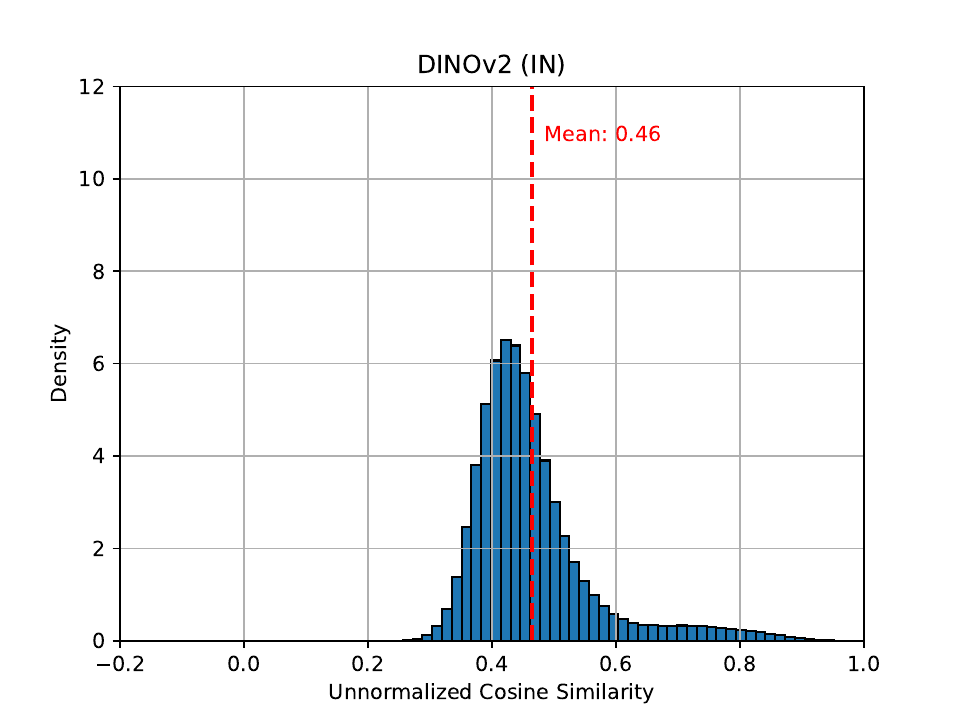}
    \includegraphics[width=0.3\linewidth]{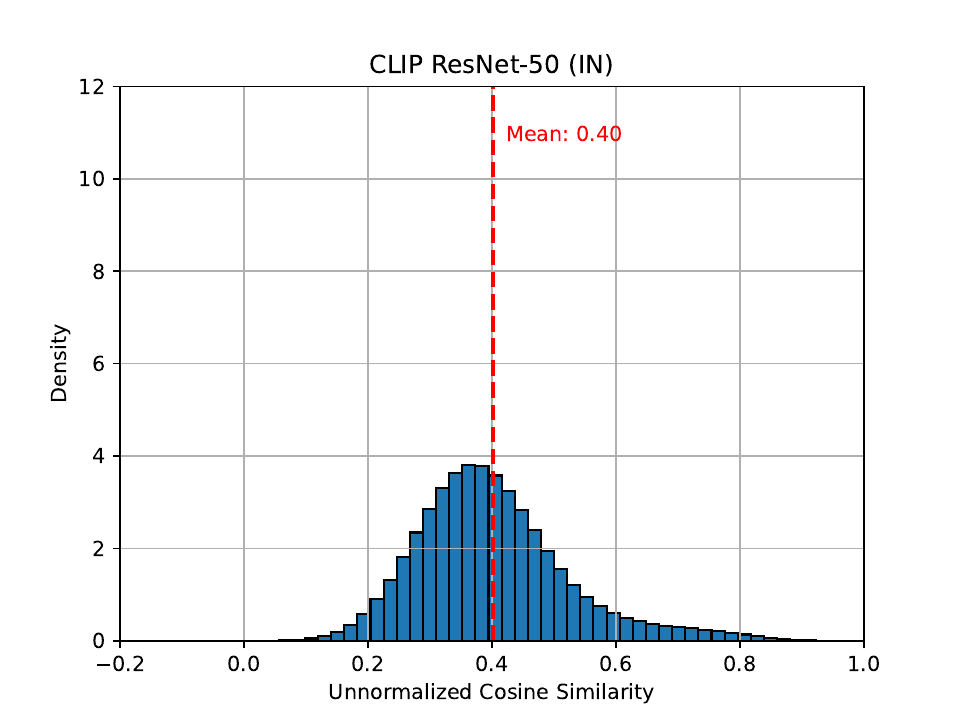}
    \includegraphics[width=0.3\linewidth]{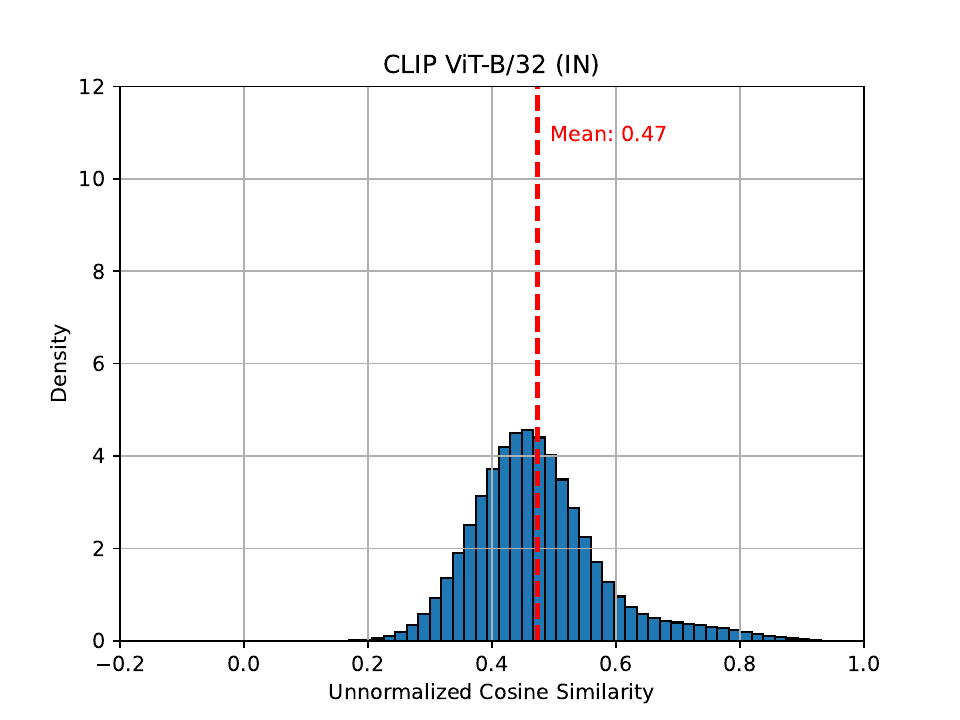}\newline
    \includegraphics[width=0.3\linewidth]{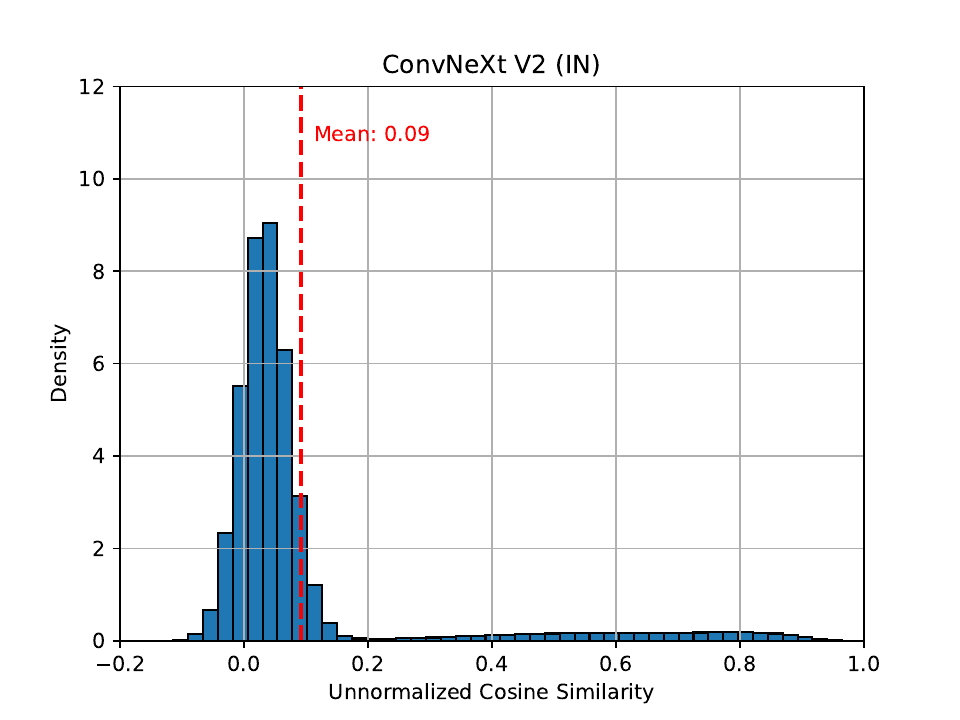} 
    \includegraphics[width=0.3\linewidth]{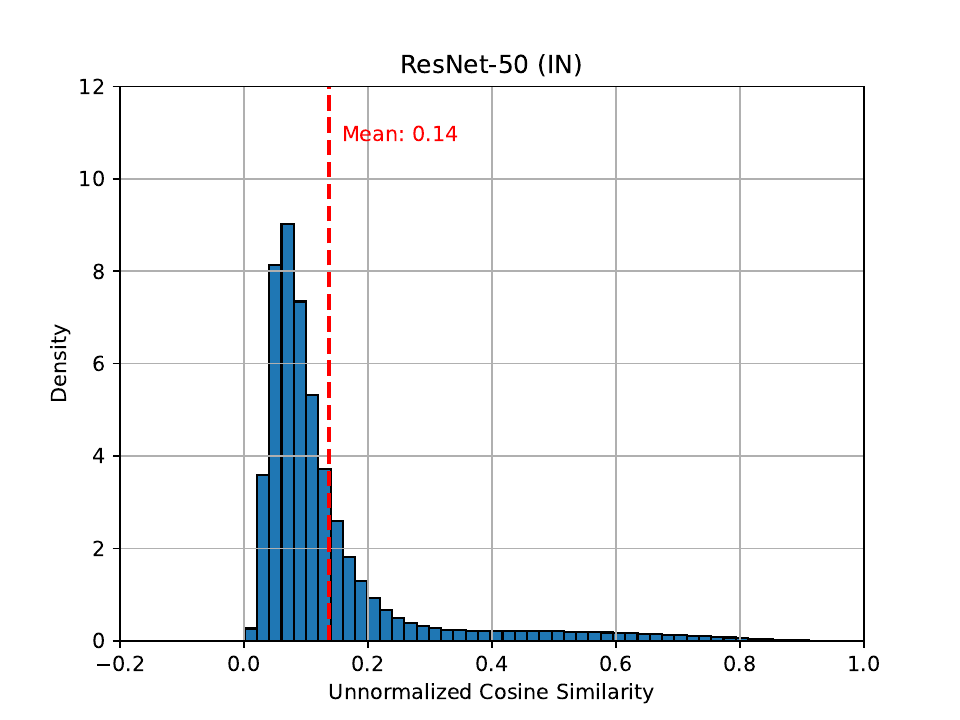}
    \includegraphics[width=0.3\linewidth]{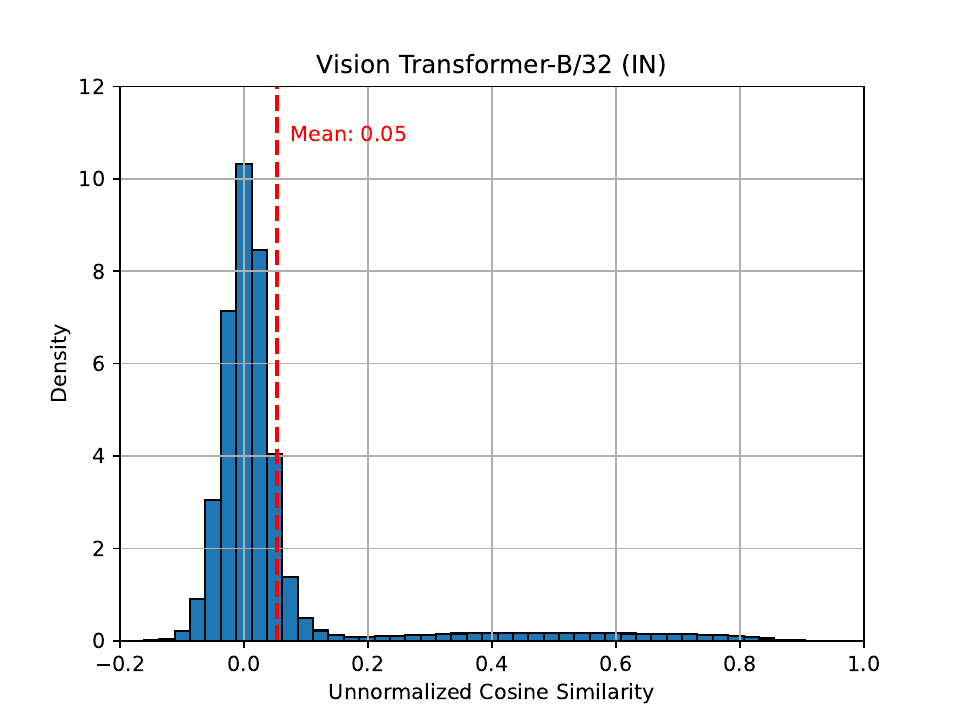}\newline
    \includegraphics[width=0.3\linewidth]{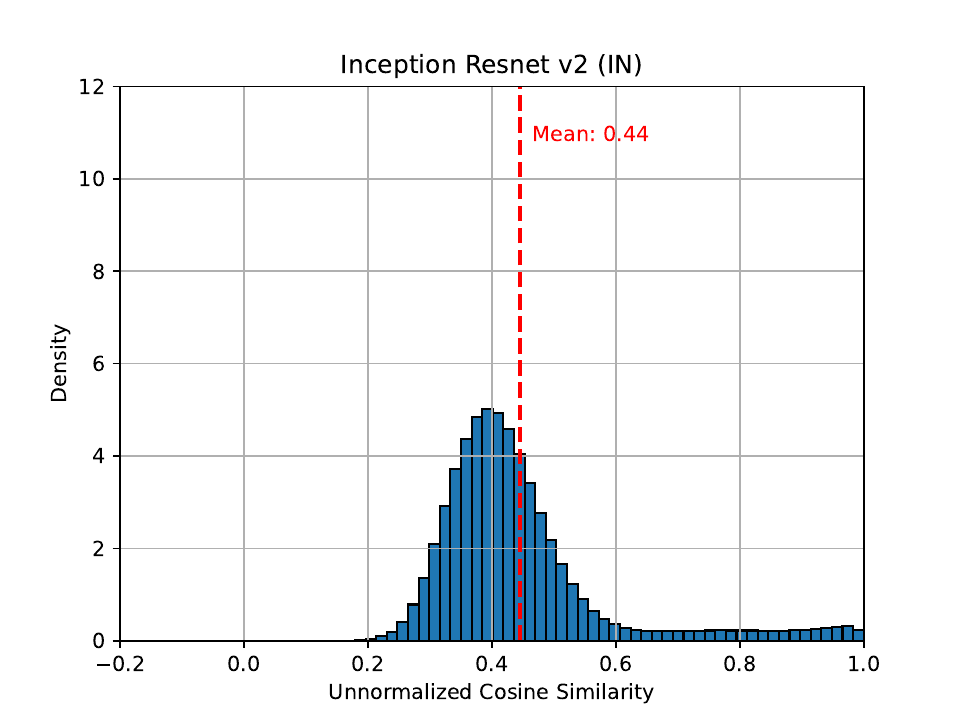}
    \caption{Probability density functions of un-normalized cosine similarity across embedding spaces for ImageNette. All distributions have a long tail towards the high cosine similarities. The \textit{narrow cone effect} can be observed for all models in the top row.}
    \label{fig:cosinesim_IN}
\end{figure*}

\cref{fig:cosinesim_IN} shows how the images' cosine similarities are distributed.

\section{Solver ablation}
\label{app:solver}

\begin{table}[ht]
    \centering
    \begin{tabular}{l@{\hspace{1cm}}cc@{\hspace{1cm}}cc}
    \toprule
        \multirow{2}{*}{Embedding space} & \multicolumn{2}{c}{Costs} & \multicolumn{2}{c}{Number of clusters} \\
        & GAEC & GAEC + KL & GAEC & GAEC + KL \\
    \midrule
        ConvNeXt V2 & $-116 \times 10^6$ & $\mathbf{-125 \times 10^6}$ &$325$& $\mathbf{281}$\\
        CLIP ViT-B/32 & $-116 \times 10^6$ & $\mathbf{-117 \times 10^6}$ &$1016$& $\mathbf{939}$\\
        DINOv2 & $-141 \times 10^6$ &  $\mathbf{-147 \times 10^6}$ &$114$&$\mathbf{102}$\\
        ViT-B/32 & $-150 \times 10^6$  & $\mathbf{-159 \times 10^6}$ &$459$& $\mathbf{404}$\\
        ResNet-50 & $-286 \times 10^6$ & $\mathbf{-309 \times 10^6}$ & $265$& $\mathbf{249}$\\
%        VGG19-BN & $-105 \times 10^6$ & && \\
    \bottomrule
    \end{tabular}
    \caption{On ClimateTV, we ablate the efficacy of using KL in conjunction with GAEC. It yields lower costs and fewer clusters in every case.}
    \label{tab:solver_abl}
\end{table}

\section{Results}
\label{app:res}

\subsection{ImageNette and ImageWoof}
\textbf{ImageNette}'s cluster sizes vary greatly, with the largest spread of ConvNeXt V2, followed by ResNet-50.
All model's median cluster sizes is below 5.
Most models have few outliers outside the quartiles, except for ConvNeXt V2.
All models with a narrow cone produce highly mixed clusters, while the other models have clusters sizes up to the class size (red line).
While most model's mean cluster size is between 298 and 362, the CLIP model's is 119, as it contains the largest number of single image clusters \ie 0.3\% of all images.
The majority (84\%) of ConvNeXt V2 image clusters contain images from a single class.
The mixed clusters are all combinations of 2 classes, which contain 1 or 2 mis-clustered samples.
While 71\% of all images are clustered in accordance with their class membership, few images have been added to incorrect classes.
%\vspace{-1em}

\begin{figure}[ht]
    \centering
    \begin{subfigure}[c]{0.49\linewidth}
        \includegraphics[width=\linewidth]{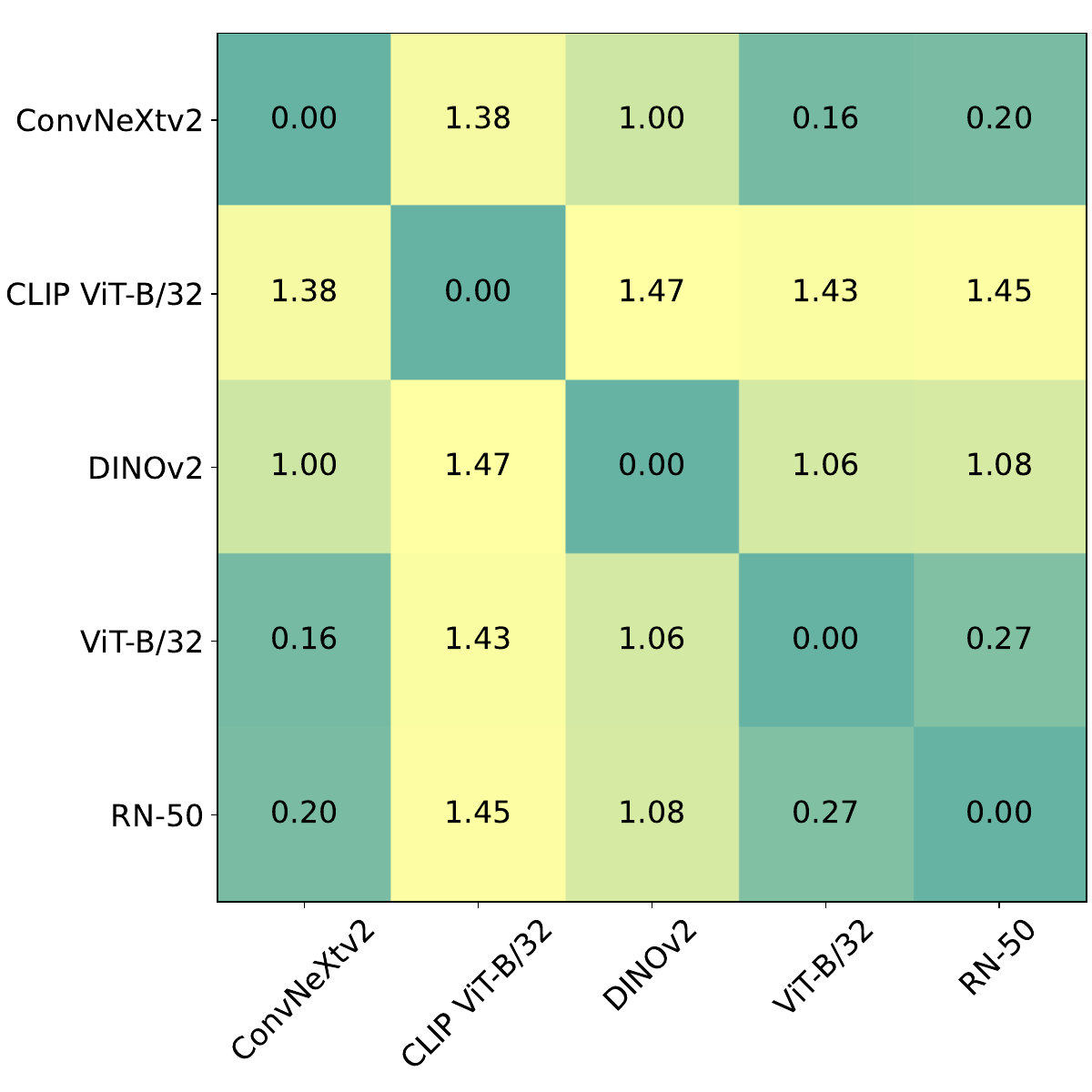}
    \caption{VI per Embedding for ImageNette.}  
    \end{subfigure}
    \begin{subfigure}[c]{0.49\linewidth}
        \includegraphics[width=\linewidth]{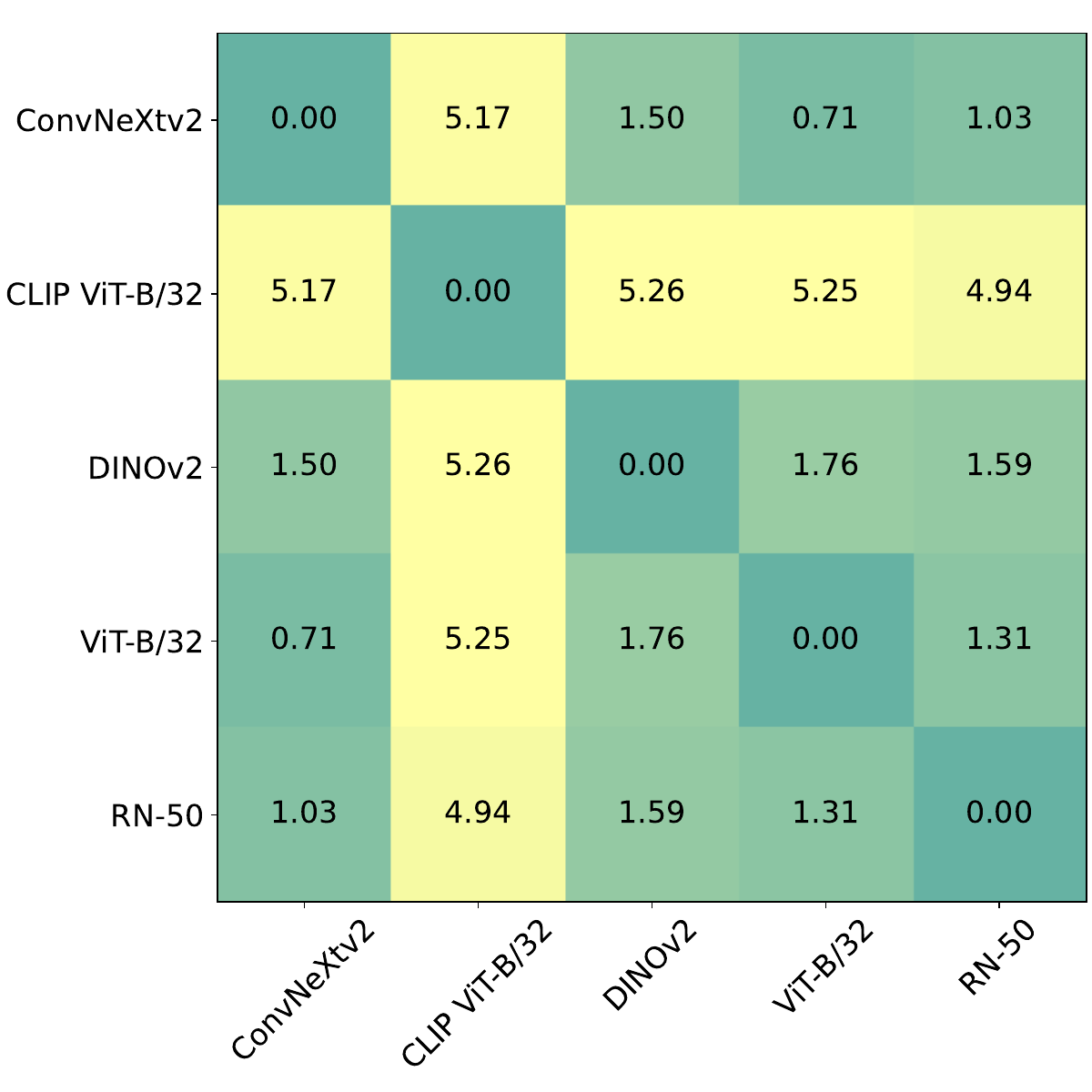}
    \caption{VI per Embedding for ImageWoof.}
    \label{fig:hm_iw}
    \end{subfigure}
    \caption{When comparing the VI per clustering, the CLIP ViT-B/32 is most distinct from the other models in both cases. This effect is even stronger in the more divere ImageWoof dataset.}
    \label{fig:hm_iniw}
\end{figure}

We evaluate the clusterings in terms of their VI.
\cref{fig:hm_iniw} compares the VI of the embedding clusterings to each other, indicating the disagreement between the different embedding spaces in terms of image similarity. %the original class labels.
For both datasets, the same trends are visible.
CLIP Vit-B/32 is the most distinct from the other clusterings, while ConvNeXt V2 and ViT-B/32 share the highest similarity.
Overall, the differences in VI are smaller for the more distinct classes in ImageNette compared to the fine-grained classification classes in ImageWoof.

%\newpage

\begin{figure}[ht]
    \centering
    \begin{subfigure}[c]{0.49\linewidth}
        \includegraphics[width=\linewidth]{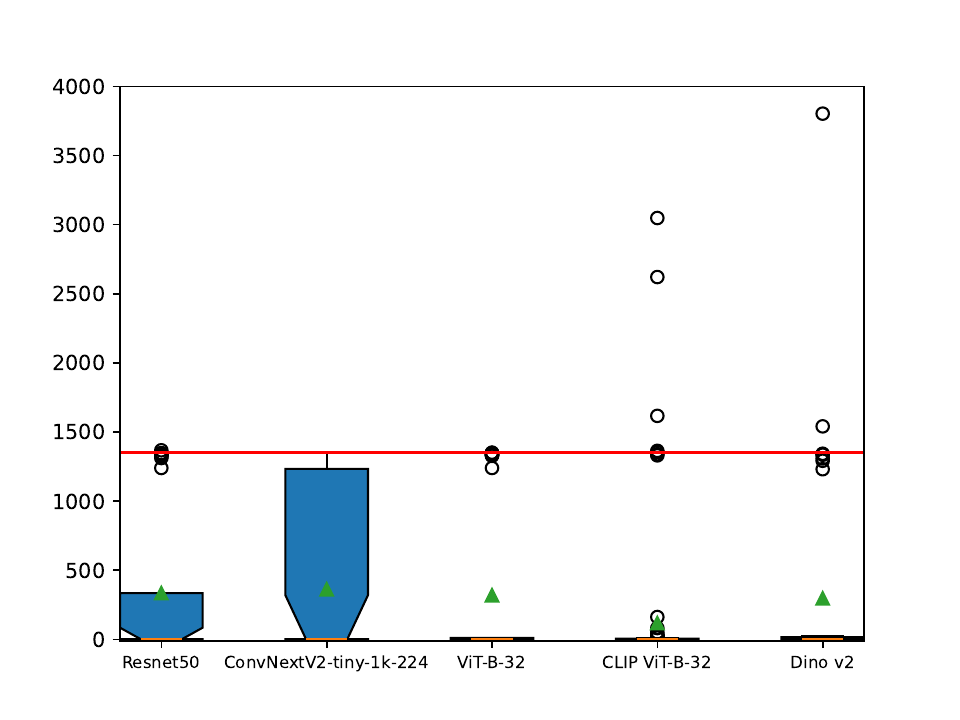}
        \caption{ImageNette cluster sizes are mainly smaller than the class size.}
        \label{fig:cs_iw}
    \end{subfigure}
    \begin{subfigure}[c]{0.49\linewidth}
        \includegraphics[width=\linewidth]{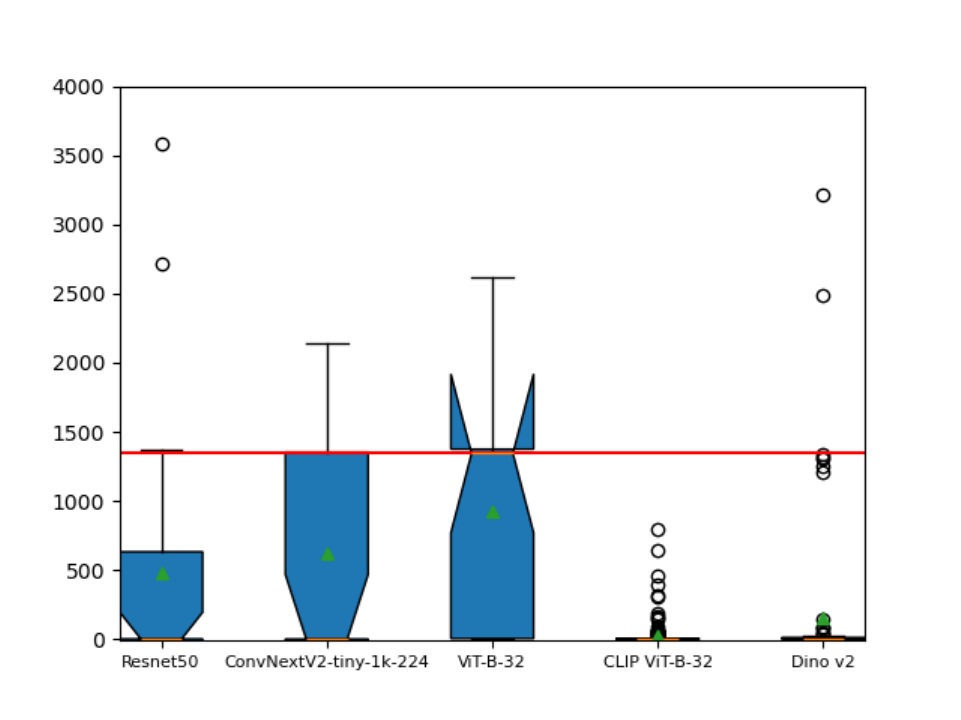}
        \caption{ImageWoof cluster sizes highly vary.}
        \label{fig:clust_sizes}
    \end{subfigure}
    \caption{Visualization of cluster sizes for embedding spaces employed (red line indicates class size). Overall, Resnet-50 and ConvNeXt V2 return larger clusters compared to CLIP ViT-B 32. Cluster sizes for the vision transformer (ViT-B 32) differ highly between datasets.}

\end{figure}

\cref{fig:cs_iw} shows the ImageNette cluster sizes for the optimal $\mathit{cal}$. 
In contrast to ImageWoof (\cref{fig:clust_sizes}, the ViT-B/32 cluster sizes are much less distributed.
For ImageWoof, cluster sizes are generally larger and also ResNet-50 and ConvNeXt V2 clusters now also exceed the number of images per class.
The same trends can be observed for the other models.
\newpage
\subsection{ClimateTV}
\label{appi: climtv}

\begin{figure}[ht]
    \centering
    \includegraphics[width=0.5\linewidth]{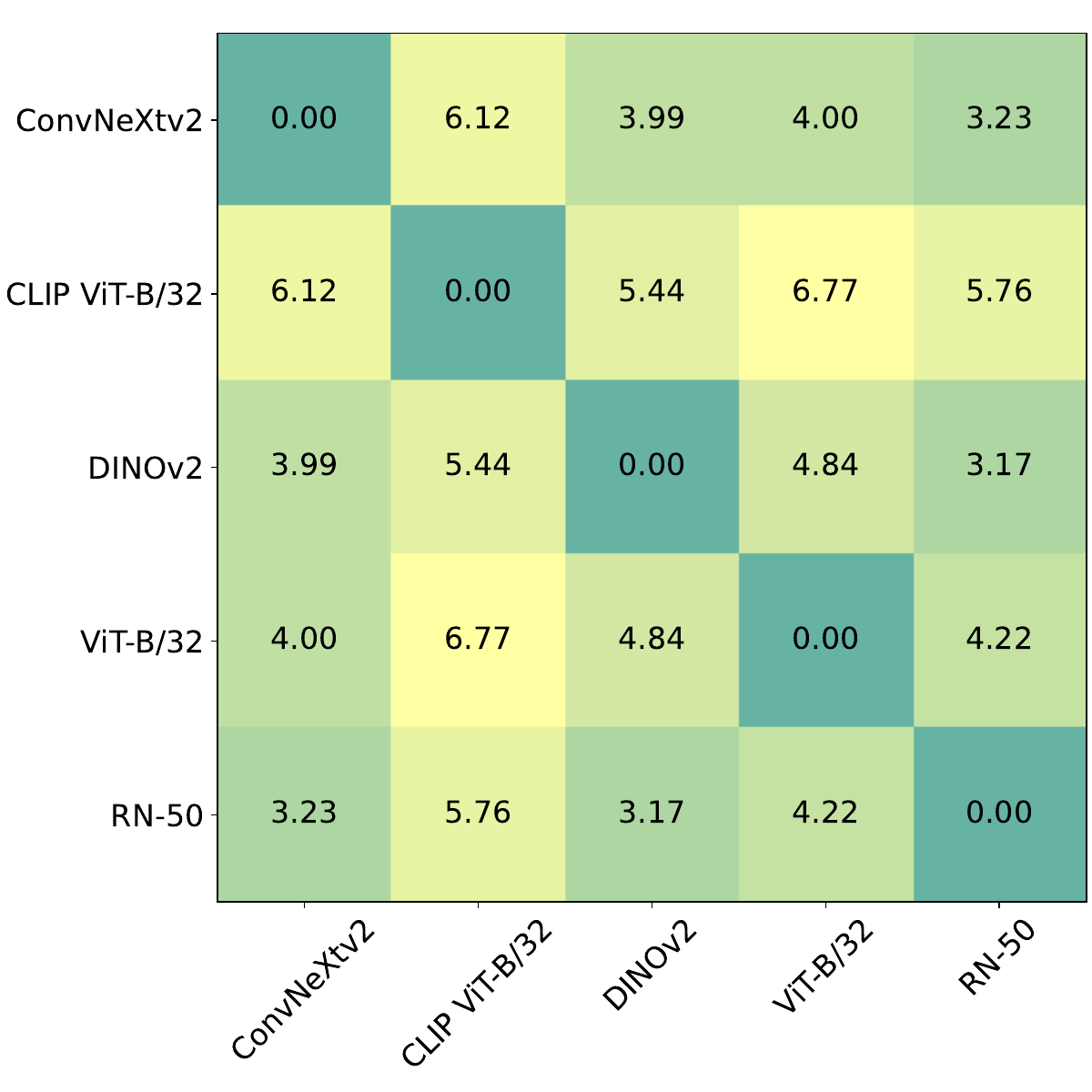}
    \caption{VI Heat Map for ClimateTV clusterings.}
    \label{fig:heatmap_tvjan}
\end{figure}

When comparing the clustering performance for ClimateTV, the differences between the embedding models become larger in terms of VI.
The greatest similarity can be found between the DINOv2 and ResNet-50 model.
Again, the CLIP clustering appears to be most dis-similar to the other clusterings, as shown in \cref{fig:heatmap_tvjan}.

\begin{figure}[ht]
    \centering
    \includegraphics[width=0.95\linewidth]{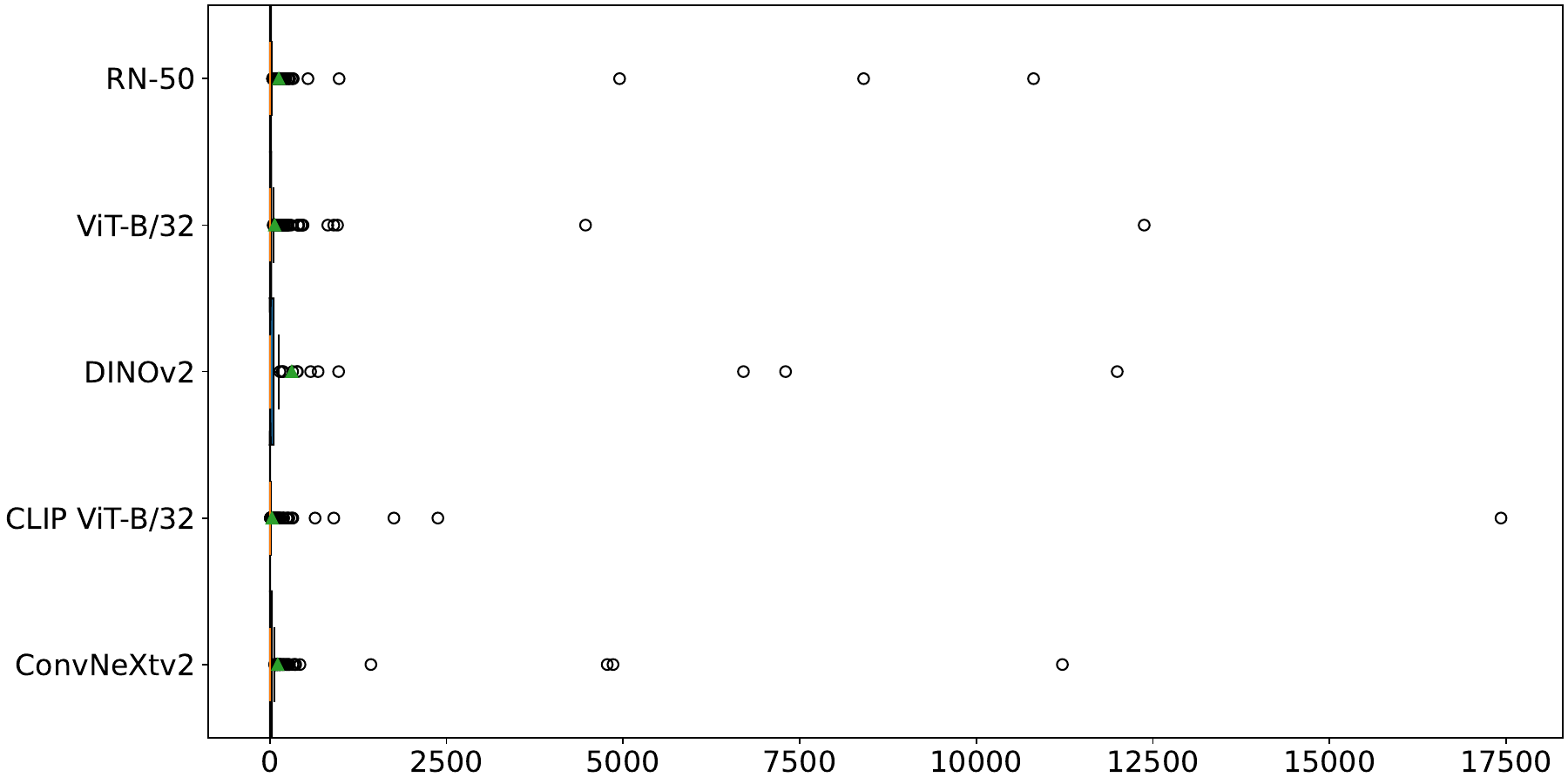}
    \caption{The comparison of ClimateTV cluster sizes shows that CLIP has by far the largest cluster. In contrast to the other models, CLIP's average cluster size is below 35.}
    \label{fig:cs_tvjan}
\end{figure}

ClimateTV produces clusters of up to 17k images.
All embedding clusterings have a handful of large clusters with a large majority below 500, as shown in \cref{fig:cs_tvjan}.
\cref{tab:tv_stats} contains the detailed statistics for all embedding clusterings. 
It stands out that CLIP ViT-B/32 has a large number of cluster of size 1.
Since the dataset includes 37k images, this makes almost 0.4k images in single image clusters.
Moreover, CLIP ViT-B/32 clusterings contain many small classes, as the median shows.

\begin{table}[ht]
    \centering
    \begin{tabular}{lcccccc}
    \toprule
        Embedding & \# clust & min. size & median size & mean size & max. size & \% size 1 \\
        \midrule
        ConvNeXt V2 & 281 & 1 & 9 & 112.9 &11,218 &0.03\% \\
        CLIP ViT-B/32 & 1016 & 1 & 2& 31.2 & 17,426 & 1.0\% \\
        DINOv2 & 102 & 1 & 6& 310.9& 11,994& 0.06\%\\
        Vit-B/32 & 459& 1 & 7& 69.1 & 12,376 &0.08\%\\
        ResNet-50 & 249 & 1 & 5& 127.4 & 10,810 & 0.12\% \\
    \bottomrule
    \end{tabular}
    \caption{ClimateTV cluster statistics}
    \label{tab:tv_stats}
\end{table}

We compare the largest clusters' contents for each embedding clustering in \ref{tab:clust_cont}
It stands out that the largest CLIP ViT-B/32 class is very noisy. 
It contains nature or its products (\eg fruits) in natural (\eg arctic sea with iceberg), generate (\eg visualization of the globe), industrial (\eg wind energy plant), or catastropical (\eg flooded city).
It appeared highly unlikely that this many images share a common features.
We are very satisfied with the ConvNeXt V2 clusters, as they contain specific semantic concepts.
DINOv2 and CLIP embeddings also appear to contain semantic information, due to the higher specificity of their clusters compared to ResNet-50 and ViT-B/32.
The ResNet-50 cluster content was most difficult to infer, as the models had visual similarity, but varied a lot. 
It appears that the ResNet-50 embeddings have a shape-bias.
This could be an explanation for the \textit{circles} cluster which contains numerous round object from buttons to the earth.
The ViT-B/32 cluster content was quite abstract, with the common theme of \textit{blue} in the largest cluster.
One of CLIP embeddings strong suits is text understanding.
One CLIP cluster which contains political news is combined with all other computer generated content by ConvNeXt V2.
In contrast, ConvNeXt V2 can detect images that contain red soil or sand, which CLIP clusters in its large \textit{nature} cluster.
There are no two clusters of the two models which are alike, \ie share 80\% or more of images.
The combination of the two clusterings appears less helpful.
When looking at CLIP cluster for ConvNeXt V2's \textit{polar bear} cluster, the single image clusters are less helpful in detecting outliers due to several of them containing polar bears.
Moreover, the larger clusters are not semantically different from each other, so their benefit is small.

\begin{table}[ht]
    \centering
    \begin{tabular}{lccccc}
    \toprule
    & ConvNeXt V2 & CLIP ViT-B/32 & DINOv2 & ViT-B/32 & ResNet-50 \\
    \midrule
    \#1 content& comp. gen. & nature  & humans & blue color  & comp. gen. \\
    (size)& (11,218) &(17,426)  &(11,994) & (12,376)& (10,810) \vspace{1em}\\
    \#2 content & speaker & Formal humans & event info & humans&humans\\
    (size)& (4,858) & (2,379)& (7,300)& (4,469)& (8,405)\vspace{1em}\\
    \#3 content & outdoor photo & text w/ "climate change" & nature & outdoor photo& nature \\
    (size)& (4,776)  & (1,757) & (6,704)& (959) & (4,951)\vspace{1em} \\
    \#4 content & protest & poster/presentation& map/globe& words& circles \\
    (size)& (1,431) & (906)& (974)& (906) & (981)\vspace{1em}\\
   
    \#5 content & portraits & cold & food & humans w/ text& nature w/ foreground\\
     (size)& (427) &(641)& (684)&(821)& (540)\\
    %6. largest cluster (size) & satellite / earth visualizations &&&&\\
    \bottomrule
    \end{tabular}
    \caption{ClimateTV cluster size and content for its top 5 largest clusters per embedding model. comp. gen. = Computer generated content.}
    \label{tab:clust_cont}
\end{table}

\begin{figure}[ht]
    \centering
    \includegraphics[width=0.228\linewidth]{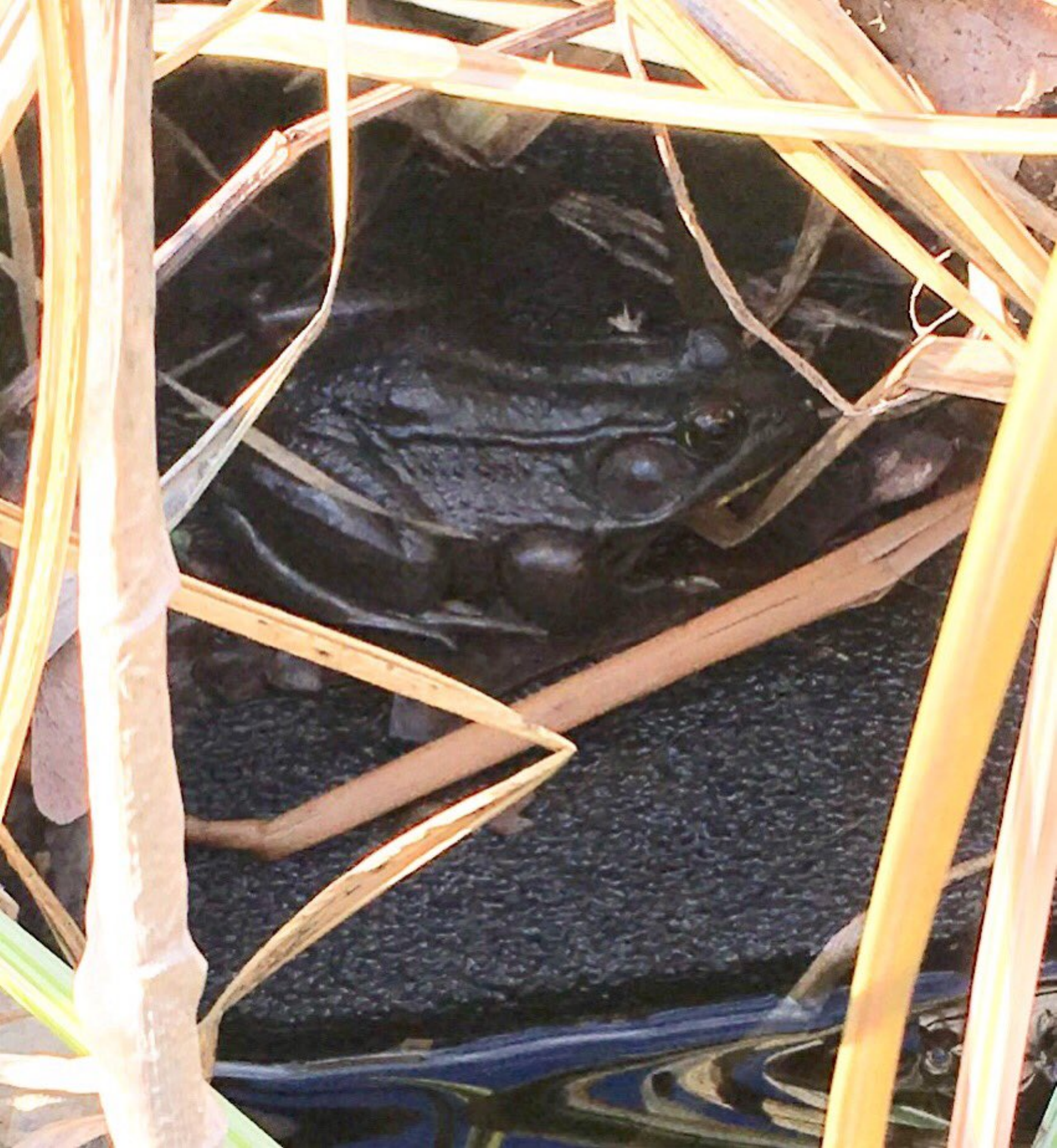}
    \hspace{0.1cm}
    \includegraphics[width=0.33\linewidth]{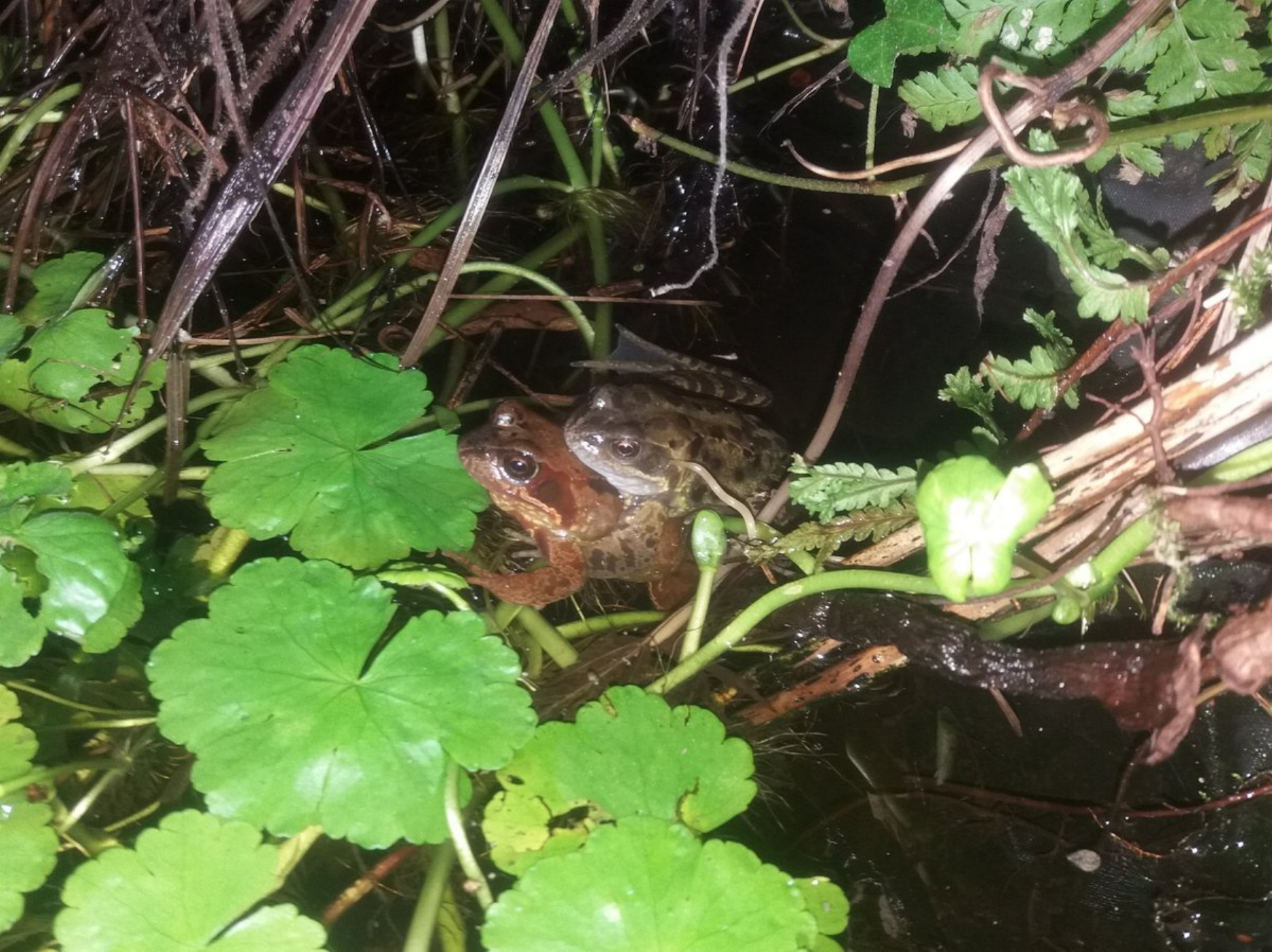}
    \hspace{0.1cm}
    \includegraphics[width=0.33\linewidth]{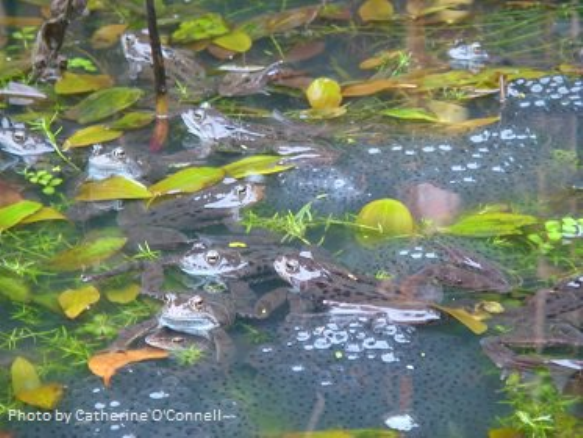} \\
    \vspace{0.1cm}
    \includegraphics[width=0.385\linewidth]{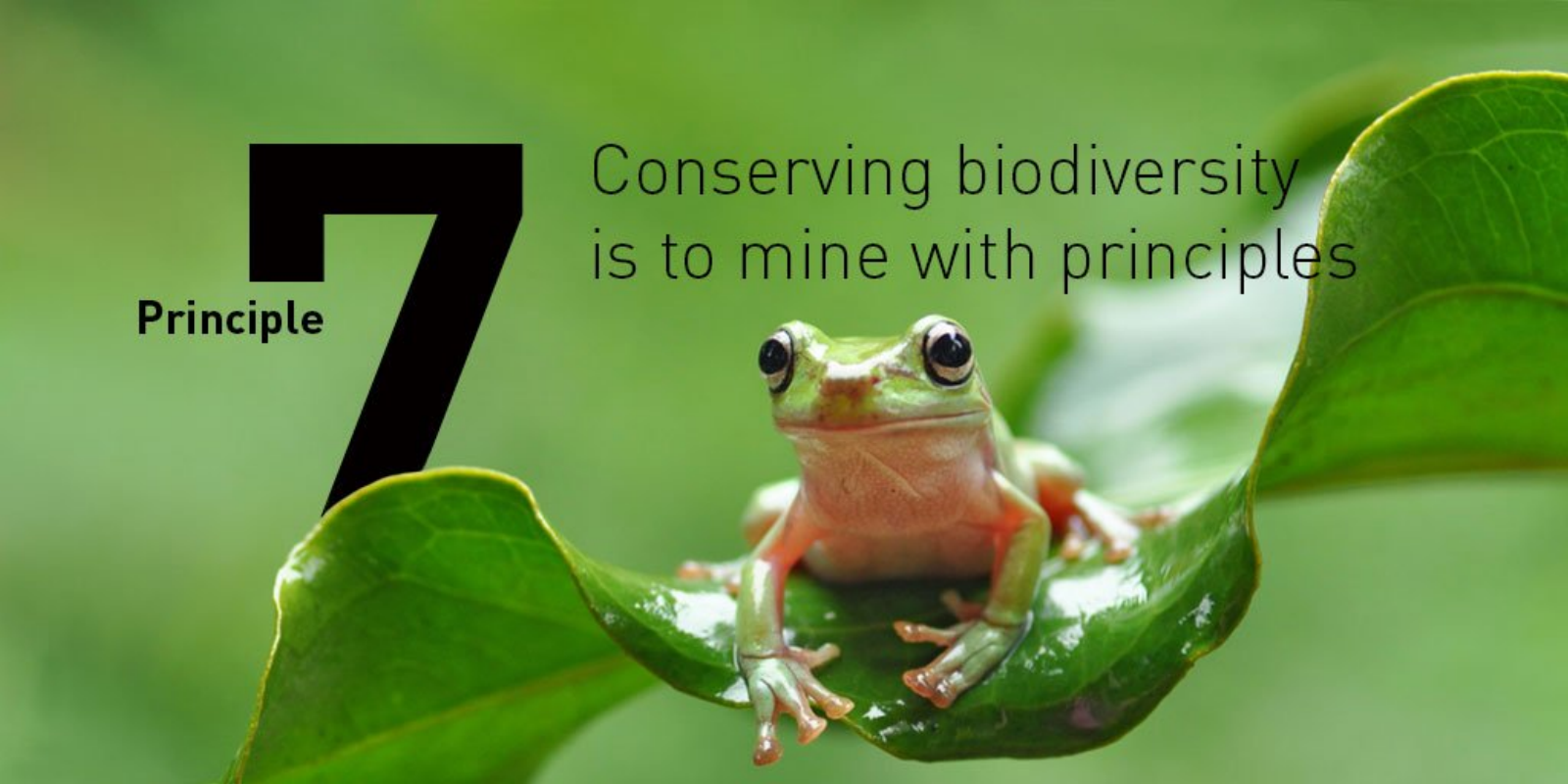}
    \hspace{0.1cm}
    \includegraphics[width=0.29\linewidth]{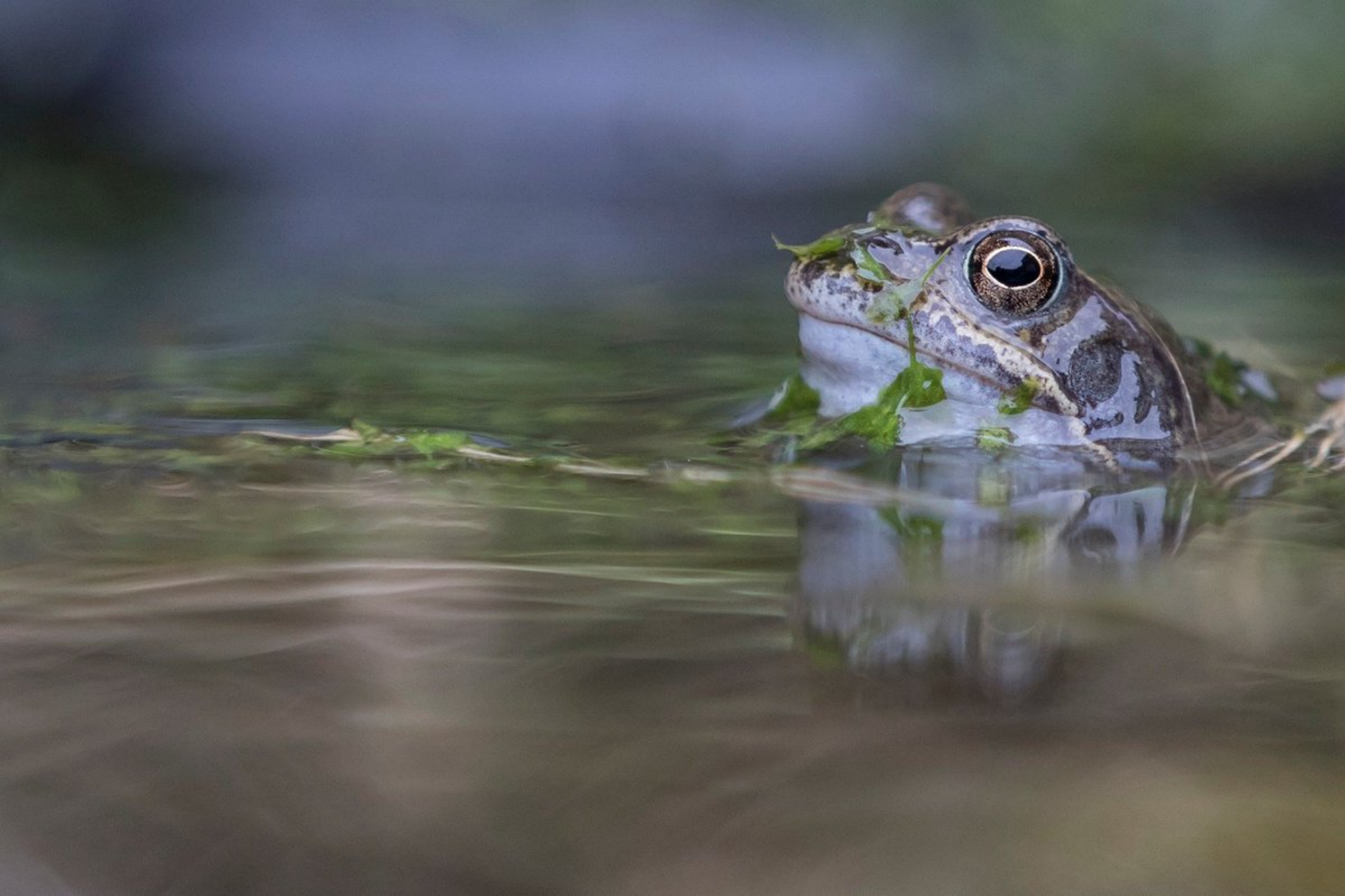}
    \hspace{0.1cm}
    \includegraphics[width=0.29\linewidth]{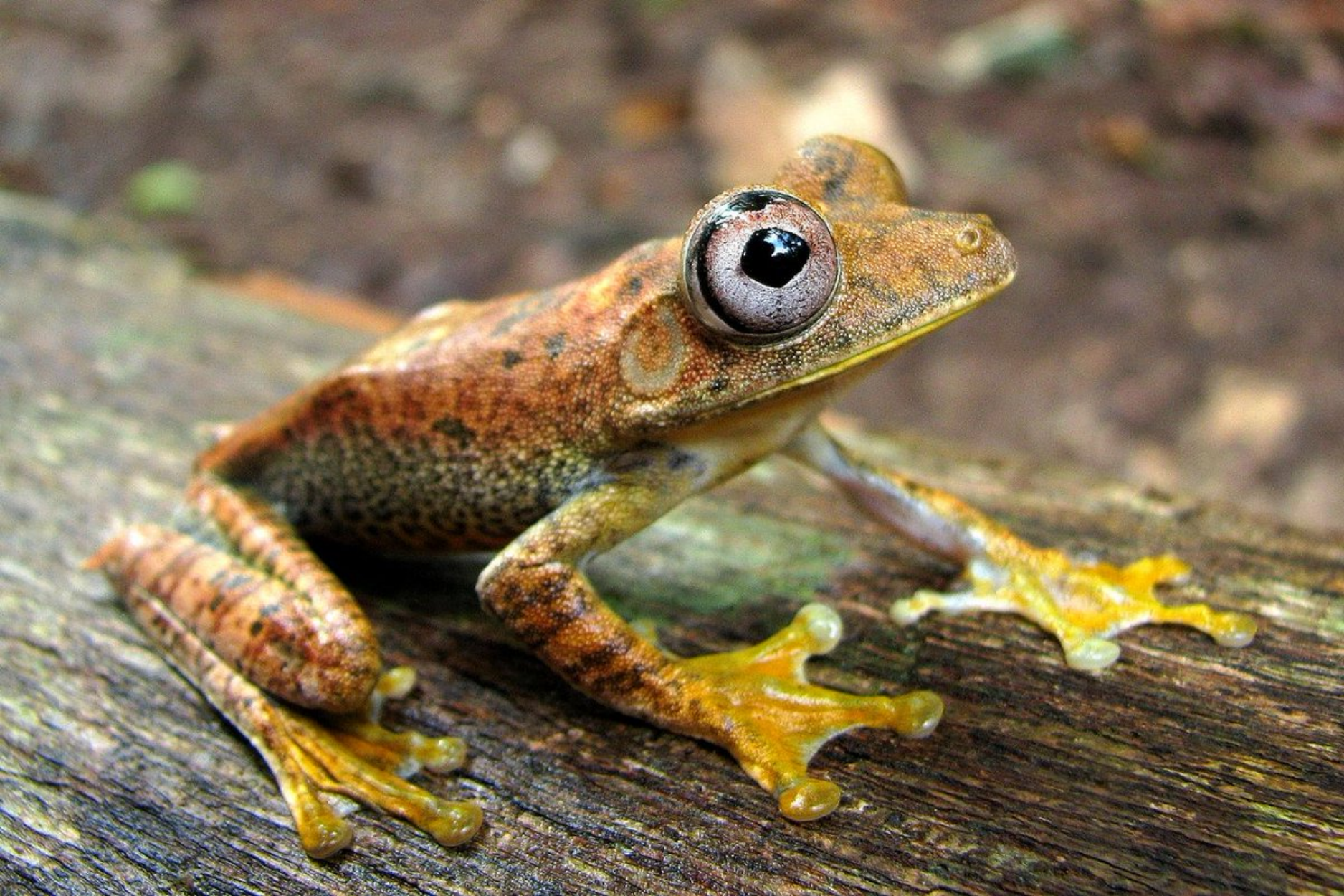} \\
    \vspace{0.1cm}
    \includegraphics[width=0.35\linewidth, trim={3cm 0 1cm 0},clip]{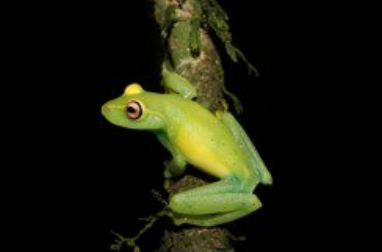}
    \hspace{0.1cm}
    \includegraphics[width=0.35\linewidth]{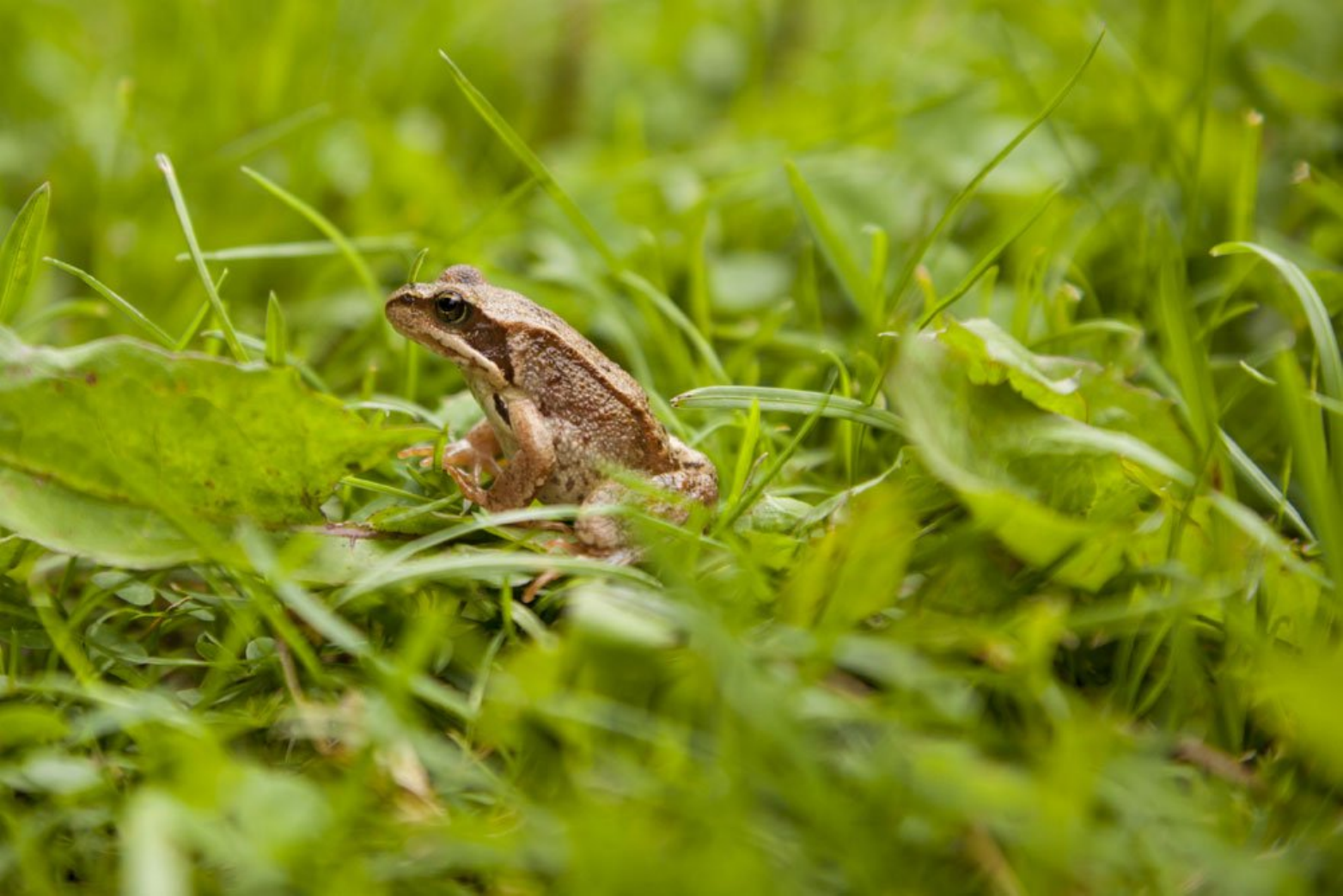}
    \hspace{0.1cm}
    \includegraphics[width=0.23\linewidth]{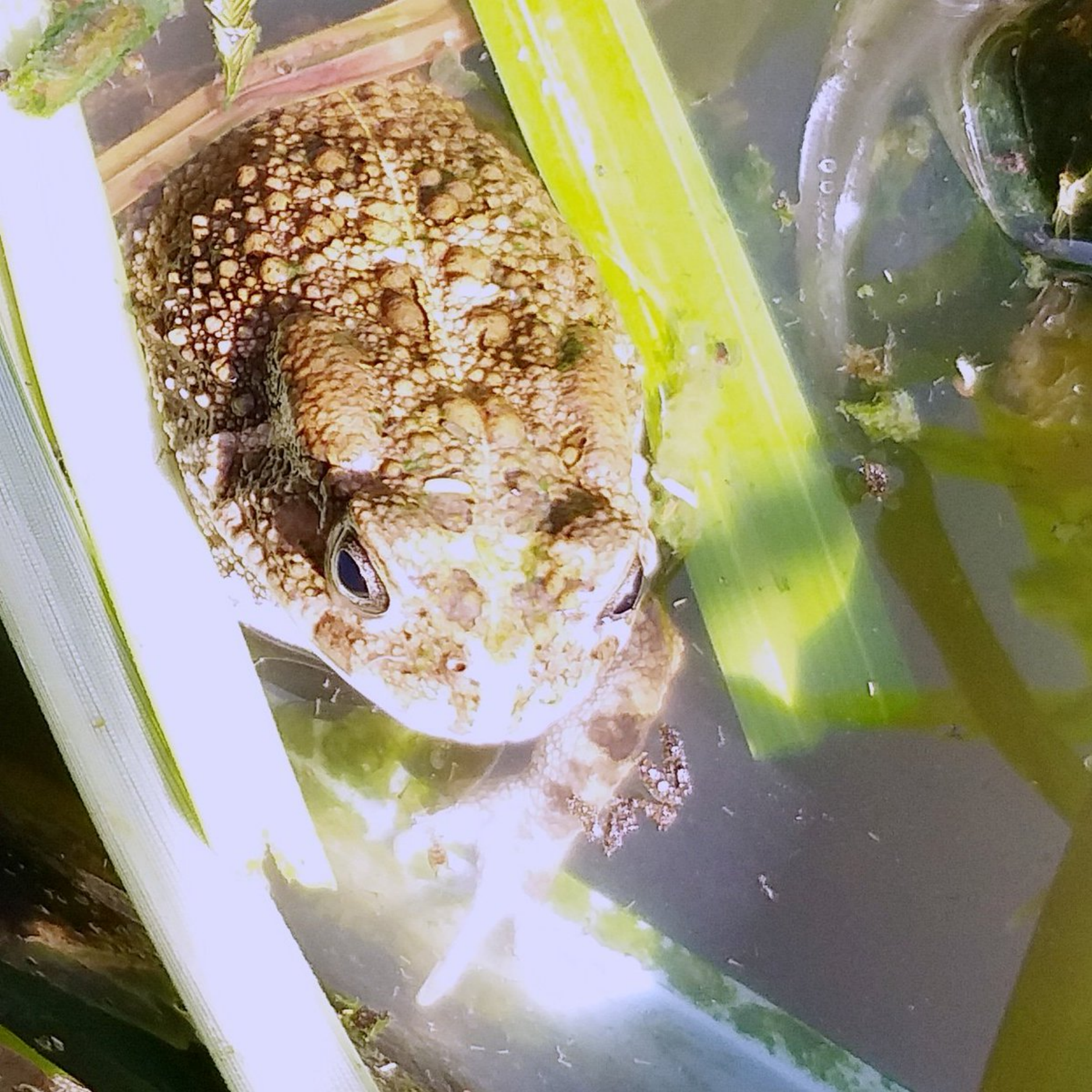} \\
     \vspace{0.1cm}
    \includegraphics[width=0.33\linewidth, trim={3cm 0 1cm 0},clip]{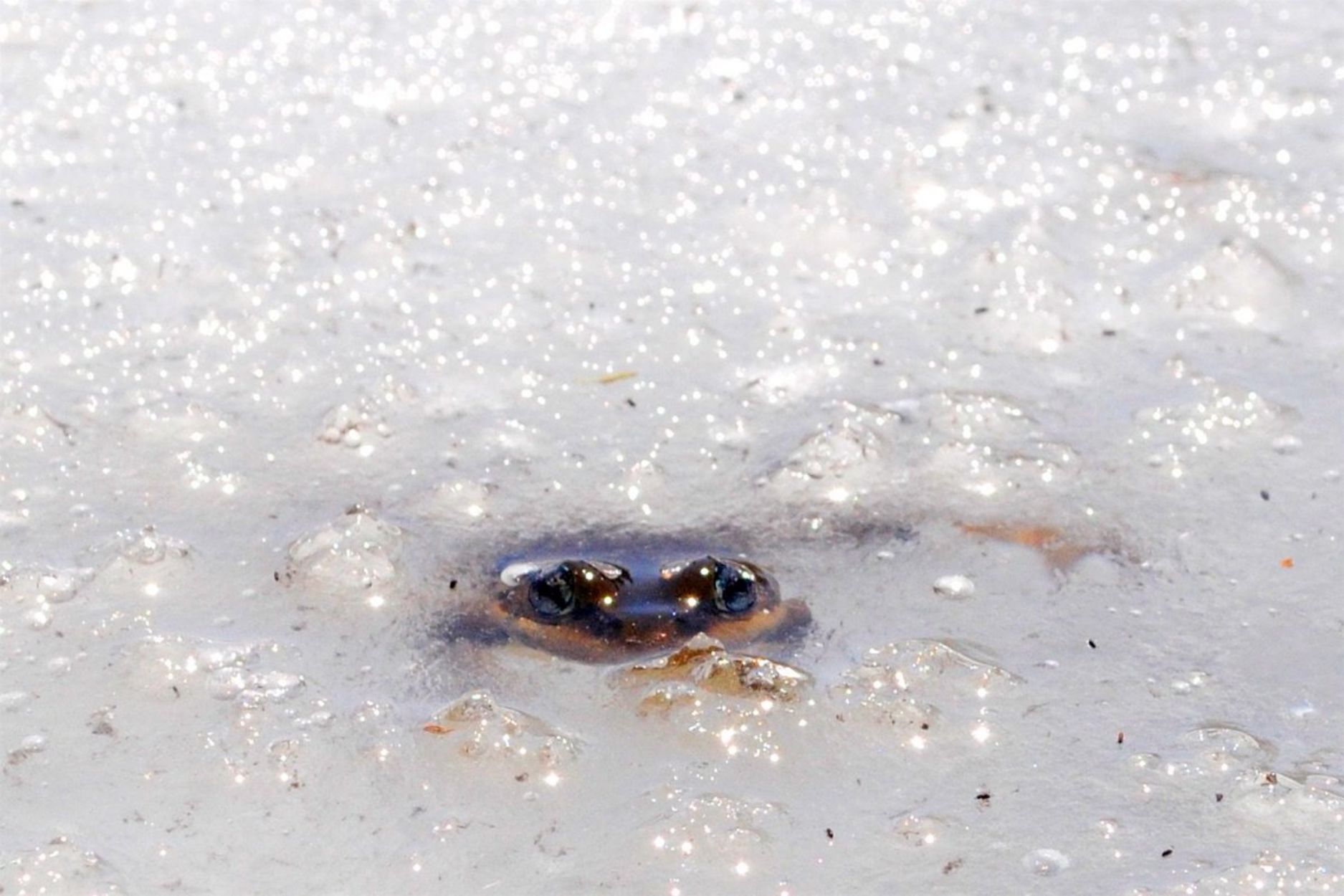}
    \hspace{0.1cm}
    \includegraphics[width=0.17\linewidth]{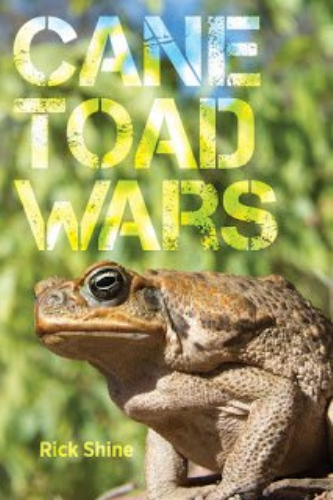}
    \hspace{0.1cm}
    \includegraphics[width=0.45\linewidth]{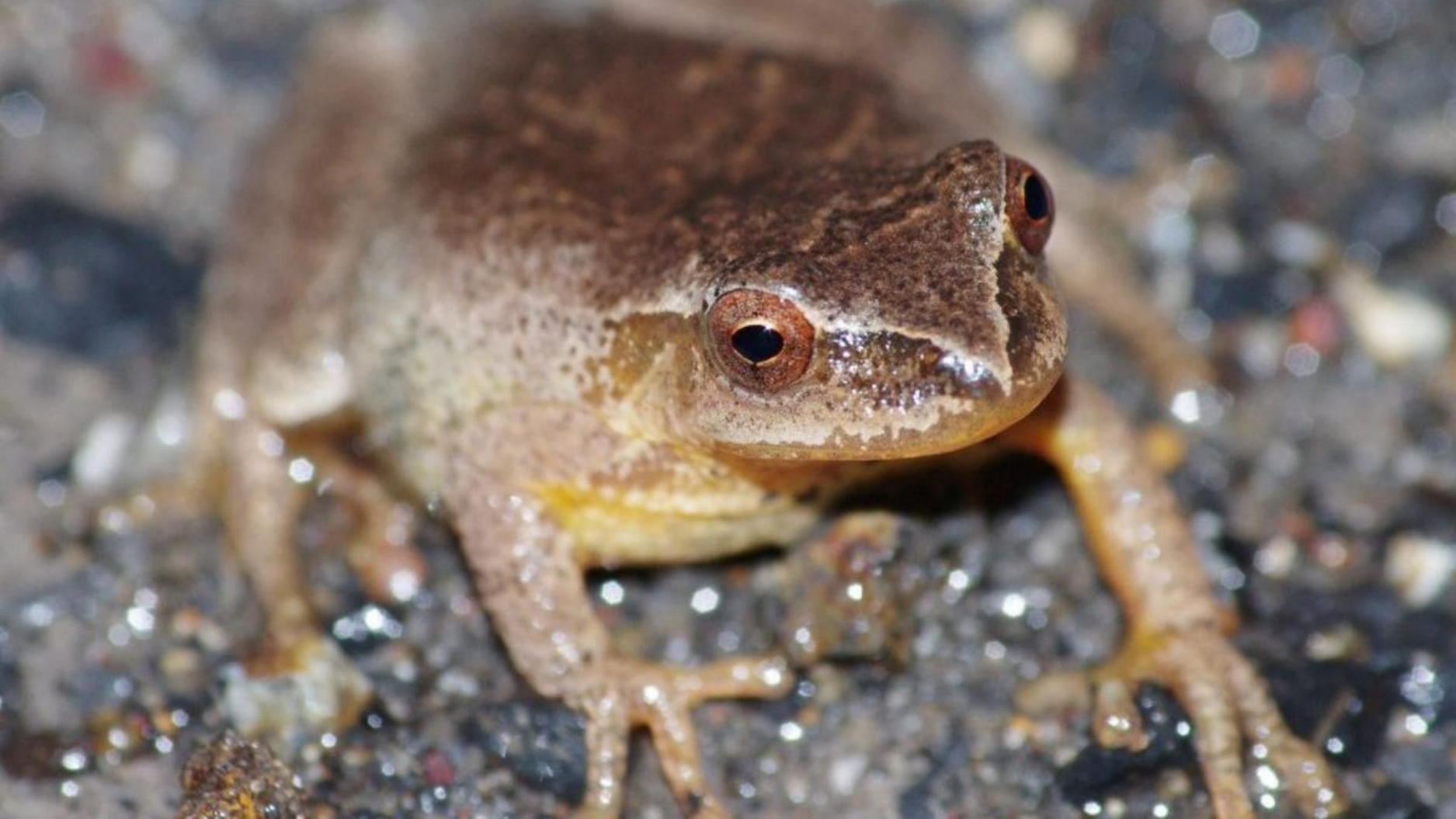} \\  
        
    \caption{The \textit{frog} cluster has the largest overlap between the clusterings of ConvNeXt V2 and DINOv2.}
    \label{fig:frog}
\end{figure}

\begin{figure}[ht]
    \centering
        \includegraphics[width=0.135\linewidth, trim={3cm 0 1cm 0},clip]{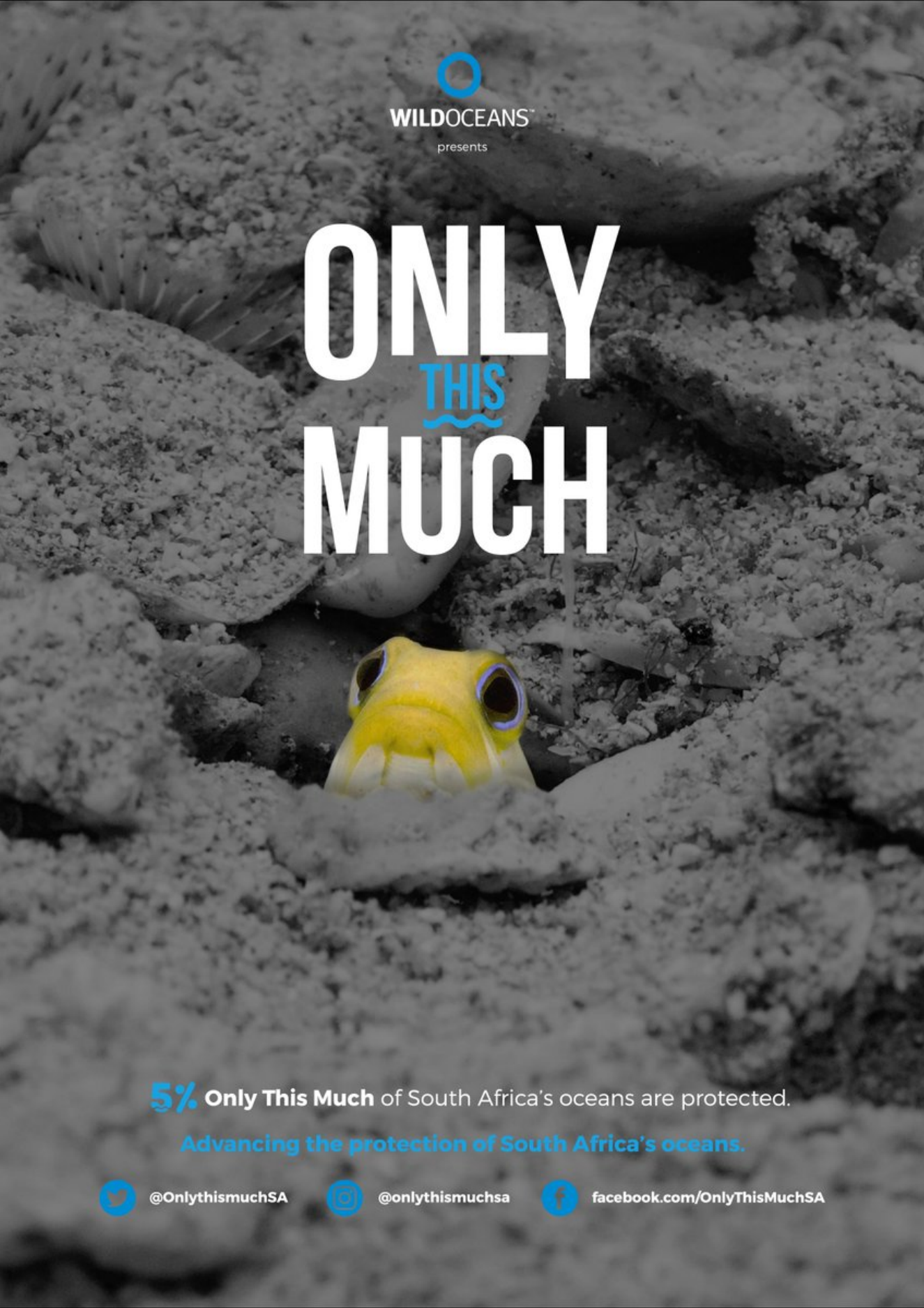}
    \hspace{0.1cm}
    \includegraphics[width=0.371\linewidth]{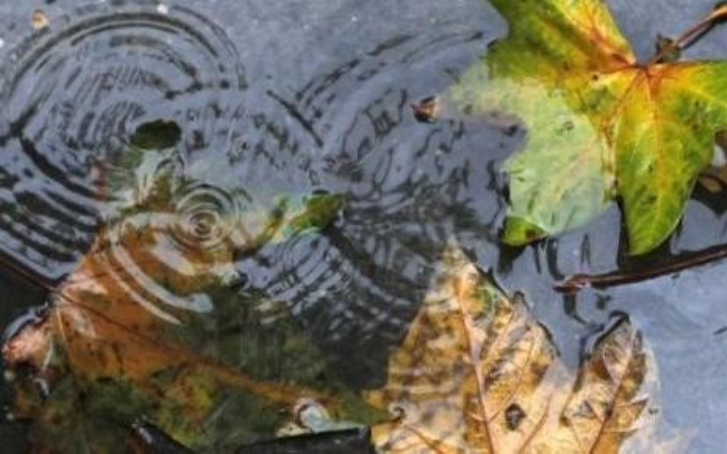}
    \hspace{0.1cm}
    \includegraphics[width=0.438\linewidth]{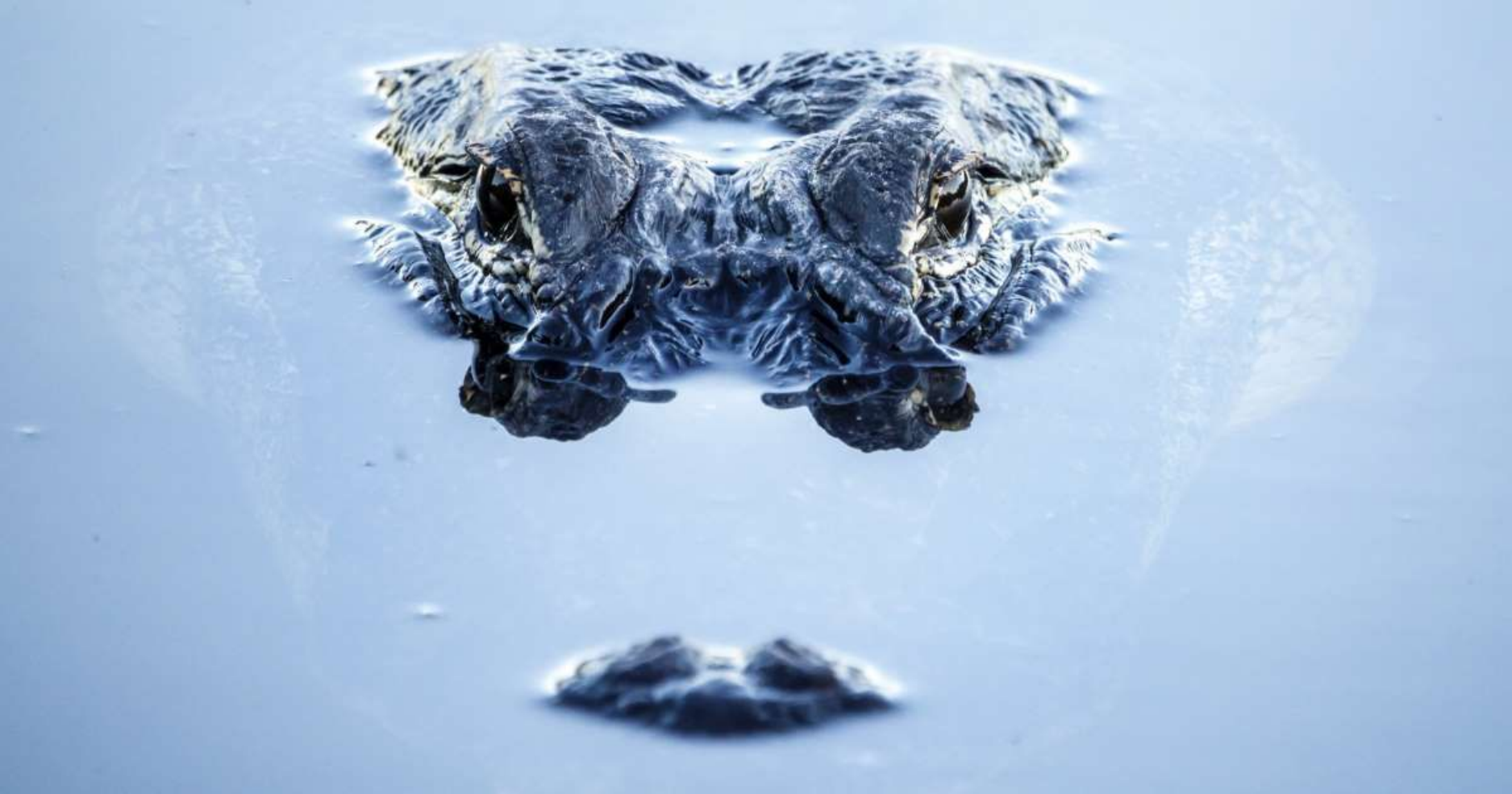} \\
    \caption{DINOv2's \textit{frog} also contained 3 images that were not shared with the ConvNeXt V2 cluster.}
    \label{fig:frog_dino}
\end{figure}

\begin{figure}[ht]
    \centering
    \includegraphics[width=0.267\linewidth, trim={3cm 0 1cm 0},clip]{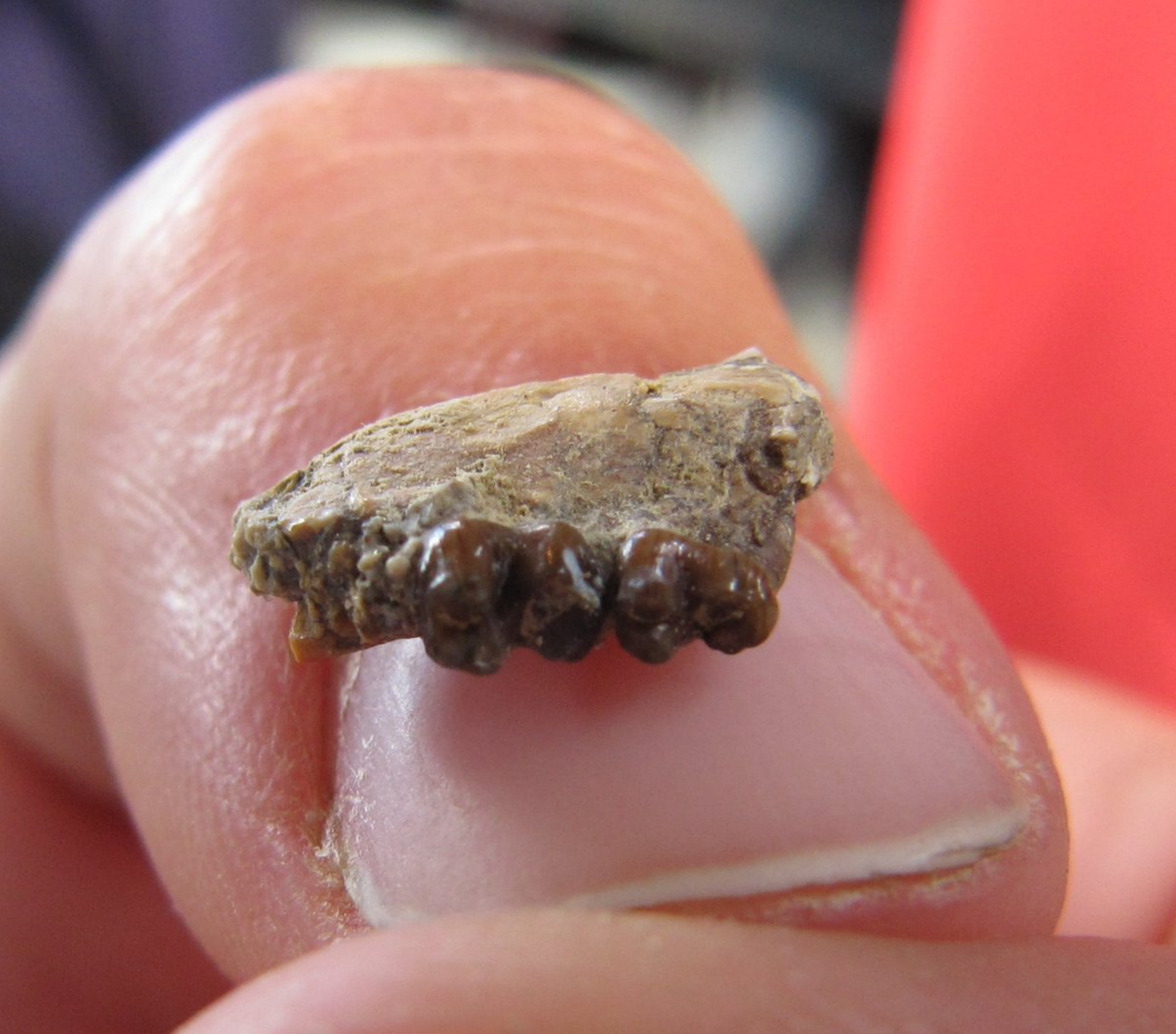}
    \hspace{0.1cm}
    \includegraphics[width=0.223\linewidth]{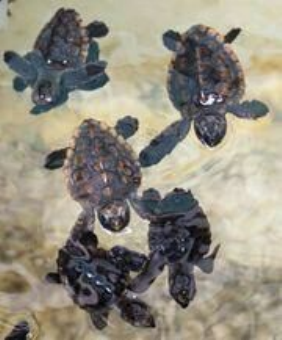}
    \hspace{0.1cm}
    \includegraphics[width=0.475\linewidth]{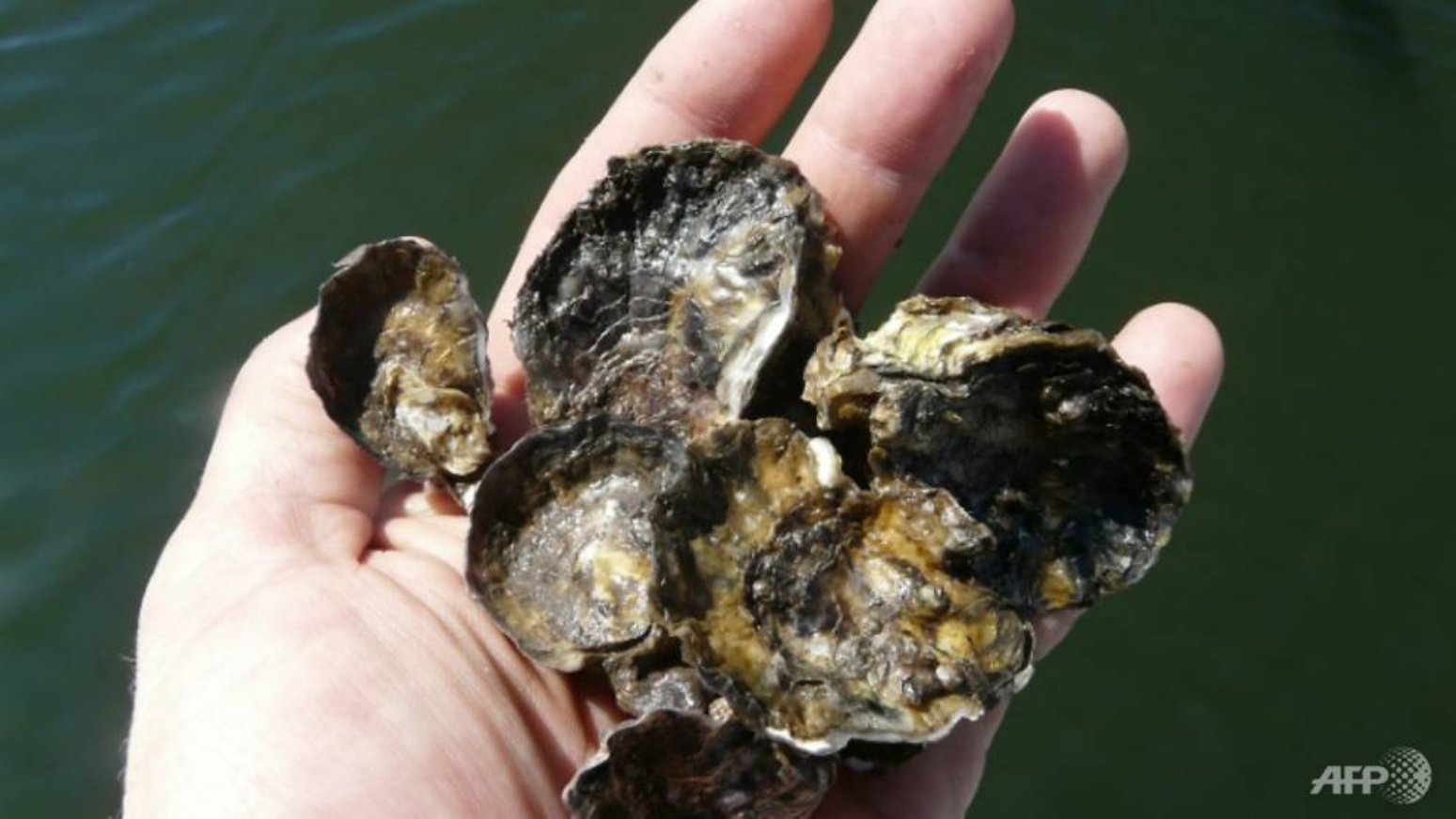} \\
    \caption{ConvNeXt V2's \textit{frog} also contained 3 images that were not shared with the DINOv2 cluster.}
    \label{fig:frog_conv}
\end{figure}

In the main paper we reported the biggest overlap between DINOv2 and ConvNeXt V2 clusters being \textit{frogs}. 
We visualized the jointly clustered images in \cref{fig:frog}.
\cref{fig:frog_conv} and \cref{fig:frog_dino} contain the images which have been included in the ConvNeXt V2 and DINOv2 cluster respectively.

\end{document}

%% file: Fig/bayesian_network.tex
\begin{figure}
    \centering
\usetikzlibrary{shapes.geometric,fit}

\begin{tikzpicture}
    \node[ellipse, draw, minimum width=1.4cm, minimum height=0.5cm] (X) at (0,0) {$x_e$};
    \node[ellipse, draw,minimum width=1.4cm, minimum height=0.5cm] (Y) at (2,0) {$y_e$};
    \node[ellipse, draw, minimum width=1.4cm, minimum height=0.5cm] (Yf) at (4,0) {$\mathscr{Y}$};
    
    \draw[->] (X) -- (Y);
    \draw[->] (Y) -- (Yf);

     % Add a rectangle around nodes x and y
    \draw[draw=black] (-1,0.75) rectangle (3,-0.75);
    \node[] at (2.5,-0.55) {$e \in E$};
\end{tikzpicture}
    \caption{The Multicut Problem can be understood as a Bayesian Network which aims to predict the optimal partitioning $\mathscr{Y}$.}
    \label{fig:MP_bayesiannetwork}
\end{figure}